\documentclass[pdflatex,sn-mathphys-num,iicol]{sn-jnl}

\usepackage{natbib}
\usepackage{verbatim}
\usepackage{bbm}
\usepackage{colortbl}
\usepackage{makecell}
\usepackage{pifont}

\usepackage{graphicx}\usepackage{multirow}\usepackage{amsmath,amssymb,amsfonts}\usepackage{amsthm}\usepackage{mathrsfs}\usepackage[title]{appendix}\usepackage{xcolor}\usepackage{textcomp}\usepackage{manyfoot}\usepackage{booktabs}\usepackage{algorithm}\usepackage{algorithmicx}\usepackage{algpseudocode}\usepackage{listings}

\def\eg{\textit{e.g.}}
\def\ie{\textit{i.e.}}

\definecolor{yellow}{rgb}{1,1, 0.7}
\definecolor{lightyellow}{rgb}{1,1, 0.8}
\definecolor{orange}{rgb}{1, 0.85, 0.7}
\definecolor{tablered}{rgb}{1, 0.7, 0.7}

\definecolor{cyan}{named}{black}

\theoremstyle{thmstyleone}

\theoremstyle{thmstyletwo}

\theoremstyle{thmstylethree}

\begin{document}

\title[4DGS]{4D Gaussian Splatting: Modeling Dynamic Scenes with Native 4D Primitives}

\author[1]{\fnm{Zeyu} \sur{Yang}}
\equalcont{These authors contributed equally to this work.}

\author[1]{\fnm{Zijie} \sur{Pan}}
\equalcont{These authors contributed equally to this work.}

\author[2]{\fnm{Xiatian} \sur{Zhu}}

\author*[1]{\fnm{Li} \sur{Zhang}}\email{lizhangfd@fudan.edu.cn}

\author[1]{\fnm{Jianfeng} \sur{Feng}}

\author[1]{\fnm{Yu-Gang} \sur{Jiang}}

\author[3]{\fnm{Philip H.S.} \sur{Torr}}

\affil*[1]{\orgname{Fudan University}, \orgaddress{\postcode{200433}, \state{Shanghai}, \country{China}}}

\affil[2]{\orgname{University of Surrey}, \orgaddress{\postcode{GU2 7XH}, \state{Guildford}, \country{UK}}}

\affil[3]{\orgname{University of Oxford}, \orgaddress{\postcode{OX1 2JD}, \state{Oxfordshire} \country{UK}}}

\affil[]{\textbf{Project page:} \url{https://fudan-zvg.github.io/4d-gaussian-splatting}}

\abstract{
Dynamic 3D scene representation and novel view synthesis are crucial for enabling immersive experiences required by AR/VR and metaverse applications. 
It is a challenging task due to the complexity of 
unconstrained real-world scenes and their temporal dynamics.
In this paper, we reformulate the reconstruction of a time-varying 3D scene as approximating its underlying spatiotemporal 4D volume by optimizing a collection of native 4D primitives, \ie, 4D Gaussians, with explicit geometry and appearance modeling.
Equipped with a tailored rendering pipeline, our representation can be end-to-end optimized using only photometric supervision while free viewpoint viewing at interactive frame rate, making it suitable for representing real world scene with complex dynamic.
This approach has been the first solution to achieve real-time rendering of high-resolution, photorealistic novel views for complex dynamic scenes.
{\color{cyan}
To facilitate real-world applications, we derive several compact variants that effectively reduce the memory footprint to address its storage bottleneck. 
Extensive experiments validate the superiority of 4DGS in terms of visual quality and efficiency across a range of dynamic scene-related tasks (e.g., novel view synthesis, 4D generation, scene understanding) and scenarios (e.g., single object, indoor scenes, driving environments, synthetic and real data).
}
}

\maketitle

\section{Introduction}
\label{sec:intro}
Modeling dynamic scenes and rendering novel views is crucial in computer vision and graphics, which has received increasing attention from both industry and academia because of its potential value in a wide range of AR/VR applications.
Whilst recent breakthroughs (e.g., NeRF~\cite{mildenhall2020nerf}) have achieved photorealistic rendering for static scenes~\cite{barron2021mip,verbin2022ref}, handling dynamic scenes remains a significant challenge. 
The temporal dynamics of scenes introduce additional complexity.
The central challenge lies in preserving intrinsic correlations and sharing relevant information across different time steps to create a compact representation, while minimizing interference between unrelated spacetime locations.

\begin{figure}[t]
    \centering
    \includegraphics[width=0.9999\linewidth]{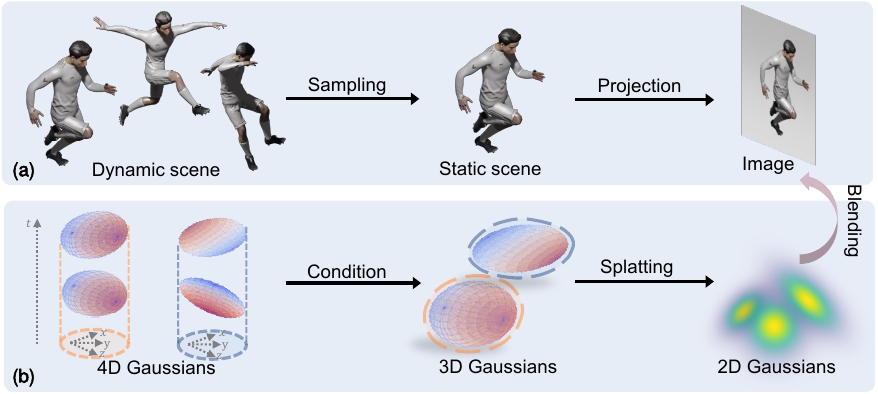}
    \caption{
    Schematic illustration of (a) our proposed spatio-temporal 4D volume learning framework for dynamic scenes, corresponding to (b) the breakdown process of our 4D Gaussian Splatting model, which transitions from 4D Gaussian representations to 2D Planar Gaussian forms in space and time. 
}
    \label{fig:teaser}
\end{figure} Existing novel view synthesis methods for dynamic scenes can be categorized into two groups. 
The first group directly learns parametric 6D plenoptic functions using implicit neural models, employing structures such as MLPs \cite{li2022neural}, grids \cite{wang2022mixed}, or their low-rank decompositions \cite{fridovich2023kplane, cao2023hexplane, attal2023hyperreel}, without explicitly modeling scene motion. 
Their ability to capture correlations across varying spatiotemporal locations relies on the intrinsic properties of the selected data structure. As a result, these methods may experience interference due to shared parameters across spatiotemporal locations or struggle to leverage the inherent correlations arising from object motion effectively.

The second group posits that scene dynamics underlie the motion and deformation of a canonical 3D model \cite{pumarola2021d, nerfplayer, abou2022particlenerf, luiten2023dynamic}. 
These approaches explicitly model scene motion, potentially improving the utilization of spatial and temporal correlations. 
However, global tracking of object motion involves inherent ambiguities and risks singularities when dealing with complex motions in real-world scenes, such as sudden appearances and disappearances, which limits their capability and scalability in unconstrained environments.

To address these fundamental limitations,
in this work we reformulate the modeling of dynamic scenes as {\bf\em a generic, spatio-temporal 4D volume learning} problem.
This novel perspective
not only introduces {\em an explicit representation}, but also makes {\em minimal assumptions about how motion is composed}, paving the way for {\em a versatile dynamic scene learning framework}.
Concretely, we propose to use a native 4D primitive \textbf{4D Gaussian} that models the geometry and motion of a time-varying 3D scene at the same time.
The adopted 4D Gaussian parameterized by anisotropic ellipses that can rotate arbitrarily in space and time fits naturally within the intrinsic spatio-temporal manifold. Furthermore, we introduce Spherindrical Harmonics, a generalized form of Spherical Harmonics, to represent the time evolution of appearance in dynamic scenes, along with a dedicated splatting-based differentiable rendering pipeline. 
Our 4DGS model is the first to support end-to-end training and real-time rendering of high-resolution, photorealistic novel views of complex dynamic scenes.

We make the following {\bf contributions}: 
\textbf{(i)} 
We propose to use a set of anisotropic 4D Gaussians equipped with 4D Spherindrical Harmonics to represent the dynamic 3D scene, which is the first native 4D primitive.
\textbf{(ii)} 
We cast the novel view synthesis of the dynamic scene as to optimize the proposed 4D Gaussian primitives for direct fitting the underlying spatio-temporal volume, and introduce the necessary differentiable rasterization pipeline.
\textbf{(iii)} 
Extensive experiments validate the superiority of 4DGS over previous alternatives
in terms of visual quality and efficiency across a variety of dynamic scene-based tasks (\eg, novel view synthesis, content generation, scene understanding) and scenarios (\eg, single object, indoor and outdoor scenes, synthetic and real data). 

{\color{cyan}
A preliminary version of this work has been presented~\cite{yang20234dgs}.
Moving beyond our initial focus on novel view synthesis,
here we significantly extend and explore the proposed 4DGS 
from {\em a versatile dynamic scene representation perspective} and manage to address the storage bottleneck.
In comparison, we summarize several important improvements:
\textbf{(i) Introducing more compact variants:} 
There is a trade-off between compactness and versatility governed by how much assumption of the scene's dynamic hold by the representation. 
We prioritize the latter while leaving room to inject motion priors via regularization. 
As a result, the size of vanilla 4DGS may increase rapidly with the inherent large 4D volume.
However, this is not a fundamental limitation. We introduce several model compression strategies across multiple dimensions, including parameter size, the number of Gaussian points, that substantially alleviate storage demands and reduce memory footprint.
\textbf{(ii) Application extension:} We first extend our model to handle large-scale dynamic urban scenes, which present additional challenges such as sparse viewpoints and unbounded environments. 
We also enhance the ability to create 4D content from monocular video via a generative approach. 
Finally, we demonstrate the proposed representation can enhance time-consistent scene understanding, providing object segmentation across both space and time.
\textbf{(iii) Extensive evaluation:} We conduct comprehensive evaluations of model compression and the newly added applications on more benchmarks.
} 
\section{Related work}\label{sec:related}

\subsection{Novel view synthesis}

\noindent \textbf{Novel view synthesis for static scenes}
The field of novel view synthesis has received widespread attention initiated by NeRF~\cite{mildenhall2020nerf}.
Querying a MLP for hundreds of points along each ray, NeRF is significantly limited in training and rendering. Subsequent works attempted to improve the speed~\cite{chen2022tensorf,sun2022dvgo,hu2022efficientnerf,Chen2023factor,fridovich2022plenoxels,muller2022instant} and enhance rendering quality~\cite{zhang2020nerf++, verbin2022ref, barron2021mip, barron2022mip360,barron2023zipnerf}.
Later, 3D Gaussian Splatting (3DGS)~\cite{kerbl3Dgaussians} 
achieves real-time rendering at high resolution thanks to its fast differentiable rasterization pipeline while retaining the merits of volumetric representations.
Subsequent works~\cite{chen2024survey,wu2024recent} have improved rendering~\cite{yu2024mip,liang2024analytic,cheng2024gaussianpro,song2024sa,ye2024absgs} and geometric accuracy~\cite{guedon2024sugar,huang20242d,yu2024gaussian}. 
Here, we explore the potential of Gaussian primitives in representing dynamic scenes.

\noindent \textbf{NeRF-based dynamic view synthesis}
Synthesizing novel views of a dynamic scene in a desired time is a more challenging task. The intricacy lies in capturing the intrinsic correlation across different timesteps. 
Inspired by NeRF, one research line attempts to learn a 6D plenoptic function represented by a well-tailored structure~\cite{li2022neural, fridovich2023kplane, cao2023hexplane, wang2022mixed, attal2023hyperreel}. 
However, these methods struggle with the coupling between parameters. 
An alternative approach explicitly models continuous motion or deformation~\cite{pumarola2021d, nerfplayer, abou2022particlenerf, luiten2023dynamic}.
Among them, point-based approaches have consistently been deemed promising. 

{\color{cyan}
\noindent \textbf{Compression of 3DGS}
Despite impressive novel view synthesis performance, 3DGS imposes substantial storage requirements due to the associated attributes of a large number of Gaussians. To reduce the memory footprint, \cite{lee2024compact} uses the hash grid to encode the color and quantifies other parameters through residual vector quantization (RVQ)~\cite{zeghidour2021soundstream}, while utilizing a learned volume mask to eliminate Gaussians with small size.~\cite{papantonakis2024reducing} introduces a scale- and resolution-aware redundancy score as a criterion for removing redundant primitives.~\cite{niedermayr2024compressed} compresses parameters of part of Gaussians in a sensitivity-aware manner. \cite{lu2024scaffold} stores the related Gaussians in the same anchor and dynamically spawns neural Gaussians during rendering.~\cite{chen2024hac} further organizes sparse anchor points using a structured hash grid.
We systematically explore these compression approaches 
with our dynamic scene representations.
}

\noindent \textbf{Dynamic 3D Gaussians} 
There have been recent efforts~\cite{luiten2023dynamic,yang2023deformable3dgs, wu20234dgaussians, liang2023gaufre,kratimenos2023dynmf,duan20244d,huang2024sc,xu2024longvolcap,wang2025freetimegs} in extending 3DGS for dynamic scenes. A straight approach is to incrementally reconstruct the dynamic scene via frame-by-frame training \cite{luiten2023dynamic}. 
\cite{yang2023deformable3dgs, wu20234dgaussians, liang2023gaufre} model the geometry and dynamics of scenes by joint optimizing Gaussians in canonical space and a deformation field. 
\cite{kratimenos2023dynmf} encourages locality and rigidity between points by factorizing the motion of the scene into a few neural trajectories.
These representations hold the topological invariance and low-frequency motion prior, thus well-suited for reconstructing dynamic scenes from monocular videos.
However, they assume that dynamic scenes are induced by a fixed set of 3D Gaussians and that all scene elements remain visible. 
Circumventing the need to maintain ambiguous global tracking, our approach can more flexibly deal with complex real-world scenes.

{\color{cyan}
\subsection{Applications with 3D/4D scene representations}

\noindent \textbf{Driving scene synthesis}
Efficient simulation of realistic sensor data in diverse driving scenes is crucial for scaling autonomous driving systems. 
Early works~\cite{dosovitskiy2017carla,shah2018airsim} employ traditional graphics techniques but suffer from domain gap of synthesized data and laborious manual modeling.~\cite{chen2021geosim} creates realistic novel driving videos by inserting 3D assets according to the depth map.
~\cite{xie2023s,wu2023mars,yang2023unisim,guo2023streetsurf,chen2024s} utilize neural implicit fields to reconstruct 
the scene from captured driving data, enabling novel view synthesis.
Recent works~\cite{lin2024vastgaussian,yan2024street,liu2024citygaussian,zhou2024hugs} have applied 3DGS to driving scenes, significantly improving fidelity while achieving real-time simulation. Other notable efforts focus on reconstructing scenes without relying on 2D masks or 3D bounding boxes of dynamic objects. Among them, SUDS~\cite{turki2023suds} and EmerNeRF~\cite{yang2023emernerf} decompose dynamic and static regions by optimizing optical flow, while $S^3$Gaussian~\cite{huang2024textit} and PVG~\cite{chen2023periodic} achieve this through self-supervision.

\noindent \textbf{3D/4D content generation}
Generative capabilities of 3D/4D representations have been explored. In 3D generation, DreamFields~\cite{jain2022dreamfields} and DreamFusion~\cite{dreamfusion} utilize NeRF as a base 3D representation. Magic3D~\cite{lin2023magic3d} incorporates a textured mesh based on the deformable tetrahedral grid in the refinement stage. Also, 3DGS has been applied to 3D generation \cite{tang2023dreamgaussian, yi2024gaussiandreamer,chen2024gsgen,gu2024tetsplatting}. 
For 4D generation, 4D radiance fields~\cite{cao2023hexplane,bahmani20244dfy,fridovich2023kplane} have been adopted~\cite{singer2023mav3d,bahmani20244dfy,jiang2023consistent4d,xie2024sv4d,yao2024sv4d2},
followed by Gaussian-based representations~\cite{yang2023deformable3dgs,wu20234dgaussians,yang20234dgs,huang2024sc} explored in \cite{ren2023dreamgaussian4d,yin20234dgen,zeng2024stag4d, pan2024efficient4d,wu2024sc4d,gao2024gaussianflow,wu2024cat4d}. 
We show the advantages of 4DGS in terms of both generation quality and speed in video-to-4D generation task.

\noindent \textbf{3D/4D scene segmentation}
Following the success of Segment Anything Model (SAM)~\cite{sam} in 2D image segmentation, recent works \cite{ji2024sa4d,cen2023saga,zhi2021semanticnerf,kim2024garfield,cen2023segment,ye2023gaussian,ying2024omniseg3d,chen2024gaussianeditor} extend NeRF and 3DGS for scene understanding and segmentation. 
SAGA~\cite{cen2023saga} applies multi-granularity segmentation to 3DGS using a scale gate mechanism. 
SA4D~\cite{ji2024sa4d} conducts 4D segmentation based on deformed 3DGS with single-granularity.
We validate the general efficacy of 4DGS for 4D scene understanding.
} 
\section{Method}\label{sec:method}

\begin{figure*}
    \centering
    \includegraphics[width=0.95\linewidth]{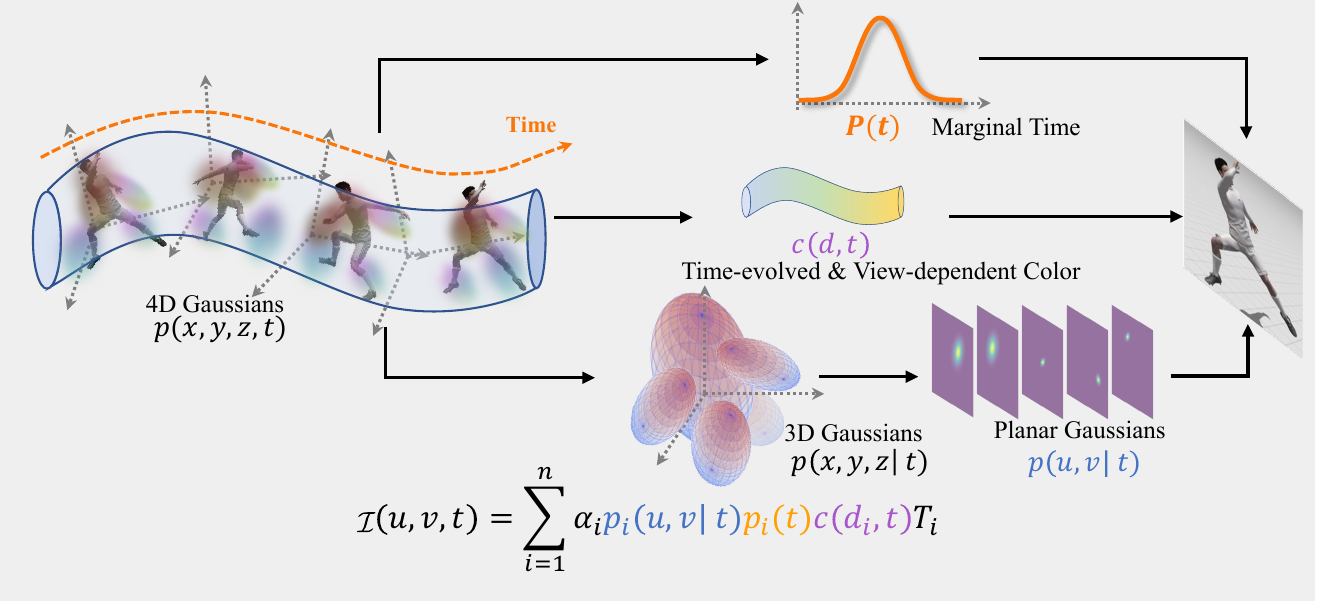}
    \caption{
    \textbf{Overview of our 4DGS.} Given a time $t$ and view $\mathcal{I}$, each 4D Gaussian is first decomposed into a conditional 3D Gaussian and a marginal 1D Gaussian. Subsequently, the conditional 3D Gaussian is projected to a 2D splat. Finally, we integrate the planar conditional Gaussian, 1D marginal Gaussian, and time-evolving view-dependent color to render the view $\mathcal{I}$.
}
    \label{fig:pipeline}
\end{figure*} 
We propose a generic scene representation, 4D Gaussian splatting (4DGS), for modeling dynamic scenes, as shown in Figure~\ref{fig:pipeline}. 
Next, we will delineate each component of 4DGS and corresponding optimization process. 
In Section~\ref{sec3.1}, we review 3D Gaussian splatting~\cite{kerbl3Dgaussians}. In Section~\ref{sec3.2}, we detail how to represent dynamic scenes and synthesize novel views with our native 4D primitives, \ie, 4D Gaussian. Finally, the optimization framework will be introduced in Section~\ref{sec3.3}.

\subsection{Preliminary: 3D Gaussian splatting}
\label{sec3.1}

3D Gaussian splatting (3DGS)~\cite{kerbl3Dgaussians} employs anisotropic Gaussian to represent static 3D scenes. Facilitated by a well-tailored GPU-friendly rasterizer, this representation enables real-time synthesis of high-fidelity novel views.

\noindent \textbf{Representation of 3D Gaussians} 
In 3DGS, a scene is represented as a cloud of 3D Gaussians. Each Gaussian has a theoretically infinite scope and its influence on a given spatial position $ x \in \mathbb{R}^3$ defined by an unnormalized Gaussian function:
\begin{equation}
\label{gaussianfunction}
    p(x|\mu, \Sigma) = e^{-\frac{1}{2} \left( x-\mu \right)^\top \Sigma^{-1} \left( x-\mu \right)},
\end{equation}
where $\mu \in \mathbb{R}^3$ is mean vector, and $\Sigma \in \mathbb{R}^{3 \times 3}$ is an anisotropic covariance matrix. 
Although the Gaussian function defined as equation~\eqref{gaussianfunction} is unnormalized, it also holds desired properties of normalized Gaussian probability density function critical for our methodology, i.e., the unnormalized Gaussian function of a multivariate Gaussian can be factorized as the production of the unnormalized Gaussian functions of its condition and margin distributions.
Hence, 
for brevity and without causing misconceptions, 
we do not specifically distinguish between equation~\eqref{gaussianfunction} and its normalized version in subsequent sections.

In~\cite{kerbl3Dgaussians}, the mean vector $\mu$ of a 3D Gaussian is parameterized as $\mu = (\mu_x, \mu_y, \mu_z)$, and the covariance matrix $\Sigma$ is factorized into a scaling matrix $S$ and a rotation matrix $R$ as $\Sigma = R S S^{\top} R^{\top}$. 
Here $S$ is summarized by its diagonal elements $S=\mathrm{diag}(s_x, s_y, s_z)$, whilst $R$ is constructed from a unit quaternion $q$. Moreover, a 3D Gaussian also includes a set of coefficients of spherical harmonics (SH) for representing view-dependent color, along with an opacity $\alpha$.

All of the above parameters can be optimized under the supervision of the rendering loss. 
During the optimization process, 3DGS also periodically performs densification and pruning on the collection of Gaussians to further improve the geometry and the rendering quality.

\noindent \textbf{Differentiable rasterization via Gaussian splatting} In rendering, given a pixel $(u, v)$ in view $\mathcal{I}$ with extrinsic matrix $E$ and intrinsic matrix $K$, its color $\mathcal{I}(u,v)$ can be computed by blending visible 3D Gaussians that have been sorted according to their depth, as described below: 
\begin{equation}
\label{alphablending}
\begin{aligned}
    \mathcal{I}(u,v) = 
    &\sum^{N}_{i=1} p_{i}(u,v;\mu_{i}^{2d}, \Sigma_{i}^{2d}) \alpha_i c_{i}(d_i) \\
    &\prod^{i-1}_{j=1} (1 - p_{i}(u,v;\mu_{i}^{2d}, \Sigma_{i}^{2d}) \alpha_j),
\end{aligned}
\end{equation}
where $c_i$ denotes the color of the $i$-th visible Gaussian from the viewing direction $d_i$, $\alpha_i$ represents its opacity, and $p_{i}(u,v)$ is the probability density of the $i$-th Gaussian at pixel $(u,v)$.

To compute $p_{i}(u,v)$ in the image space, we linearize the perspective transformation as in~\cite{zwicker2002ewa,kerbl3Dgaussians}. Then, the projected 3D Gaussian can be approximated by a 2D Gaussian. The mean of the derived 2D Gaussian is obtained as:
\begin{equation}
\label{meansplatting}
    \mu_i^{2d} = \mathrm{Proj}\left( \mu_i | E, K \right)_{1:2},
\end{equation}
where $\mathrm{Proj}\left( \cdot | E, K \right)$ denotes the transformation from the world space to the image space given the intrinsic $K$ and extrinsic $E$. The covariance matrix is given by 
\begin{equation}
\label{covsplatting}
    \Sigma_i^{2d} = (J E \Sigma E^{\top} J^{\top})_{1:2,1:2},
\end{equation}
where $J$ is the Jacobian matrix of the perspective projection. 

\subsection{4D Gaussian for dynamic scenes}
\label{sec3.2}

\noindent \textbf{Reformulation} 
To enable Gaussian splatting for modeling dynamic scenes, 
reformulation is necessary.
In dynamic scenes, a pixel under view $\mathcal{I}$ can no longer be indexed solely by a pair of spatial coordinates $(u,v)$ in the image plane; But an additional timestamp $t$ comes into play and intervenes. 
Formally this is formulated by extending equation~\eqref{alphablending} as:

{
\begin{equation}
\begin{aligned}
    \mathcal{I}(u,v,t) = &\sum^{N}_{i=1} ( p_{i}(u,v,t) \alpha_i c_{i}(d) \\
    &\prod^{i-1}_{j=1} (1-p_{j}(u,v,t)\alpha_j)). \label{eq:pixel_blending}
\end{aligned}
\end{equation}
}

Note that $p_{i}(u,v,t)$ can be further factorized as a product of a conditional probability $p_i(u,v|t)$ and a marginal probability $p_i(t)$ at time $t$, yielding:

{
\small
\begin{equation}
\begin{aligned}
    \mathcal{I}(u,v,t) = & \sum^{N}_{i=1} ( p_{i}(t) p_{i}(u,v|t) \alpha_i c_{i}(d)\\
    &\prod^{i-1}_{j=1} (1- p_{j}(t) p_{j}(u,v|t) \alpha_j)).
\label{eq:4dblending}
\end{aligned}
\end{equation}
}

We consider the underlying $p_i(x,y,z,t)$ follows a 4D Gaussian distribution. 
As the conditional distribution $p(x,y,z|t)$ is also Gaussian, we can derive $p(u,v|t)$ similarly as a planar Gaussian whose mean and covariance matrix are parameterized by equations~\eqref{meansplatting} and~\eqref{covsplatting}, respectively. 

Subsequently, it comes to the question of {\em how to represent a 4D Gaussian}.
An intuitive solution is that we adopt a distinct perspective for space and time. Consider that $(x,y,z)$ and $t$ are independent of each other, i.e., $ p_{i}(x,y,z|t) = p_{i}(x,y,z)$, then equation~\eqref{eq:4dblending} can be implemented by adding an extra 1D Gaussian $p_i(t)$ into the original 3D Gaussian. 
This design can be viewed as imbuing a 3D Gaussian with temporal extension, or weighting down its opacity when the rendering timestep is away from the expectation of $p_i(t)$. 
However, our experiment shows that while this approach can achieve a reasonable fitting of 4D manifold, it is difficult to capture the underlying motion of the scene (see Section~\ref{sec:ablation_main} for details).

\noindent \textbf{Representation of 4D Gaussian}
To address the mentioned challenge, we suggest treating the time and space dimensions equally by formulating a coherent integrated 4D Gaussian model.
Similar to~\cite{kerbl3Dgaussians}, we parameterize its covariance matrix $\Sigma$ as the configuration of a 4D ellipsoid for easing model optimization:
\begin{equation}
    \Sigma = R S S^\top R^\top,
\end{equation}
where $S=\mathrm{diag}(s_x,s_y,s_z,s_t)$ is a diagonal scaling matrix and $R$ is a 4D rotation matrix. On the other hand, a rotation in 4D Euclidean space can be decomposed into a pair of isotropic rotations, each of which can be represented by a quaternion. 
Specifically, given $q_l=(a,b,c,d)$ and $q_r=(p,q,r,s)$ denoting the left and right isotropic rotations respectively, $R$ can be constructed by:
\begin{equation}
\begin{aligned}
R &= L(q_l) R(q_r) \\
  &=
\left(
\begin{array}{rrrr}
a & -b & -c & -d \\
b &  a & -d &  c \\
c &  d &  a & -b \\
d & -c &  b &  a
\end{array}
\right)
\left(
\begin{array}{rrrr}
p & -q & -r & -s \\
q &  p &  s & -r \\
r & -s &  p &  q \\
s &  r & -q &  p
\end{array}
\right).
\end{aligned}
\end{equation}
The mean of a 4D Gaussian is represented by four scalars as $\mu =(\mu_x, \mu_y, \mu_z, \mu_t)$. Thus far we arrive at a complete representation of the general 4D Gaussian. 

Subsequently, the conditional 3D Gaussian can be derived from the properties of the multivariate Gaussian with
\begin{equation}
\begin{aligned}
    \mu_{xyz|t} &= \mu_{1:3} + \Sigma_{1:3,4}\Sigma_{4,4}^{-1} (t - \mu_t), \\
    \Sigma_{xyz|t} &= \Sigma_{1:3,1:3} - \Sigma_{1:3,4} \Sigma_{4,4}^{-1} \Sigma_{4,1:3},
\end{aligned}
\end{equation}
where the numerical subscripts denote the index across each dimension of the matrix.
Since $p_i(x,y,z|t)$ is a 3D Gaussian, $p_i(u,v|t)$ in equation~\eqref{eq:4dblending} can be derived in the same way as in equations~\eqref{meansplatting} and~\eqref{covsplatting}. Moreover, the marginal $p_i(t)$ is also a Gaussian in one-dimension:
\begin{equation}
p(t) = \mathcal{N} 
    \left( t; \mu_{4}, \Sigma_{4,4} \right).
\end{equation}
So far we have a comprehensive implementation of equation~\eqref{eq:4dblending}. Next, we can adapt the highly efficient tile-based rasterizer proposed in~\cite{kerbl3Dgaussians} to approximate this process, through considering the marginal distribution $p_i(t)$ when accumulating colors and opacities.

\noindent \textbf{Proof of unnormalized Gaussian property}
The previous formulation treats the unnormalized Gaussian defined in equation~\eqref{gaussianfunction} as a specialized probability distribution and assumes it can be factorized into the product of the unnormalized Gaussian functions of its conditional and marginal distributions. In this paragraph, we provide a concise proof for this property.

First we demonstrate that Gaussian conditional probability formula also holds in equations~\eqref{eq:pixel_blending} and~\eqref{eq:4dblending}, i.e.
\begin{equation}
    p(u,v,t) = p(t)p(u,v|t)
    \label{eq:condition_unnorm},
\end{equation}
where
\begin{align}
    p(u,v,t) &= (2\pi)^{-\frac{3}{2}}\det{(\Sigma)}^{-\frac{1}{2}} \mathcal{N}(u,v,t |\mu, \Sigma) \\
    p(t) &= (2\pi)^{-\frac{1}{2}}\det{(\Sigma_{t})}^{-\frac{1}{2}} \mathcal{N}(t |\mu_t, \Sigma_t) \\
    p(u,v|t) &= (2\pi)^{-1}\det{(\Sigma_{uv|t})}^{-\frac{1}{2}} \mathcal{N}(u,v |\mu_{uv|t}, \Sigma_{uv|t}).
\end{align}

To prove equation~\eqref{eq:condition_unnorm}, 
since Gaussian conditional probability formula holds for normalized version: $\mathcal{N}(u,v,t |\mu, \Sigma) = \mathcal{N}(t |\mu_t, \Sigma_t) \mathcal{N}(u,v |\mu_{uv|t}, \Sigma_{uv|t})$, 
we only need to prove 
\begin{equation}
    \det{(\Sigma)} = \det{(\Sigma_{t})} \det{(\Sigma_{uv|t})}.
    \label{eq:cov_det}
\end{equation}
From Gaussian property, we know that $\Sigma_{uv|t} = \Sigma_{uv} - \Sigma_{uv,t}\Sigma_{t}^{-1}\Sigma_{t, uv}$.
Then from the decomposition of $\Sigma$ by
\begin{align}
    \Sigma=&\begin{bmatrix}
                \Sigma_{uv} & \Sigma_{uv,t}\\
                \Sigma_{uv,t} & \Sigma_{t}
                \end{bmatrix} \\
          = &\begin{bmatrix}
                I & -\Sigma_{uv,t}\Sigma_{t}^{-1}\\
                0 & I
                \end{bmatrix} \\
            &\begin{bmatrix}
                \Sigma_{uv} - \Sigma_{uv,t}\Sigma_{t}^{-1}\Sigma_{t, uv} & 0\\
                0 & \Sigma_{t}
                \end{bmatrix} \\
            &\begin{bmatrix}
                I & 0\\
                -\Sigma_{t}^{-1}\Sigma_{t,uv} & I
                \end{bmatrix},
    \label{eq:mat_decompose}
\end{align}
equation~\eqref{eq:cov_det} holds immediately. The proof can be easily extended onto $p(x,y,z,t)=p(t)p(x,y,z|t)$.

\noindent \textbf{4D spherindrical harmonics} 
The view-dependent color $c_i(d)$ in equation~\eqref{eq:4dblending} is represented by a series of SH coefficients in the original 3DGS. 
To more faithfully model the dynamic scenes of the real world, we need to enable appearance variation with both varying viewpoints and evolving time.

Leveraging the flexibility of our framework, a straightforward solution is to directly use different Gaussians to represent the same point at different times. However, this approach leads to duplicated and redundant Gaussians that represent an identical object, making it challenging to optimize.
Instead, we choose to exploit 4D extension of the spherical harmonics (SH) that represents the time evolution of appearance. 
The color in equation~\eqref{eq:4dblending} could then be manipulated with $c_i(d, \Delta t)$, where $d=(\theta, \phi)$ is the normalized view direction under spherical coordinates and $\Delta t$ is time difference between the expectation of the given Gaussian and the viewpoint. 
 
Inspired by studies on head-related transfer function, we propose to represent $c_i(d,\Delta t)$ as the combination of a series of 4D spherindrical harmonics (4DSH) which are constructed by merging SH with different 1D-basis functions. For computational convenience, we use the Fourier series as the adopted 1D-basis functions. Consequently, 4DSH can be expressed as:
\begin{equation}
    Z_{nl}^{m}(t, \theta, \phi) = \cos\left( \frac{2 \pi n }{T} t \right) Y_{l}^{m} (\theta, \phi),
\end{equation}
where $Y^{m}_{l}$ is the 3D spherical harmonics. The index $l\geq0$ denotes its degree, and $m$ is the order satisfying $-l \leq m \leq l$. The index $n$ is the order of the Fourier series. The 4D spherindrical harmonics form an orthonormal basis in the spherindrical coordinate system.

\subsection{Training}
\label{sec3.3}
As 3DGS~\cite{kerbl3Dgaussians}, we conduct interleaved 
optimization and density control during training. It is worth highlighting that our optimization process is entirely end-to-end, capable of processing entire videos, with the ability to sample at any time and view, as opposed to the traditional frame-by-frame or multi-stage training approaches.

\noindent \textbf{Optimization} In optimization, we only use the rendering loss as supervision. 
In most cases, combining the representation introduced above with the default training schedule as in~\cite{kerbl3Dgaussians} is sufficient to yield satisfactory results. 
However, for scenes with more dramatic changes, we observe issues such as temporal flickering and jitter. We consider that this may arise from suboptimal sampling techniques. 
Rather than adopting the prior regularization, we discover that batch sampling in time turns out to be effective, resulting in more seamless and visually pleasing appearance of dynamic contents. 

\noindent \textbf{Densification in spacetime} 
In terms of density control, simply considering the magnitude of spatial position gradients is insufficient to assess under- and over-reconstruction over time. To address this, we incorporate the gradients of $\mu_t$ as an additional density control indicator. Furthermore, in regions prone to over-reconstruction, we employ joint spatial and temporal position sampling during Gaussian splitting.

{\color{cyan}
\section{Compact 4D Gaussian splatting}
\label{compact4dgs}
\subsection{Analysis on memory bottleneck}
4DGS can be interpreted as modeling the motion of dynamic scenes using \textit{piecewise}-linear function, analogous to how 3D Gaussian model geometry. 
Given sufficient coverage of training views, the piecewise-linear function can offer a better fitting capability than those modeling the motion with trigonometric or polynomial functions, since enough piecewise-linear functions can also approximate these functions.
However, this leads to a large number of primitives to model the motion of the same object, resulting in considerable redundancy and substantial storage demands.

To mitigate the storage bottleneck suffered by 4DGS, 
we consider two perspectives: (i) lowering the encoding cost for the attributes of each Gaussian (Section~\ref{sec:paramcompression}); (ii) decreasing the total number of primitives (Section~\ref{sec:mask4dgs}), resulting in a compact version of 4DGS, dubbed 4DGS$_{C}$.

\subsection{Parameter compression}
\label{sec:paramcompression}
As the rendering quality exhibits varying sensitivity to different attributes, we compile a few strategies to reduce their memory footprint. 
First, storing high-order SH coefficients for each Gaussian leads to considerable redundancy, particularly when modeling the time evolution of view-dependent color. To reduce this memory cost, we employ the residual vector quantization (R-VQ)~\cite{lee2024compact, zeghidour2021soundstream}.
We construct two separate quantizers for the base color and the rest components of 4DSH coefficients, respectively. 
R-VQ model is also used to compress the shape-related attributes, including scaling and rotation.
For convenience, the codebooks are constructed after the training of Gaussians. 
We fine-tune the codebook parameters after quantization to recover the lost information and performance.
To further reduce storage cost, we apply Huffman encoding~\cite{huffman1952method} for lossless compression of the codebook indices.
The reconstruction quality is more sensitive to the position parameters~\cite{lee2024compact, papantonakis2024reducing, niedermayr2024compressed}, we thus only apply half-precision quantization to them. 
We use a 8-bit min-max quantization for opacity for its negligible storage.

\subsection{Insignificant Gaussian removal}
\label{sec:mask4dgs}
We leverage an adaptive mask to learn the significance of each Gaussian during training.
We introduce a mask parameter $m \in \mathbb{R}$ per Gaussian,
with the score defined as $M = \text{Sigmoid} (m)$.
The Gaussians with low score will be masked.
To enable the gradient of $m$ during optimization, as \cite{lee2024compact} we modify the scaling matrix and opacity of each Gaussian:
\begin{align}
\label{eq:mask_prune}
    &\hat{M} = \text{stop\_gradient}(\mathbbm{1}\{M>\varepsilon\} - M) + M, \\
    &\hat{S} = \hat{M}S = \text{diag} (\hat{M}s_x, \hat{M}s_y, \hat{M}s_z, \hat{M}s_t), \\
    &\hat{\alpha} = \hat{M}\alpha,
\end{align}
where $\varepsilon$ is the threshold and $\mathbbm{1}$ denotes indicator function.

Aiming at minimizing Gaussian redundancy, we incorporate a mask compression loss: 
\begin{equation}
    \mathcal{L}_{\mathrm{mask}} = \frac{1}{N} \sum_i^N M_i,
\end{equation}
where $N$ is the number of 4D Gaussians.
During training, we also discard the Gaussians with low significance at some specific interval to reduce the computational burden.

\section{Urban scene 4D Gaussian}

\label{4dgsinthewild}
We further extend 4DGS to urban scenes 
where the unbounded range of the scene itself and limited views of driving videos present unique challenges.
To address these issues, we tailor a set of customizations to adapt the proposed representation to such scenes.
First, we leverage LiDAR point clouds, typically available with corresponding timestamps aggregated across all training frames, to initialize the 4D position of 4D Gaussians.
As the distance to the sky is much bigger than the scene's scale, we employ a learnable cube map, $M_\mathrm{sky}$, to model its appearance, which can be queried by view direction $\mathbf{r}$. For each pixel, the cube map is used to consume the remaining transmittance: 
\begin{equation}
    \mathcal{I}(u,v,t) = \mathcal{I}_{gs}(u,v,t) + (1-O_{gs}) M_\mathrm{sky}(\mathbf{r}),
\end{equation}
where $\mathcal{I}_{gs}(u,v,t)$ and $O_{gs}$ represent the color and opacity rendered by 4D Gaussian, respectively. During training, we penalize the inverse depth in the sky region using $L_{1}$ loss $\mathcal{L}_\mathrm{sky}$, with the sky mask obtained via SegFormer~\cite{xie2021segformer}.

To tackle sparse observations, we introduce several regularizations.
For each Gaussian, we apply an $L_{1}$ loss on the difference between its $q_l$ and $q_r$ to encourage sparse motion and persistence in the scene: 
\begin{equation}
\label{eq:qreg}
\mathcal{L}_\mathrm{sparse} = \frac{1}{N} \sum_{i}^{N} \| q_{l}^{i} - q_{r}^{i} \|,
\end{equation}
where $N$ is the number of 4D Gaussians. $\mathcal{L}_\mathrm{sparse}$ injects useful static prior into reconstruction under such sparse views. 

To expand supervision to more frames and make its behavior more physically plausible, we extend its lifetime by encouraging a large covariance along the temporal dimension:
\begin{equation}
\mathcal{L}_\mathrm{covt} = \frac{1}{N} \sum_{i}^{N} e^{-\Sigma^{i}_{4,4}}.
\end{equation}

As most motions in urban scenes are rigid, we deploy a rigid loss to ensure that each Gaussian exhibits similar motion to its nearby Gaussians at the current frame: 
\begin{equation}
\mathcal{L}_\mathrm{rigid} = \frac{1}{kN} \sum_{i}^{N} \sum_{j \in \text{KNN}(i)} p_{i}(t) p_{j}(t) \|v_i - v_j\|,
\end{equation}
where $\text{KNN}(i)$ is the set of $k$ nearest Gaussians at current time $t$.

Additionally, we incorporate depth supervision from LiDAR point clouds, given by the inverse $L_{1}$ loss:
\begin{equation}
\mathcal{L}_\mathrm{lidar} = \| \frac{1}{\hat{\mathcal{D}}} - \frac{1}{\mathcal{D}}\|,
\end{equation}
where $\hat{\mathcal{D}}$ is the rendered depth and $\mathcal{D}$ is the groundtruth depth derived from LiDAR data. 

The overall loss function is defined as:
\begin{equation}
\begin{aligned}
\mathcal{L}_\mathrm{urban} = &\lambda_\mathrm{l1} \mathcal{L}_\mathrm{l1}
\\
+&\lambda_\mathrm{d-ssim} \mathcal{L}_\mathrm{d-ssim} + \lambda_\mathrm{sky} \mathcal{L}_\mathrm{sky} \\ +&\lambda_\mathrm{lidar} \mathcal{L}_\mathrm{lidar}
+\lambda_\mathrm{covt} \mathcal{L}_\mathrm{covt} \\
+&\lambda_\mathrm{sparse} \mathcal{L}_\mathrm{sparse}  + \lambda_\mathrm{rigid} \mathcal{L}_\mathrm{rigid},
\end{aligned}
\end{equation}
where $\lambda_{*}$ represents the weight per term. 
To further prevent overfitting, we deprecate the time marginal filter and deactivate the temporal coefficient of 4DSH. 
The optimization process begins with several warm-up steps, during which the dual quaternions are kept equal to simulate a static scene. 
During training, we add random perturbations on time to minimize the undesired motions of the background regions.

\section{Generative 4D Gaussian}
In the preliminary version, we demonstrated that our proposed representation is capable of creating 4D content from monocular video via a reconstruction-based pipeline.
However, effectively addressing such an inherently ill-posed problem in the more general case requires stronger priors as regularization.
Existing methods~\cite{jiang2023consistent4d,yin20234dgen,ren2023dreamgaussian4d,pan2024efficient4d,zeng2024stag4d} achieve this by incorporating generative models.
As a fundamental representation, our approach is also well-suited to such approaches.
For fair comparison, we use only anchored images~\cite{yin20234dgen} and score distill sampling (SDS)~\cite{dreamfusion} as supervision, independent of 4D representation.

Suppose the input video includes $N$ frames $\{\mathcal{I}_{i}^{\text{ref}}\}$, we first utilize a multi-view diffusion model~\cite{liu2023syncdreamer} to generate $M$ pseudo views for each frame, resulting in a total of $MN$ images $\{\mathcal{I}_{i,j}\}_{i,j=1}^{N,M}$.
For SDS loss, we leverage 2D diffusion priors from Stable-Zero123~\cite{liu2023zero, stable0123}: 
\begin{equation}
\nabla \mathcal{L}_\text{SDS}(\mathcal{I}; y, t) = \mathbb{E}\left[\omega(t)(\epsilon_\phi(z_t; y, t) - \epsilon) \frac {\partial z} {\partial \mathcal{I}} \frac {\partial \mathcal{I}} {\partial \Theta}\right],
\end{equation}
where $\mathcal{I}$ is the image rendered by a 4D representation, $z$ denotes the latent encoded from $\mathcal{I}$, $z_t$ is the perturbed latent at noise level $t$, $\omega(t)$ is a weighting function at noise level $t$, and $\epsilon_\phi(z_t; y, t)$ represents the noise estimated by the Stable-Zero123 U-Net $\epsilon_\phi$ conditioned on a reference frame $y$. 

The overall loss function is defined by equation~(\ref{eq:loss_generation}):

{\footnotesize
\begin{equation}
\begin{aligned}
        \mathcal{L} = &
        \lambda_{\text{img}} \sum_{i=1}^{N} \Vert \hat{\mathcal{I}}_{i}^{\text{ref}} - \mathcal{I}_{i}^{\text{ref}}\Vert_2^2 
        + \lambda_{\text{img}}\sum_{i=1}^{N}\sum_{j=1}^{M} \Vert \hat{\mathcal{I}}_{i,j} - \mathcal{I}_{i,j}\Vert_2^2 
        \nonumber \\
        &+ \lambda_{\text{SDS}} \sum_{i=1}^{N} \mathcal{L}_\text{SDS}(\mathcal{I}^{\text{random}}_i; \mathcal{I}_{i}^{\text{ref}}, t),\label{eq:loss_generation}
\end{aligned}
\end{equation}
}
where $\lambda_{*}$ denotes the balancing weight, $\hat{\mathcal{I}}_{i}^{\text{ref}}$ and $\hat{\mathcal{I}}_{i,j}$ are the rendered images under fixed reference camera poses at timestamp $i$, and $\mathcal{I}^{\text{random}}_i$ is 
the rendered image under randomly sampled camera pose at timestamp $i$.
We note rendered images $\hat{\mathcal{I}}_{i}^{\text{ref}}$, $\hat{\mathcal{I}}_{i,j}$ and $\mathcal{I}^{\text{random}}_i$ are 4D representation specific.

\section{Segment anything in 4D Gaussian
}
\label{sec:segment}

Semantic 4D scene understanding is critical where scene representation plays a key role. We formulate a 4DGS based segmentation model inspired by 3D segmentation \cite{cen2023saga}.
Specifically, attaching a scale-gated feature vector to each 4D Gaussian, we train these features on a pre-trained 4DGS using 2D SAM~\cite{sam} masks through contrastive learning. Once trained, we can segment the 4D Gaussians based on feature matching, such as clustering or similarity with queried features. 

\noindent \textbf{Scale-gated features}
We need to obtain the mask scale $s_{\mathbf{M}}$ for a 2D mask $\mathbf{M}$. To that end, $\mathbf{M}$ is first projected into 4D space to get the projected point cloud $\mathcal{P}$. Then $s_{\mathbf{M}}$ is calculated from $\mathcal{P}$ as: 

{
\footnotesize
\begin{align}
    &s_{\mathbf{M}} = \label{eq:mask_scale} \\
    &2 \sqrt{\text{std}(\mathcal{X}(\mathcal{P}))^2 + \text{std}(\mathcal{Y}(\mathcal{P}))^2 + \text{std}(\mathcal{Z}(\mathcal{P}))^2 + \text{std}(\mathcal{T}(\mathcal{P}))^2
    }, \nonumber
\end{align}
}
where $\mathcal{X}(\mathcal{P}), \mathcal{Y}(\mathcal{P}), \mathcal{Z}(\mathcal{P})$, $\mathcal{T}(\mathcal{P})$ represent the 4D coordinates of the point cloud $\mathcal{P}$, and $\text{std}(\cdot)$ denotes standard deviation. 
Note each 2D mask only has one timestamp and the rendering of 4DGS is first conditioned on a given timestamp, the term $\text{std}(\mathcal{T}(\mathcal{P}))$ is actually zero. 

For a pre-trained 4DGS, we introduce a feature vector $\mathbf{f} \in \mathbb{R}^D$ to each 4D Gaussian, where $D$ denotes the feature dimension. To enable multi-granularity segmentation, we adopt a scale mapping $\mathcal{S}: [0, 1] \rightarrow [0, 1]^D$ to obtain a scale-gated feature at a given scale $s \in [0,1]$: 
\begin{equation}
    \mathbf{f}^s = \mathcal{S}(s) \odot  \mathbf{f}.
\end{equation}
The 2D feature $\mathbf{F}^{s}$ at a specific view could be rendered following the same process as in Section~\ref{sec3.2}, with the color $c$ replaced by the normalized feature vector $\mathbf{f^{s}}$.

\noindent \textbf{Training}
We train the soft scale gate mapping $\mathcal{S}$ and Gaussian affinity feature $\mathbf{f}$ via scale-aware contrastive learning. Specifically, the correspondence distillation loss is given by:

{\footnotesize
\begin{align}
    \mathcal{L}_{\mathrm{corr}}(s, p_1, p_2) = &\left( 1-2\mathbbm{1}\left( \mathbf{V}(s, p_1) \cdot \mathbf{V}(s, p_2) > 0 \right)\right) \nonumber \\
    &\max\left( \frac{\mathbf{F}^{s}(p_1) \cdot \mathbf{F}^{s}(p_2)}{ \| \mathbf{F}^{s}(p_1) \| \| \mathbf{F}^{s}(p_2) \|} \right),
\end{align}}
where $p_1$ and $p_2$ are different pixels at same view, $\mathbf{V}$ is the scale-aware pixel identity vector derived from 2D SAM masks (refer to~\cite{cen2023saga} for details of $\mathbf{V}$ and additional training strategy). The trained features are further smoothed by averaging the $k$-nearest-neighbors in 4D space. 

\noindent \textbf{Inference}
We consider two means of semantic analysis.
For automatic decomposition of a dynamic scene, 4D Gaussians are clustered by their smoothed features (see colored clusters in Section~\ref{sec:exp_seg}). 
This allows consistent 2D segmentation results to be rendered from any viewpoint at any timestamp.
In addition, we enable 4D segmentation using 2D point prompts from a specific view. By extracting a 2D feature from the rendered feature map as input, we filter the 4D Gaussians according to the similarity between their features and the 2D query features.
By integrating the scale mapping and scale-gated features, both segmentation algorithms support multi-granularity, controlled by the input scale.
} 
\section{Experiments}

\subsection{4D from multi-view videos}
\subsubsection{Experimental setup}

\noindent \textbf{Implementation details}
\label{sec:implement_details}
To assess the versatility of our approach, we did not extensively fine-tune the training schedule across different datasets.  
By default, we conducted training with a total of 30,000 iterations, a batch size of 4, and halted densification at the 15,000 iteration. 
We adopted the settings of~\cite{kerbl3Dgaussians} for hyperparameters such as loss weight, learning rate, and threshold.
Before optimization, we initialized both $q_l$ and $q_r$ as unit quaternions to construct identity rotation matrix and set the initial time scaling to half of the scene's duration. While the 4D Gaussian theoretically extends infinitely, we applied a Gaussian filter with marginal $p(t)<0.05$ when rendering the view at time $t$.
The LPIPS~\cite{zhang2018unreasonable} are computed using AlexNet~\cite{krizhevsky2012imagenet}.

\noindent \textbf{Plenoptic Video dataset~\cite{li2022neural}} contains six real-world scenes, each lasting ten seconds. 
For each scene, one view is reserved for testing while the other views are used for training.
To initialize the Gaussians for this dataset, we utilize the colored point cloud generated by COLMAP from the first frame of each scene.
The timestamps of each point are uniformly distributed across the scene's duration. 
Following common practice, we downsample the resolution by $2\times$ in training and evaluation.
Since the colmap may fail to reconstruct the distant background out of windows in scenes \textit{flames salmon} and \textit{coffee martini}. 
We simply adopt an environment sphere to model this part and terminate its optimization after 10,000 iterations.

{\color{cyan}
\noindent \textbf{Technicolor dataset~\cite{sabater2017technicolor}} comprises light field videos of 12 scenes captured by a $4\times4$ camera rig. 
Following the evaluation protocol in previous arts~\cite{attal2023hyperreel,STG_2024_CVPR,ex4dgs}, we evaluate our method on 5 scenes (\textit{Birthday}, \textit{Fabien},
\textit{Painter}, \textit{Theater}, \textit{Trains}) holding out the camera at the second row and the second column for testing.
We retain the spatial resolution of $2560 \times 1920$ for training and evaluation. 
For a fair comparison, we initialized our 4D Gaussian from the same sparse sfm points as~\cite{STG_2024_CVPR}. 
Since there is negligible risk of floaters in some scenes (\eg, \textit{Painter}), we disable the periodic reset of opacity in these scenes to achieve better performance. The SSIM is evaluated using \textit{sckit-image} library function with $data range=1.0$.
}

\subsubsection{Results}
\label{sec:results_dynvs}
\begin{table}[t]
\caption{
    \textbf{Comparison with the state-of-the-art methods on the Plenoptic Video benchmark.} 
}
\label{tab:dynerf}
\renewcommand\tabcolsep{1pt}
\renewcommand\arraystretch{1.2}
\scriptsize
\begin{tabular}{l|ccccc}
\hline

\hline

\hline

\hline
Method & PSNR$\uparrow$ & DSSIM$\downarrow$ & LPIPS$\downarrow$ & FPS$\uparrow$ \\  
\hline

\hline

\hline

\multicolumn{5}{l}{\textit{- Plenoptic Video}}             \\

\hline

\hline

Neural Volumes~\cite{lombardi2019neural}\footnotemark[1] & 22.80 & 0.062  & 0.295 & - \\
LLFF~\cite{mildenhall2019local}\footnotemark[1] & 23.24 & 0.076 & 0.235 & - \\
DyNeRF~\cite{li2022neural}\footnotemark[1]  & 29.58 & 0.020 & 0.099 & 0.015 \\
HexPlane~\cite{cao2023hexplane} & 31.70 & 0.014 & 0.075 & 0.56\footnotemark[3] \\
K-Planes-explicit~\cite{fridovich2023kplane} & 30.88 & 0.020 &- & 0.23\footnotemark[3] \\
K-Planes-hybrid~\cite{fridovich2023kplane} & 31.63 & 0.018 & - & - \\
MixVoxels-L~\cite{wang2022mixed} & 30.80 & 0.020 & 0.126 & 16.7 \\
StreamRF~\cite{li2022streaming}\footnotemark[1] & 29.58 & - & - & 8.3 \\
NeRFPlayer~\cite{nerfplayer} & 30.69 & 0.035\footnotemark[2] & 0.111 & 0.045 \\
HyperReel~\cite{attal2023hyperreel} & 31.10 & 0.037\footnotemark[2] & 0.096 & 2.00 \\
4DGaussian~\cite{wu20234dgaussians}\footnotemark[4] & 31.02 & 0.030 & 0.150 & 36 \\
4D-Rotor-Gaussian~\cite{duan20244d} & 31.62 & 0.030 & 0.140 & \textbf{277} \\
\rowcolor[gray]{.9}
\textbf{4DGS (Ours)} & \textbf{32.01} & \textbf{0.014} & \textbf{0.055} & 114 \\

\hline

\hline

\hline

\hline
\end{tabular}
\footnotetext[1]{Only report the result on the scene \textit{flames salmon}.}
\footnotetext[2]{Only report SSIM instead of MS-SSIM like others.}
\footnotetext[3]{Measured by ourselves using their officially released code.}
\footnotetext[4]{Results on \textit{Spinach}, \textit{Beef}, and \textit{Steak scenes}.}
\end{table} \begin{figure*}[t]
    \centering 
    \setlength{\tabcolsep}{2.0pt}
    \renewcommand{\arraystretch}{0.8}
    \begin{tabular}{ccccc} 

    \raisebox{-0.4\height}{\includegraphics[width=.18\textwidth,clip]{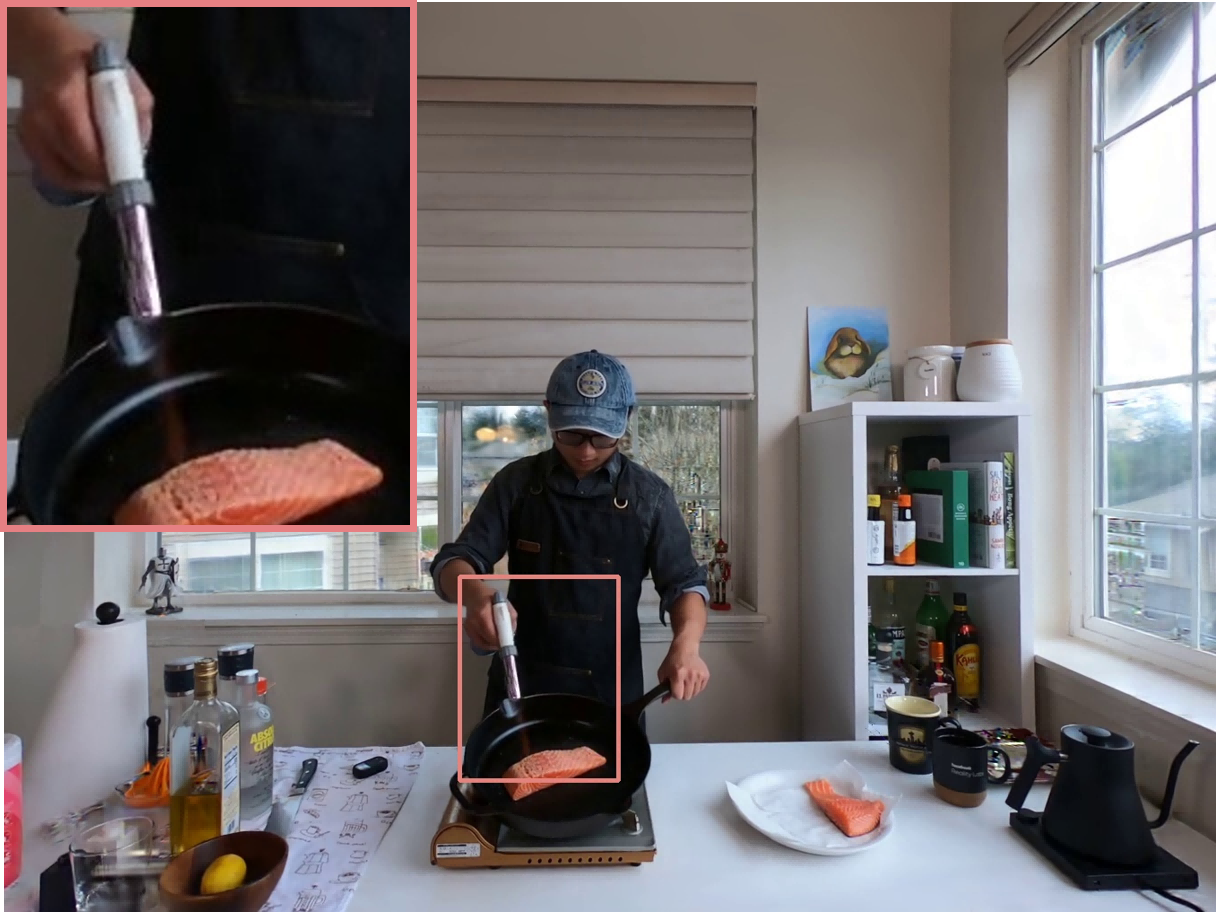}}
    &
    \raisebox{-0.4\height}{\includegraphics[width=.18\textwidth,clip]{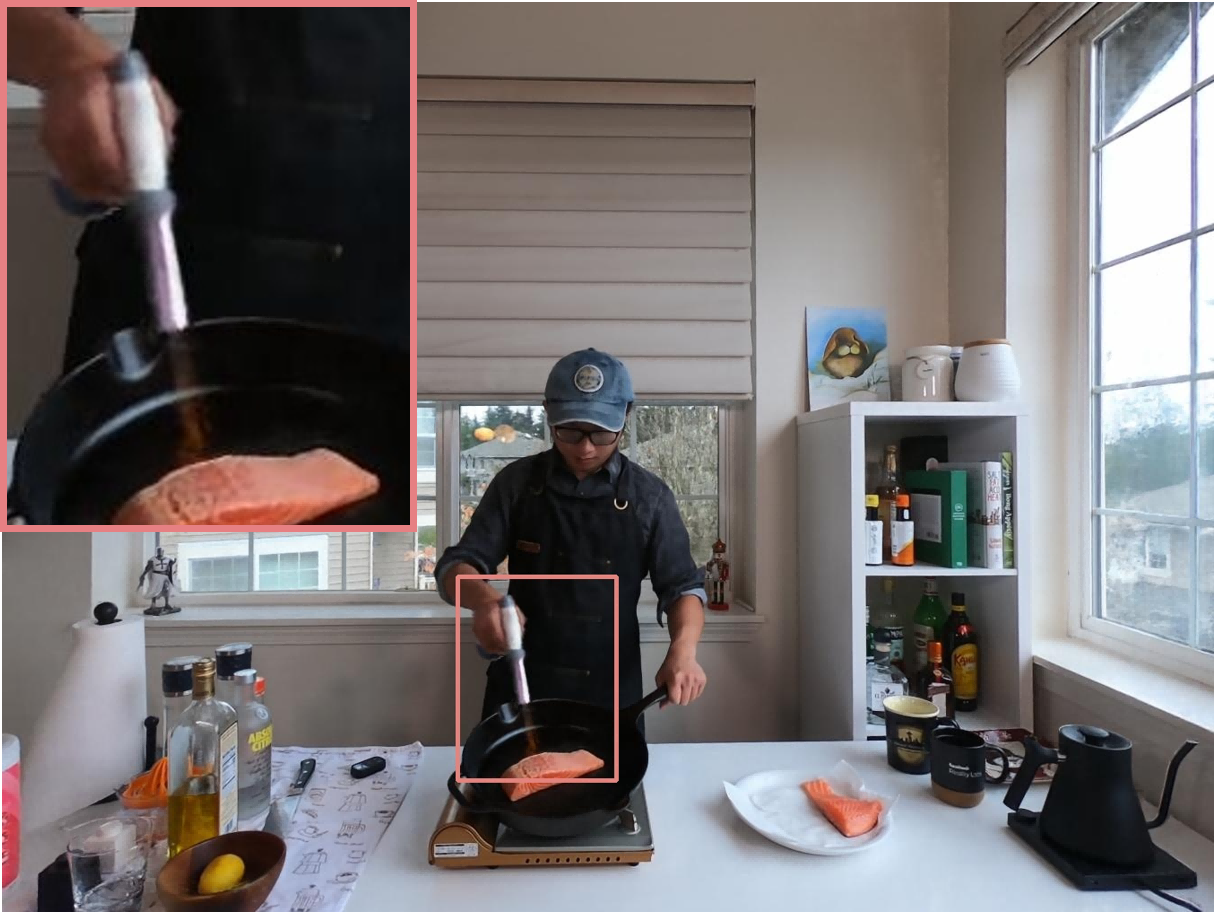}}
    &
    \raisebox{-0.4\height}{\includegraphics[width=.18\textwidth,clip]{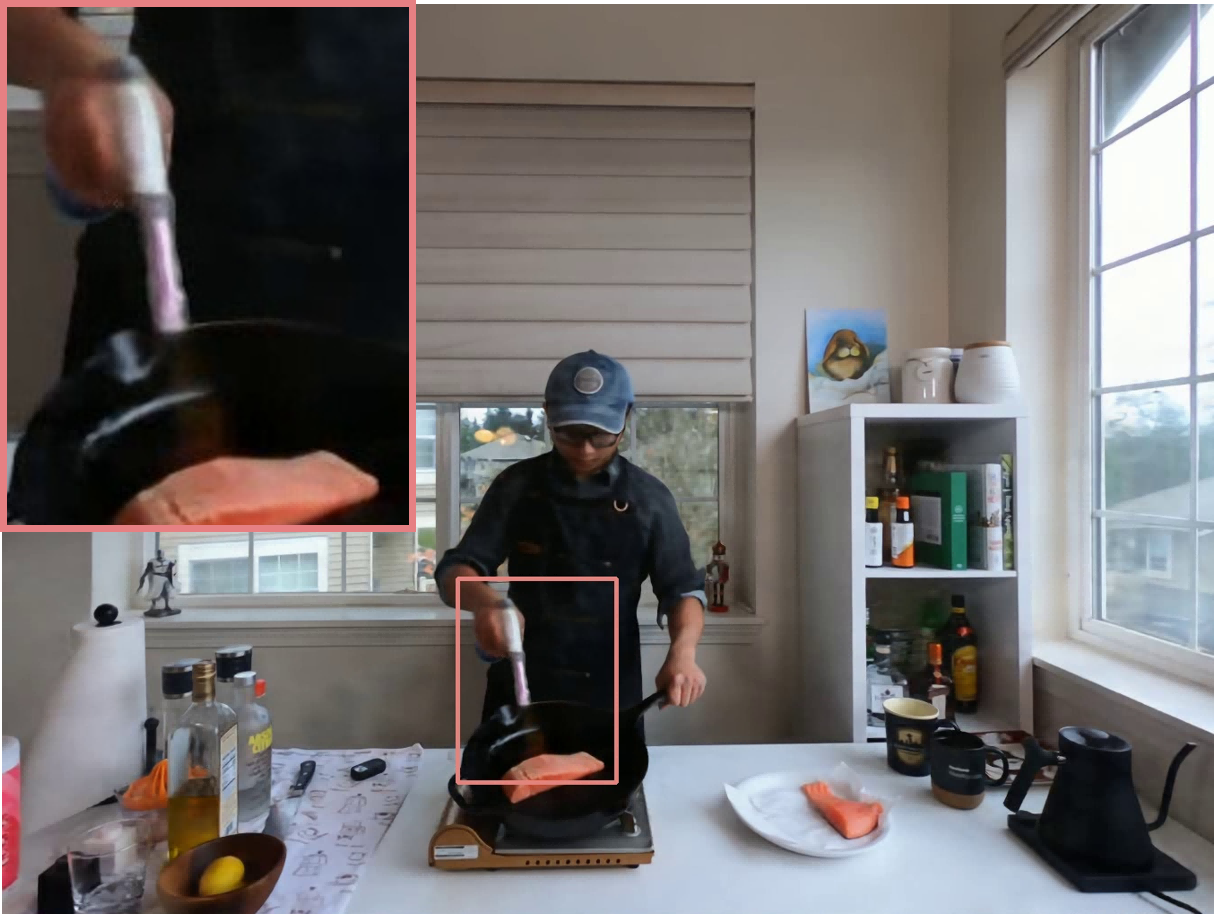}}
    &
    \raisebox{-0.4\height}{\includegraphics[width=.18\textwidth,clip]{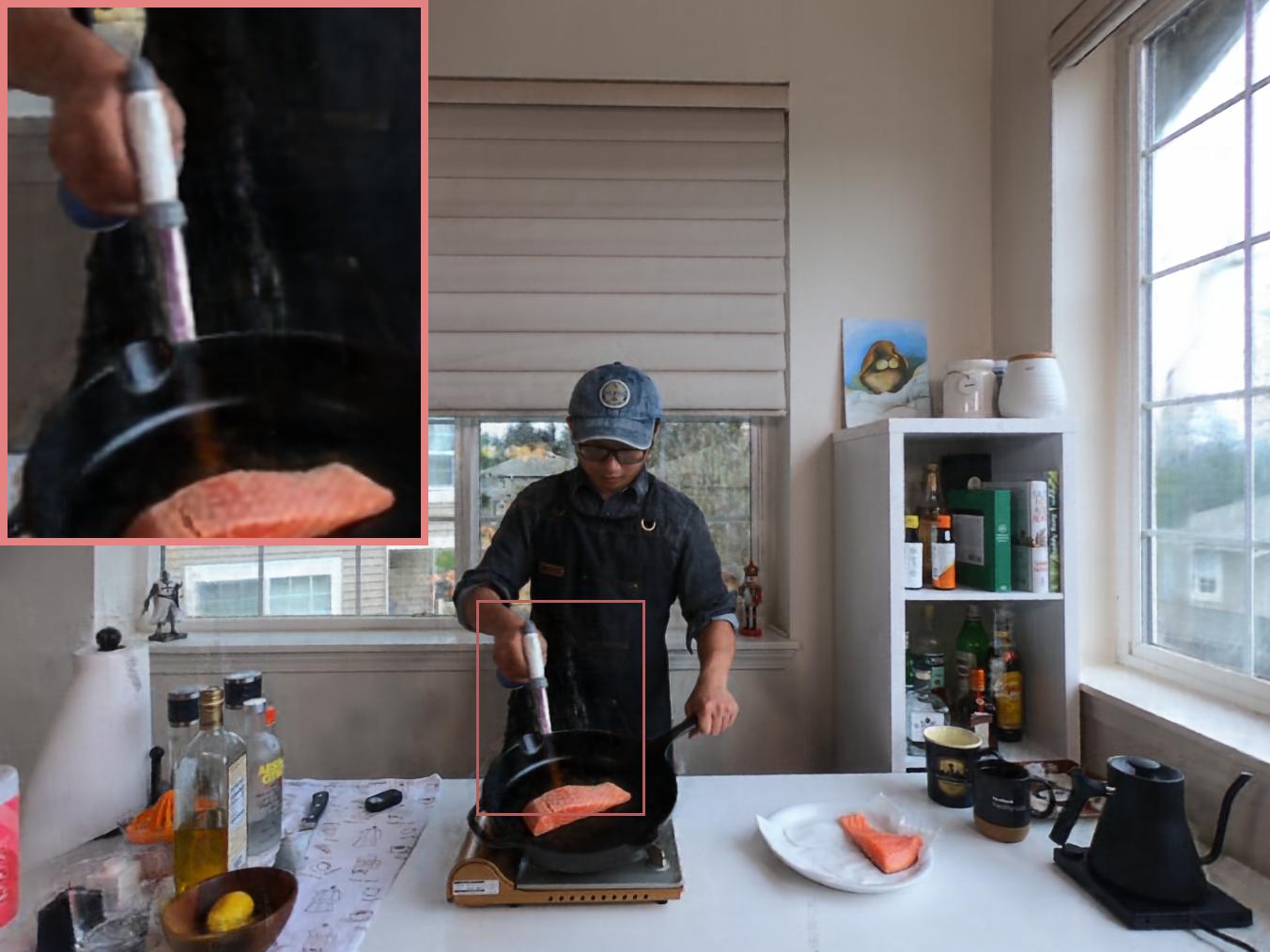}}
    &
    \raisebox{-0.4\height}{\includegraphics[width=.18\textwidth,clip]{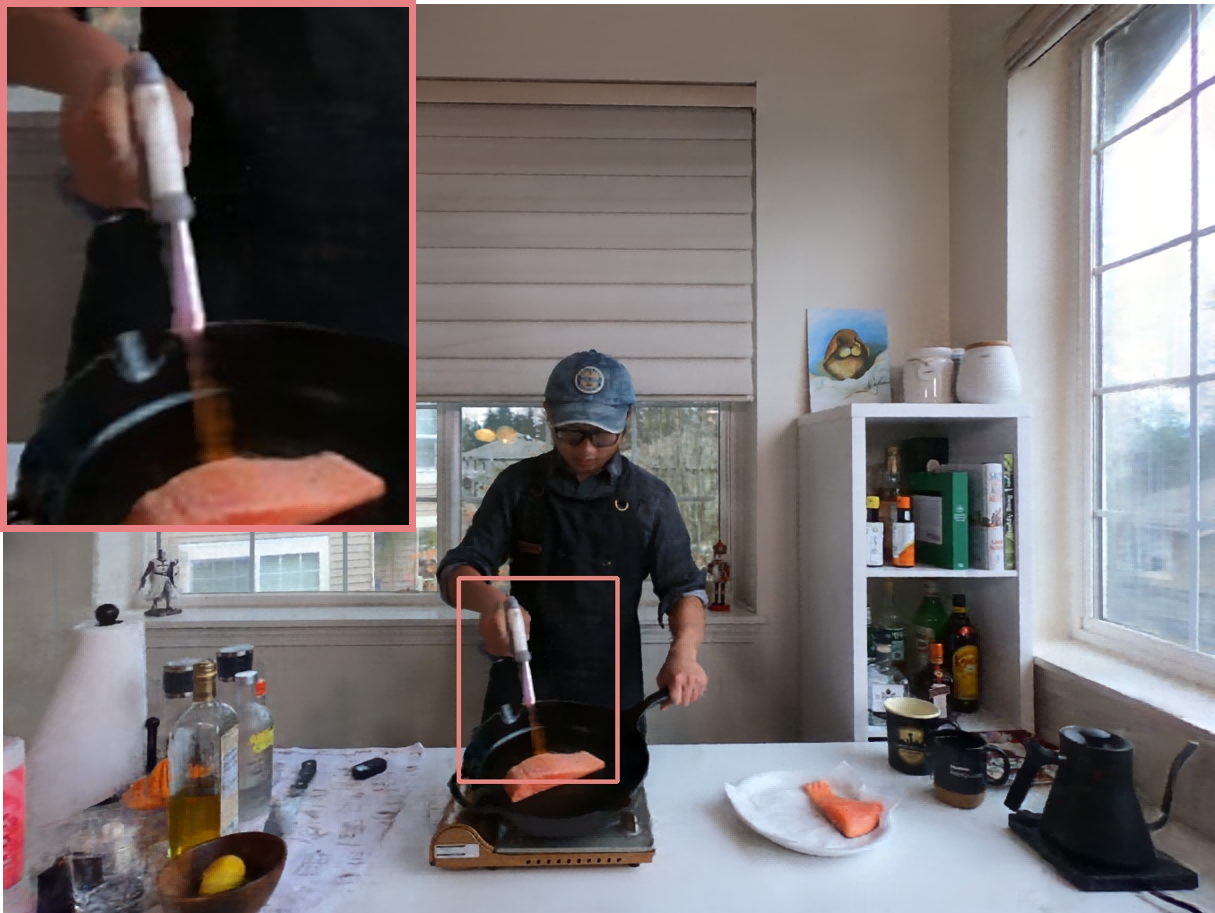}}
 \\
\footnotesize{Ours} & \footnotesize{DyNeRF~\cite{li2022neural}} & \footnotesize{K-Planes~\cite{fridovich2023kplane}} & \footnotesize{NeRFPlayer~\cite{nerfplayer}} & \footnotesize{HyperReel~\cite{attal2023hyperreel}}
 \\
\footnotesize{(114 fps)} & \footnotesize{(0.015 fps)} & \footnotesize{(0.23 fps)} & \footnotesize{(0.045 fps)} & \footnotesize{(2.00 fps)}
 \\
     \raisebox{-0.4\height}{\includegraphics[width=.18\textwidth,clip]{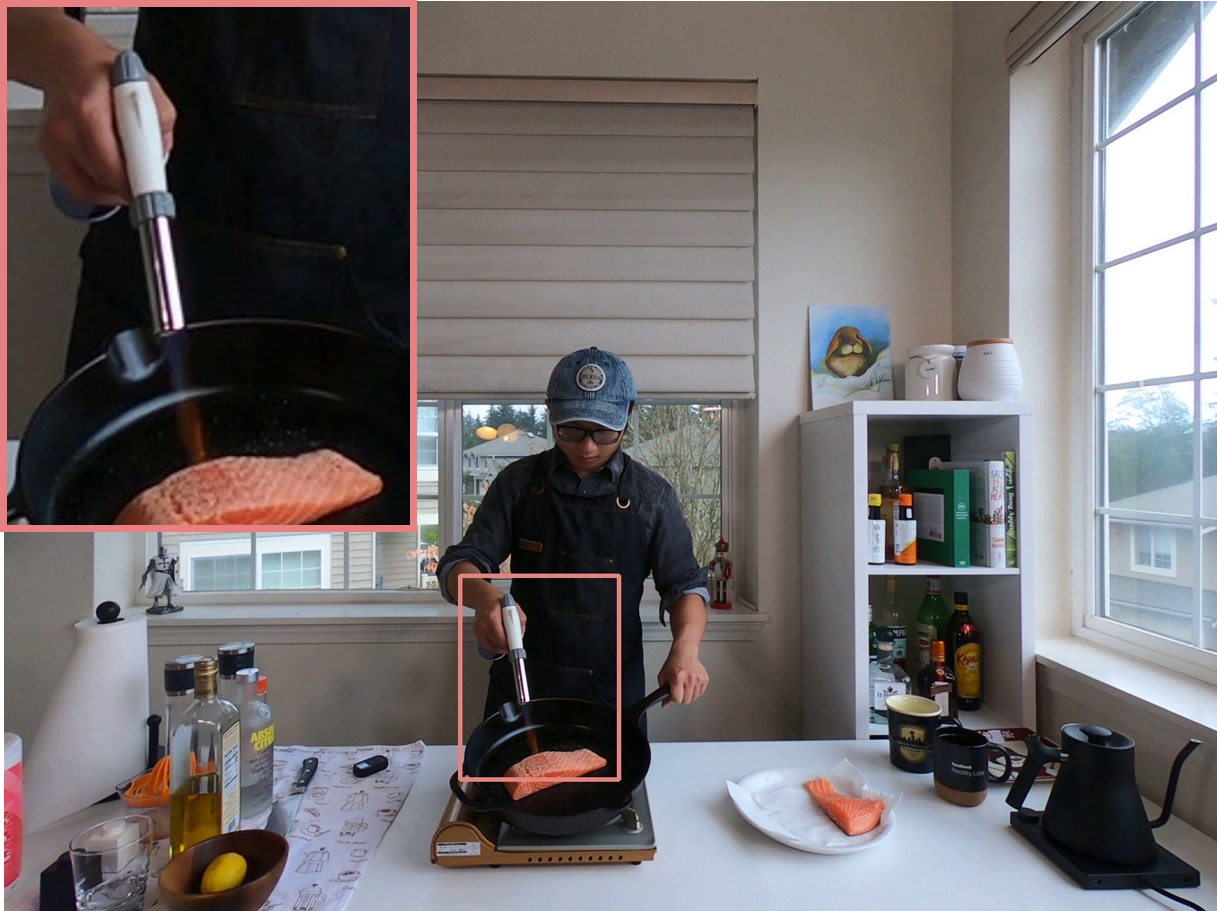}}
    &
    \raisebox{-0.4\height}{\includegraphics[width=.18\textwidth,clip]{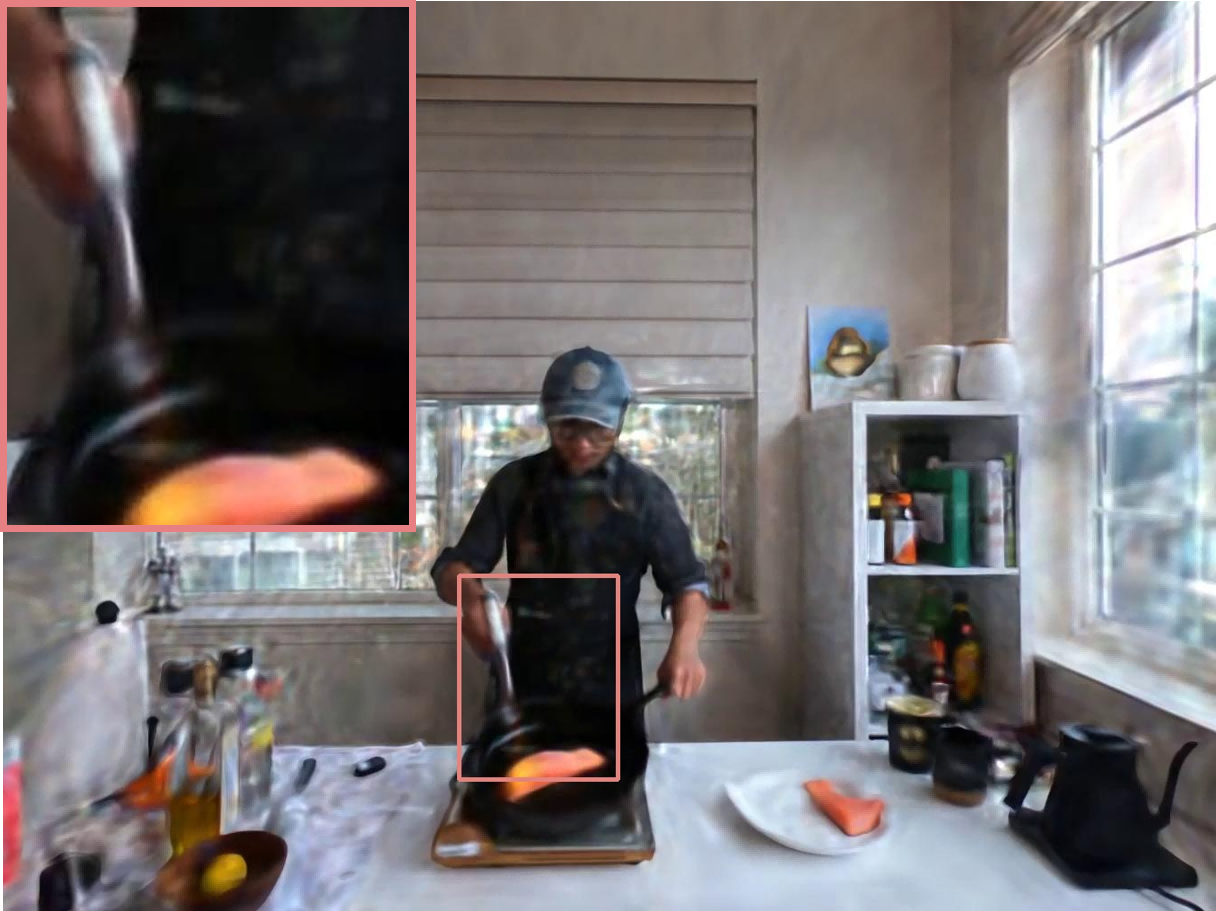}}
    &
    \raisebox{-0.4\height}{\includegraphics[width=.18\textwidth,clip]{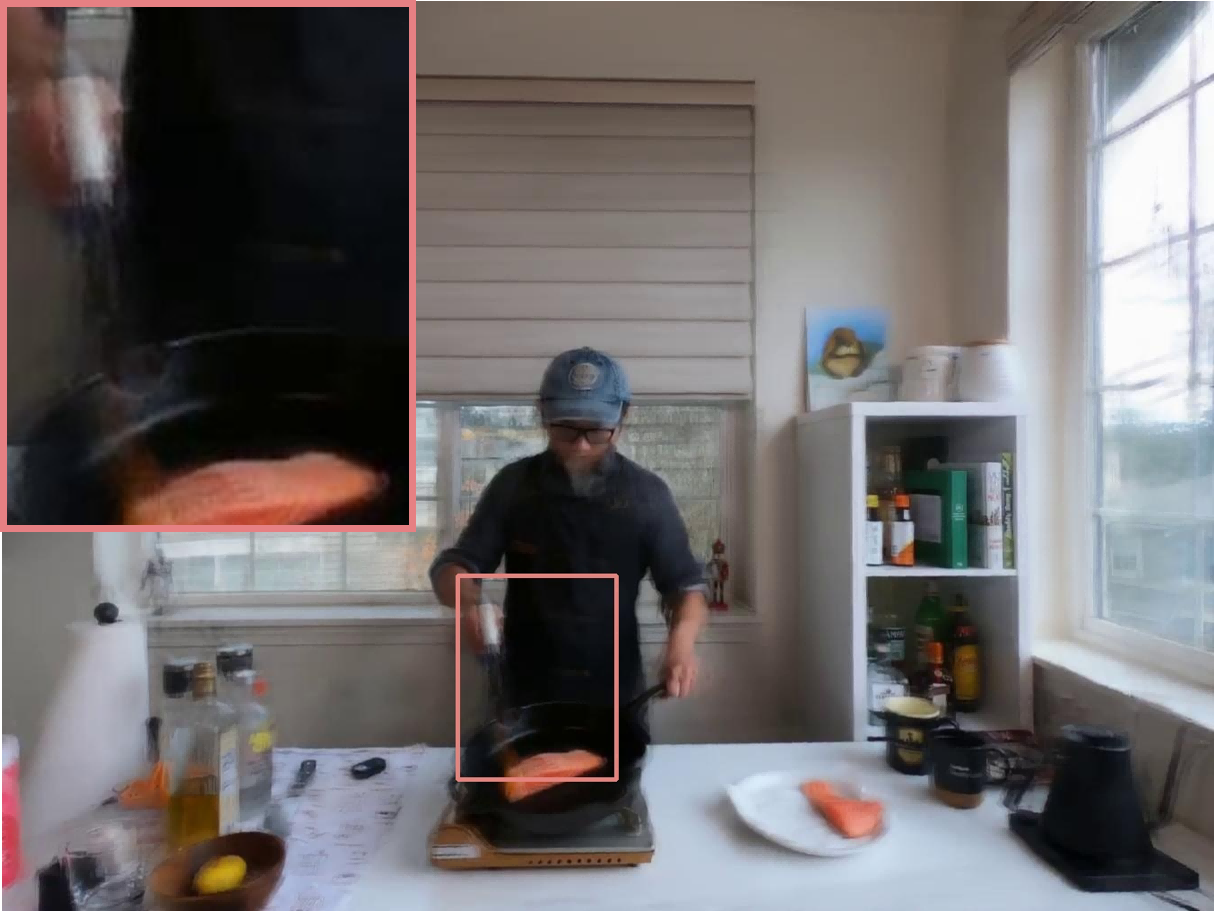}}
    &
    \raisebox{-0.4\height}{\includegraphics[width=.18\textwidth,clip]{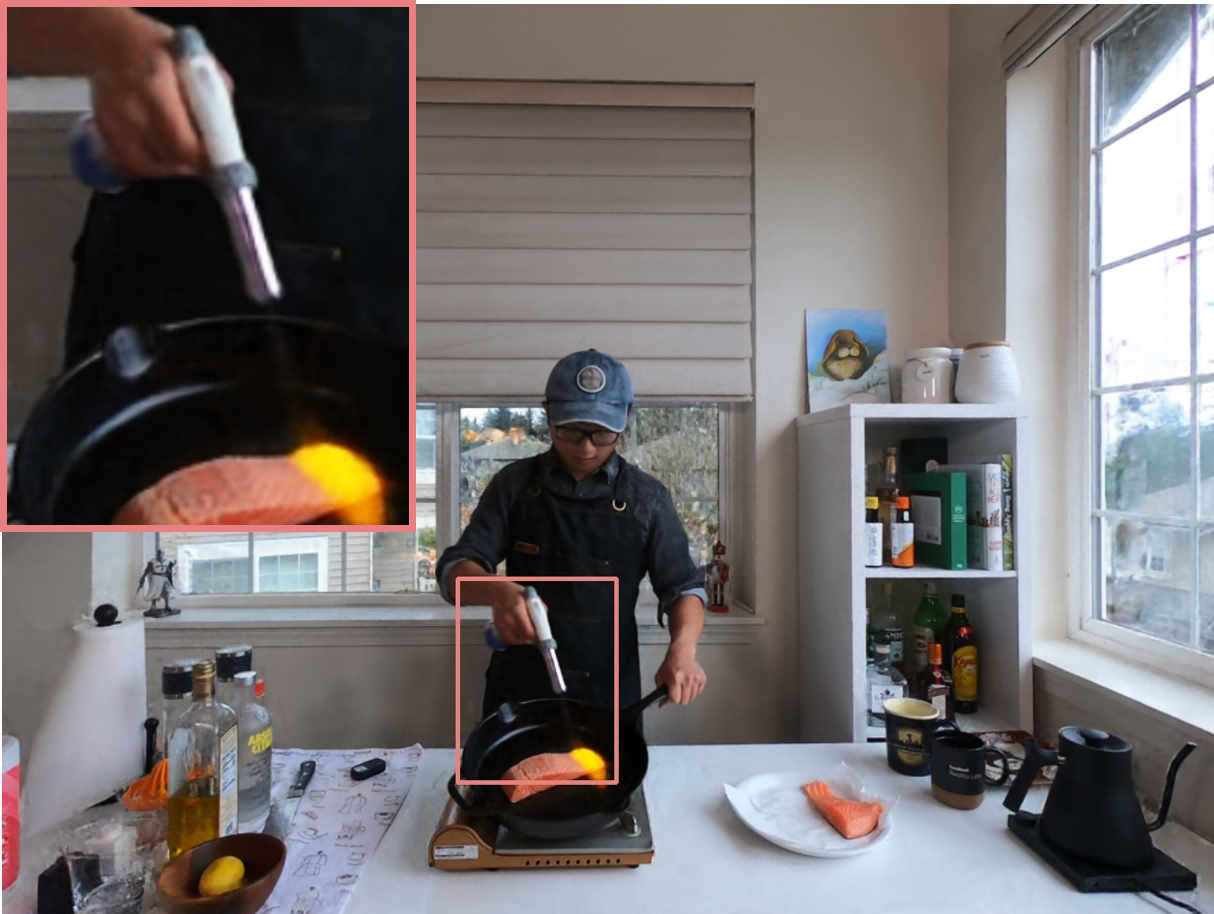}}
    &
    \raisebox{-0.4\height}{\includegraphics[width=.18\textwidth,clip]{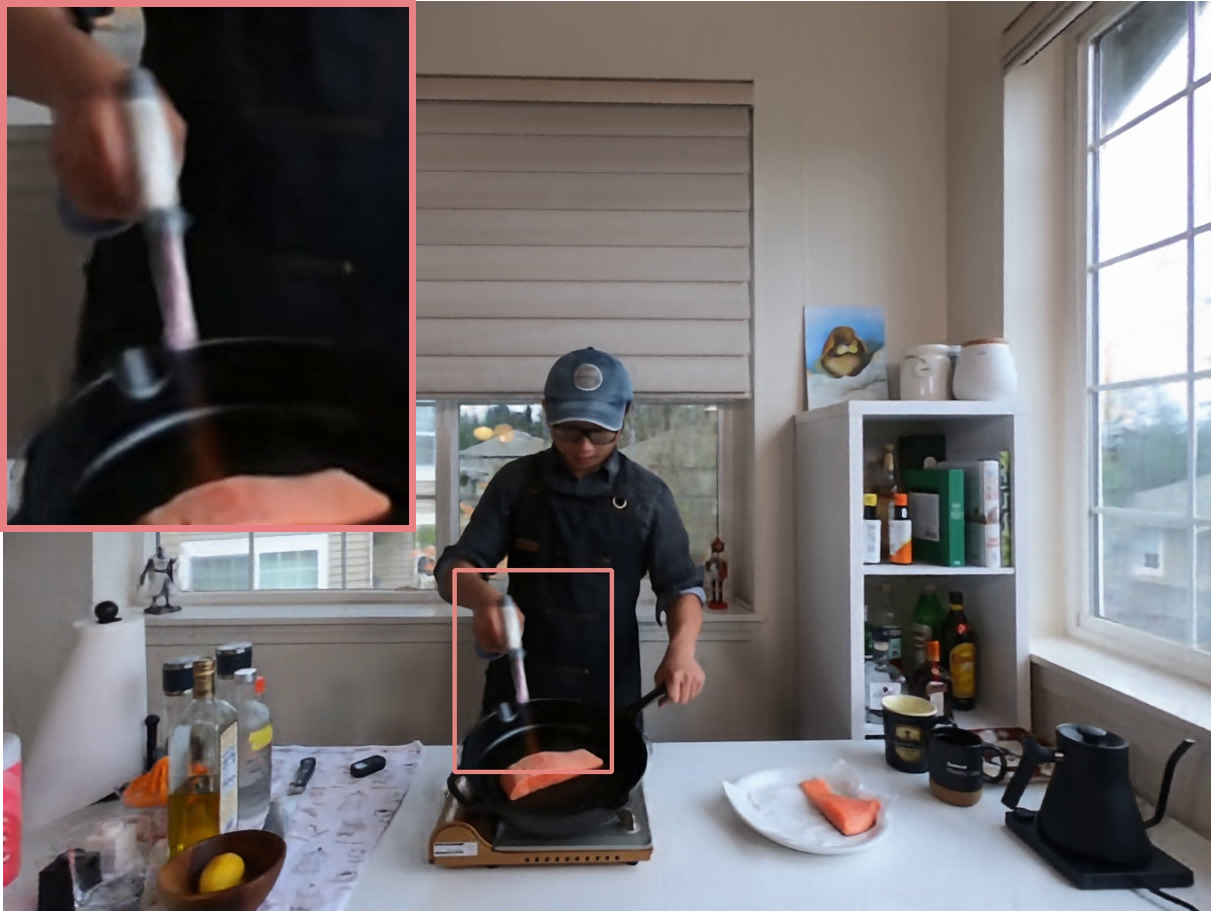}}
 \\ 
 \footnotesize{Ground truth} & \footnotesize{Neural Volumes~\cite{lombardi2019neural}} & \footnotesize{LLFF~\cite{mildenhall2019local}} & \footnotesize{HexPlane~\cite{cao2023hexplane}} & \footnotesize{MixVoxels~\cite{wang2022mixed}}
 \\
  \footnotesize{-} & \footnotesize{-} & \footnotesize{-} & \footnotesize{(0.56 fps)} & \footnotesize{(16.7 fps)}
 \\
    \end{tabular}
    \caption{Qualitative result on the \textit{flame salmon}. It can be clearly seen that the visual quality is higher than other methods in the region from the moving hands and flame gun to the static salmon. }
\label{fig:qualitative_plenoptic}
\end{figure*} \noindent \textbf{Evaluation on Plenoptic Video dataset~\cite{li2022neural}.}
Table~\ref{tab:dynerf} 
presents a quantitative evaluation on the Plenoptic Video dataset. Our approach not only significantly surpasses previous methods in terms of rendering quality but also achieves substantial speed improvements. Notably, it stands out as the sole method capable of real-time rendering while delivering high-quality dynamic novel view synthesis within this benchmark.
To complement this quantitative assessment, we also offer qualitative comparisons on the ``flame salmon'' scene, as illustrated in Figure \ref{fig:qualitative_plenoptic}.
The quality of synthesis in dynamic regions notably excels when compared to other methods. Several intricate details, including the black bars on the flame gun, the fine features of the right-hand fingers, and the texture of the salmon, are faithfully reconstructed, demonstrating the strength of our approach.

\begin{table}[t]
\caption{
{\color{cyan}
    \textbf{Quantitative comparison on the Technicolor dataset.}
}
}
\label{tab:technicolor}
\renewcommand\tabcolsep{1pt}
\renewcommand\arraystretch{1.2}
\scriptsize
\begin{tabular}{l|ccc|c}
\hline

\hline

\hline

\hline
Method & PSNR$\uparrow$ & SSIM$\uparrow$ & LPIPS$\downarrow$ & Size/Fr$\downarrow$ \\  
\hline

\hline

\hline

\multicolumn{4}{l}{\textit{- Technicolor}}             \\

\hline

\hline

DyNeRF~\cite{li2022neural}\footnotemark[1] & 31.80 & - & 0.142 & \textbf{0.6 MB}\footnotemark[2] \\
HyperReel~\cite{attal2023hyperreel}\footnotemark[1] & 32.73 & 0.906 & 0.109 & 1.2 MB\footnotemark[2] \\
4DGaussian~\cite{wu20234dgaussians}\footnotemark[1] & 30.79 & 0.843 & 0.178 & 1.0 MB\footnotemark[3] \\
Ex4DGS~\cite{ex4dgs}\footnotemark[1] & 33.62 & 0.916 & 0.088 & 1.5 MB\footnotemark[3] \\
STG~\cite{STG_2024_CVPR}\footnotemark[2] & 33.56 & 0.920 & \textbf{0.084} & 1.1 MB\footnotemark[2] \\
\rowcolor[gray]{.9}
\textbf{4DGS} & \textbf{34.03} & \textbf{0.921} & \textbf{0.084} & 37.6 MB \\
\rowcolor[gray]{.9}
\textbf{4DGS$_{C}$}\footnotemark[4] & 33.75 & 0.913 & 0.865 & 1.6 MB \\
\hline

\hline

\hline

\hline
\end{tabular}
\footnotetext[1]{Reported from~\cite{ex4dgs}.}
\footnotetext[2]{Reported from~\cite{STG_2024_CVPR}.}
\footnotetext[3]{Reported from~\cite{cho20244dscaffold}.}
\footnotetext[4]{Compact variant of our 4DGS.}
\end{table} 
{\color{cyan}
\noindent \textbf{Evaluation on Technicolor dataset.}
To validate the applicability to diverse real-world scenes, we compare our approach with previous methods on the Technicolor dataset.
The qualitative results in Figure~\ref{fig:qualitative_technicolor} demonstrate that our method has superior capability for capturing subtle details in the dynamic regions. 
The quantitative results in Table~\ref{tab:technicolor} suggest that our method achieves the highest performance across quality-related metrics. 
While there seems to be a storage bottleneck compared to other baselines since the vanilla 4DGS focuses on boosting the quality, this can be easily addressed. 
As shown in the table, when equipped with the compression strategy introduced in Section~\ref{compact4dgs}, the method (4DGS$_{C}$) achieves a significant improvement in compactness without significant compromise in quality.
}

\begin{figure*}
    \centering 
    \setlength{\tabcolsep}{2.0pt}
\begin{tabular}{cccc} 
    {\includegraphics[width=.24\textwidth,clip]{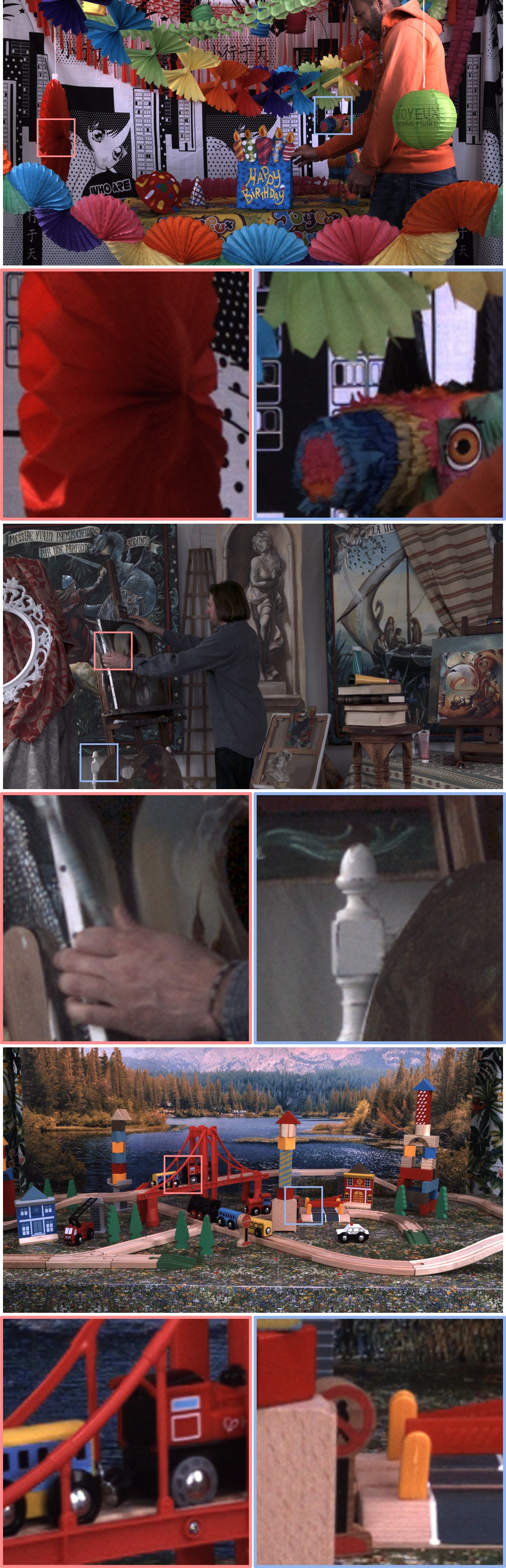}}
    &
    {\includegraphics[width=.24\textwidth,clip]{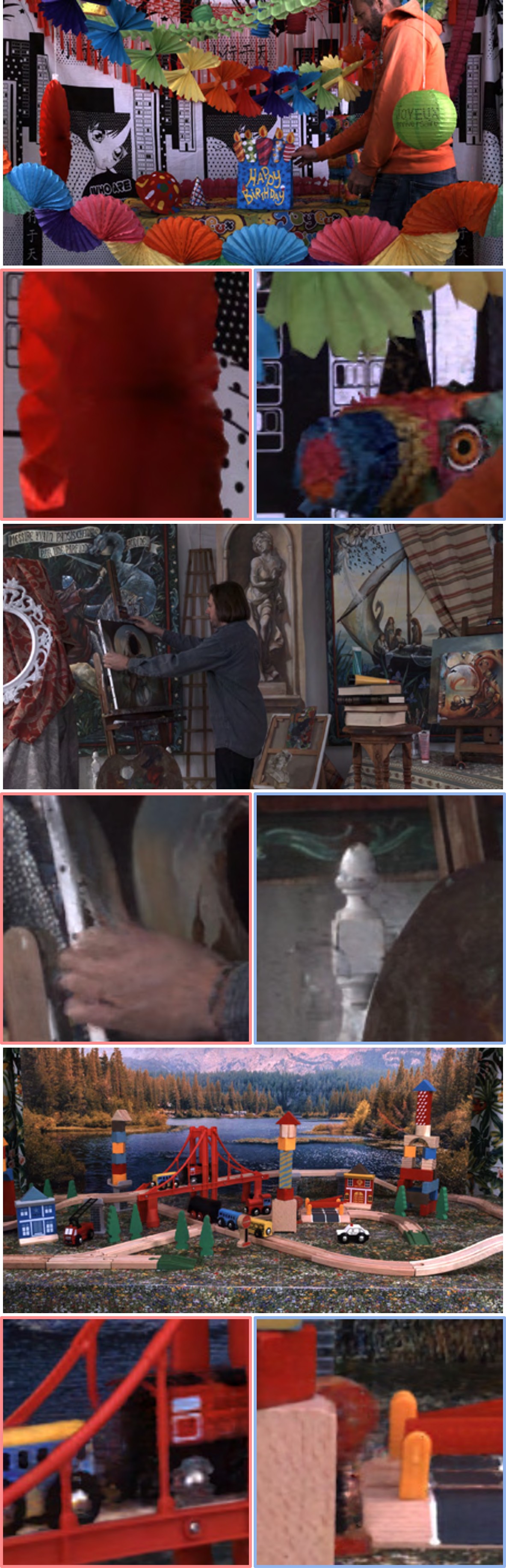}}
    &
    {\includegraphics[width=.24\textwidth,clip]{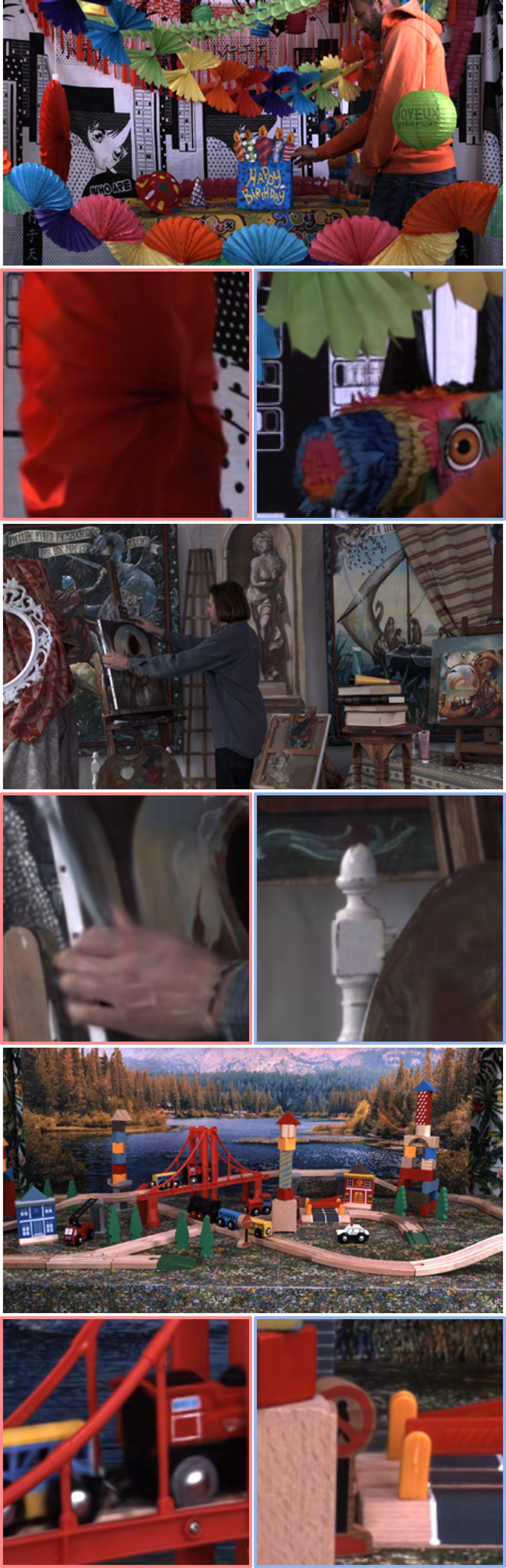}}
    &
    {\includegraphics[width=.24\textwidth,clip]{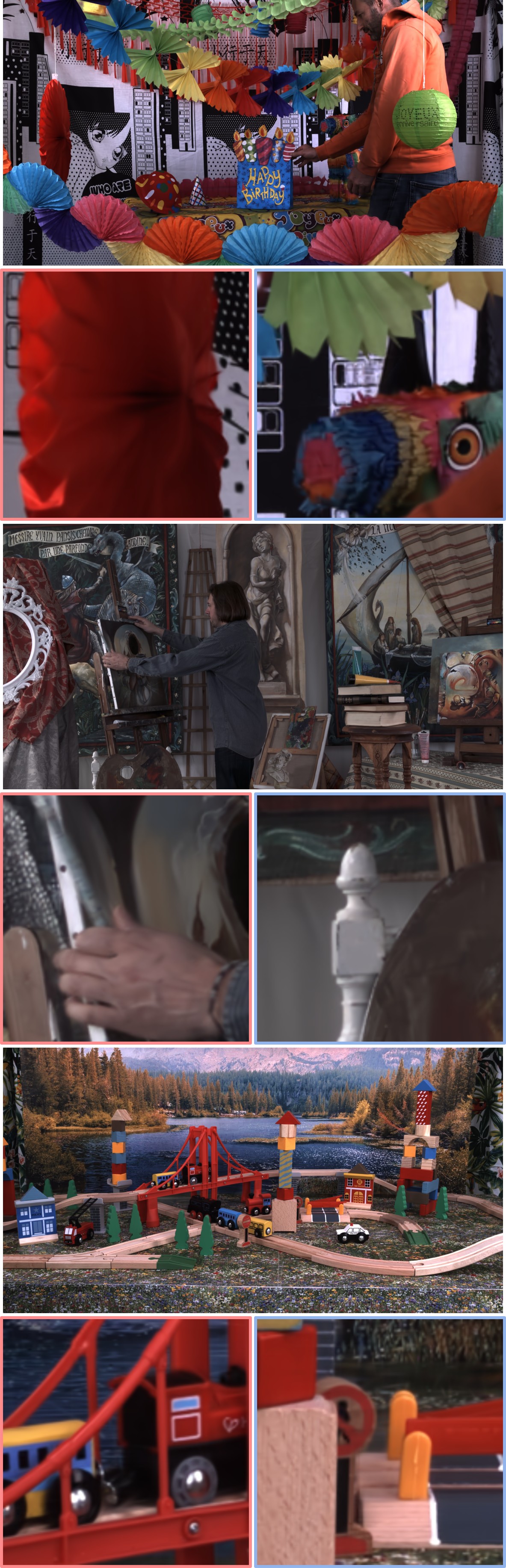}}
 \\
GT & HyperReel~\cite{attal2023hyperreel} & STG~\cite{STG_2024_CVPR} & Ours
 \\
    \end{tabular}
    \caption{{\color{cyan}
    \textbf{Qualitative comparison on the Technicolor dataset.}}
}
    \label{fig:qualitative_technicolor}
\end{figure*} 
\subsubsection{Ablation and analysis}
\label{sec:ablation_main}
\begin{table}[t]
\caption{\textbf{Ablation studies.} We ablate our framework on two representative real scenes, \textit{flame salmon} and \textit{cut roasted beef}, which have volumetric effects, non-Lambertian surfaces, and different lighting conditions. ``No-4DRot'' denotes restricting the space and time independent of each other. 
}
\label{tab:ablation_main}
\renewcommand\tabcolsep{3pt}
\renewcommand\arraystretch{1.2}
\scriptsize
\begin{tabular}{l|cc|cc}
\hline

\hline

\hline

\hline
 & \multicolumn{2}{c|}{Flame Salmon} & \multicolumn{2}{c}{Cut Roasted Beef}  \\
 & PSNR $\uparrow$ & SSIM $\uparrow$ & PSNR $\uparrow$ & SSIM $\uparrow$  \\ 
\hline

No-4DRot            & 28.78 & 0.95
& 32.81 & 0.971 \\
No-4DSH             & 29.05 & 0.96
& 33.71 & 0.97
 \\
No-Time split    & 28.89 & 0.96 & 32.86 & 0.97  \\
\rowcolor[gray]{.9}
\textbf{Full} & \textbf{29.38} & \textbf{0.96} & \textbf{33.85} & \textbf{0.98} \\

\hline
\end{tabular}
\end{table} 
\noindent \textbf{Coherent comprehensive 4D Gaussian}
\begin{figure}
    \centering
    \includegraphics[width=1.0\linewidth]{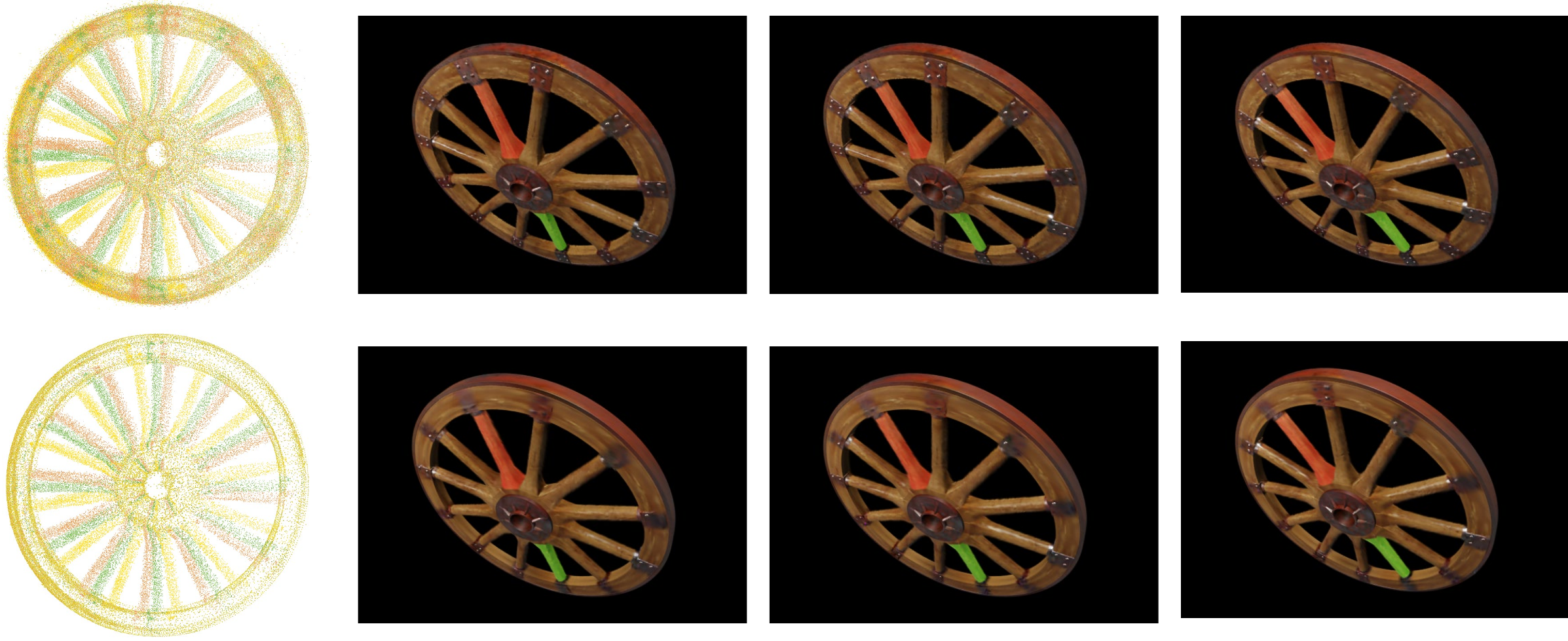}
    \caption{
    \textbf{Visualization of the time slices under Full setting (top) vs. No-4D Rot (bottom).} In the first column, we show the time slices of fitted 4D Gaussian under different settings. In the other columns, we present the rendered images. 
}
    \label{fig:slice}
\end{figure} Our novel approach involves treating 4D Gaussian distributions without strict separation of temporal and spatial elements. In Section~\ref{sec3.2}, we discussed an intuitive method to extend 3D Gaussians to 4D Gaussians, as expressed in equation~\eqref{eq:4dblending}. This method assumes independence between the spatial $(x, y, z)$ and temporal variable $t$, resulting in a block diagonal covariance matrix.
The first three rows and columns of the covariance matrix can be processed similarly to 3D Gaussian splatting.
We further additionally incorporate 1D Gaussian to account for the time dimension.

To compare our unconstrained 4D Gaussian with this baseline, we conduct experiments on two representative scenes, as shown in Table~\ref{tab:ablation_main}. 
We can observe the clear superiority of our unconstrained 4D Gaussian over the constrained baseline. This underscores the significance of our unbiased and coherent treatment of both space and time aspects in dynamic scenes.
We also qualitatively compare the sliced 3D Gaussians of these two variants in Figure~\ref{fig:slice}. It can be found that under the No-4DRot setting the rim of the wheel is not well reconstructed, and fewer Gaussians are engaged in rendering the displayed frames after filtering, despite a larger total number of fitted Gaussians under this configuration. 
This indicates that the 4D Gaussian in the No-4DRot setting has less temporal variance, which impairs the capacity of motion fitting and exchanging information between successive frames, and brings more flicker and blur in the rendered videos. 

\noindent \textbf{4D Gaussian is capable of capturing the underlying 3D movement}
\begin{figure*}[t]
    \centering 
    \setlength{\tabcolsep}{0.0pt}
    \begin{tabular}{ccccccc} 
    & \scriptsize{Coffee Martini} & \scriptsize{Cook Spinach} & \scriptsize{Cut Beef} & \scriptsize{Flame Salmon} & \scriptsize{Sear Steak} & \scriptsize{Flame Steak} \\
    \rotatebox{90}{\scriptsize{\makecell{Render Flow\\ }}} & \raisebox{-0.15\height}{\includegraphics[width=.16\textwidth,clip]{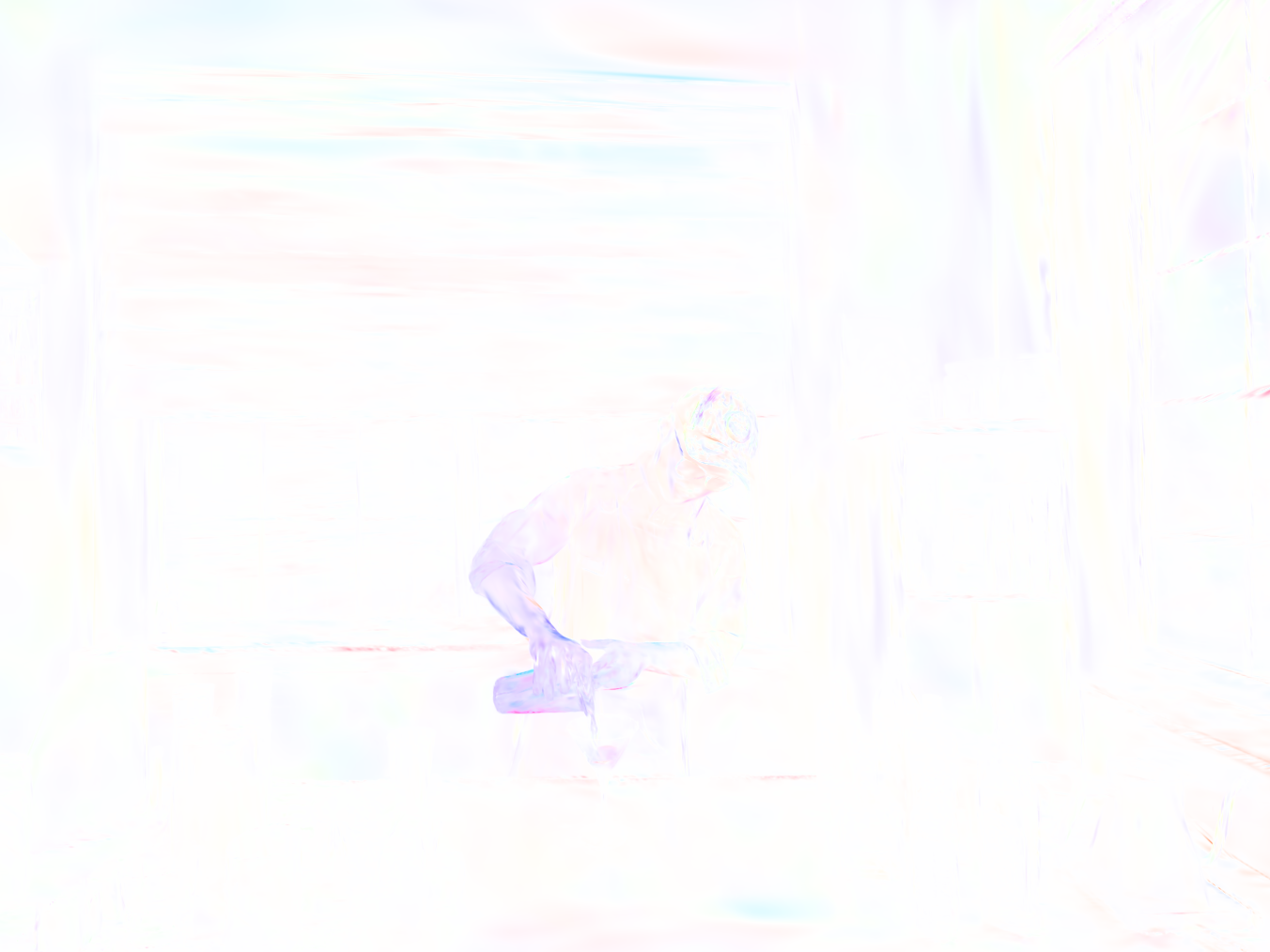}}
    &
    \raisebox{-0.15\height}{\includegraphics[width=.16\textwidth,clip]{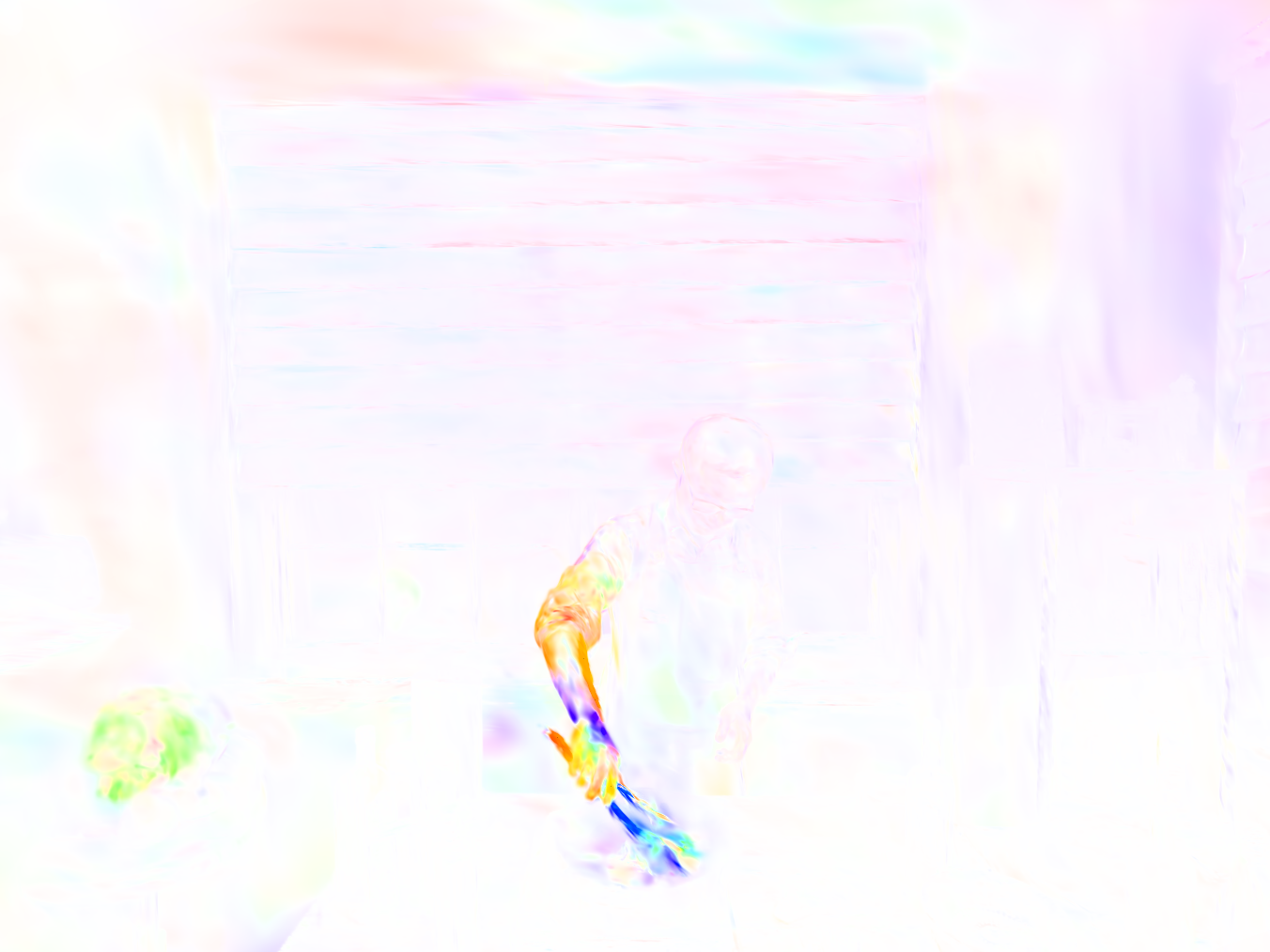}}
    &
    \raisebox{-0.15\height}{\includegraphics[width=.16\textwidth,clip]{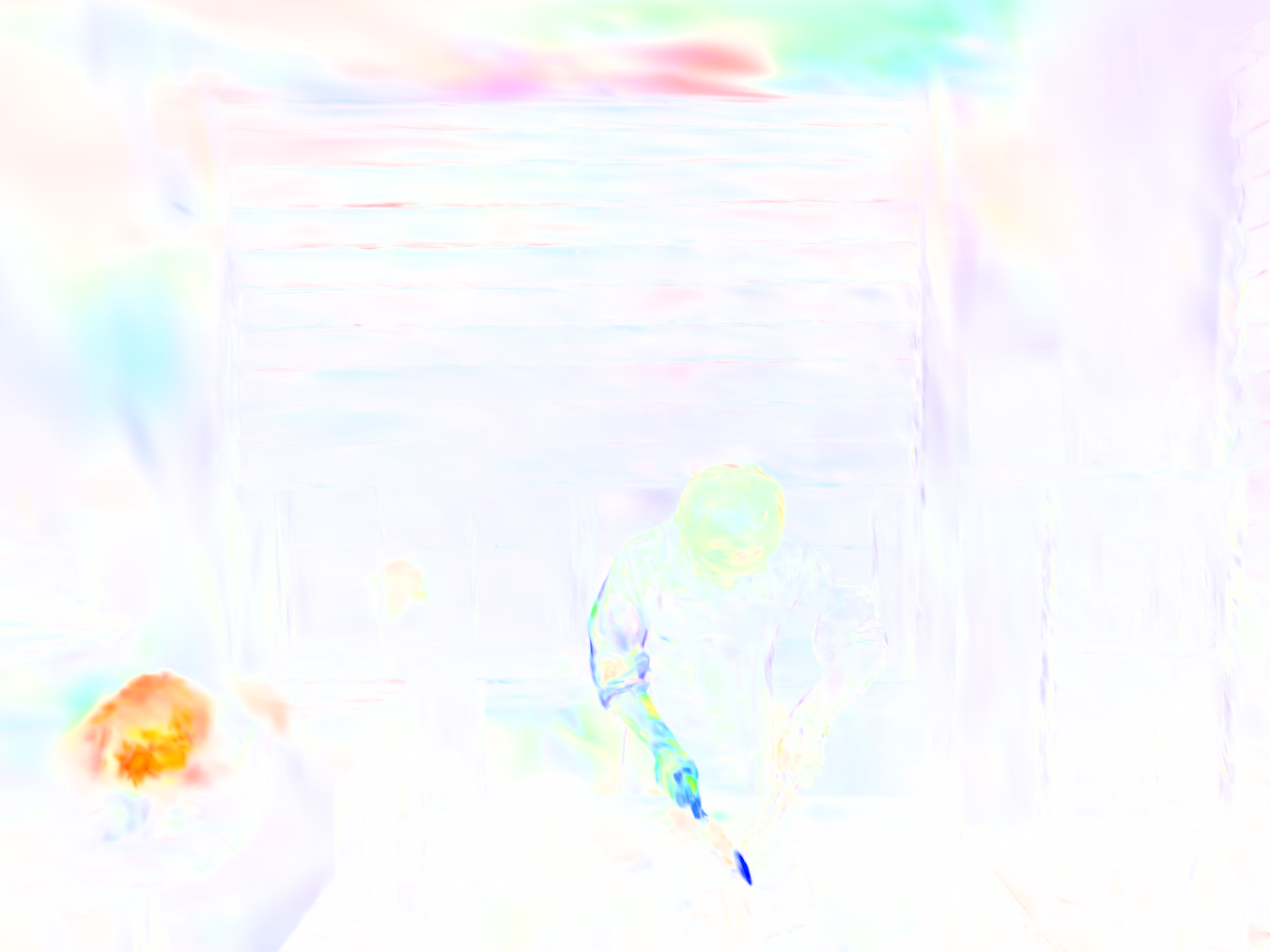}}
    &
    \raisebox{-0.15\height}{\includegraphics[width=.16\textwidth,clip]{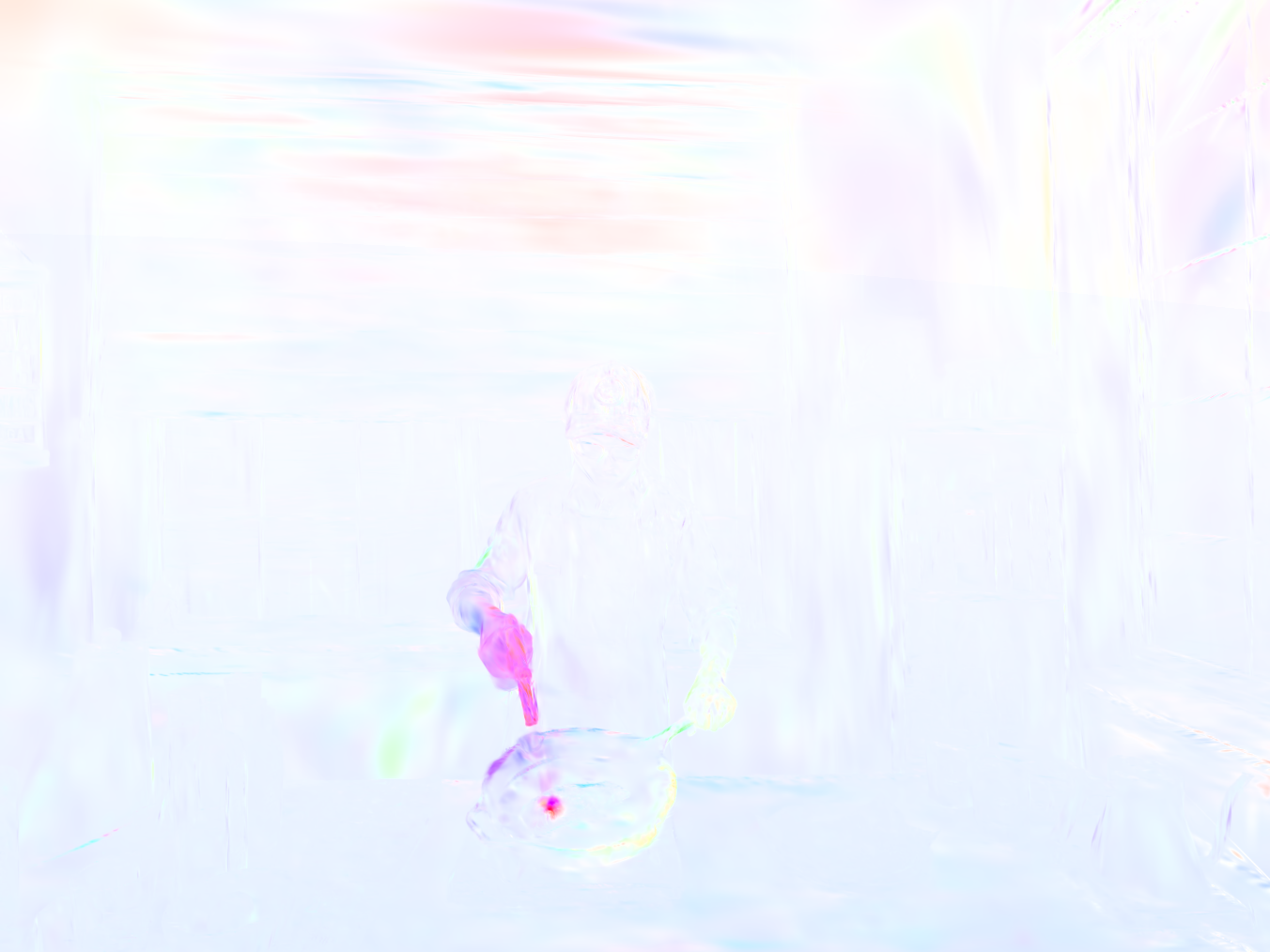}}
    &
    \raisebox{-0.15\height}{\includegraphics[width=.16\textwidth,clip]{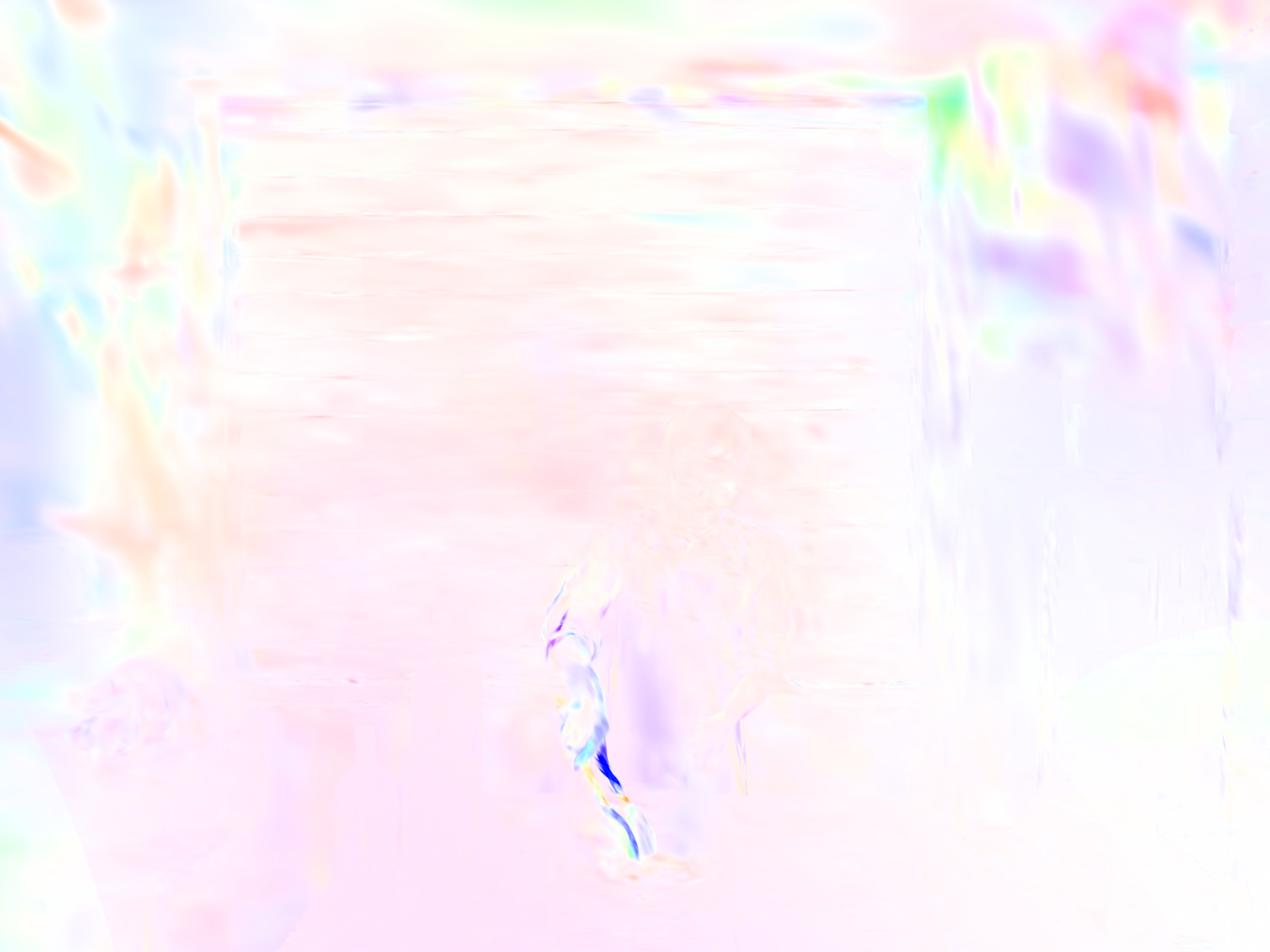}}
    &
    \raisebox{-0.15\height}{\includegraphics[width=.16\textwidth,clip]{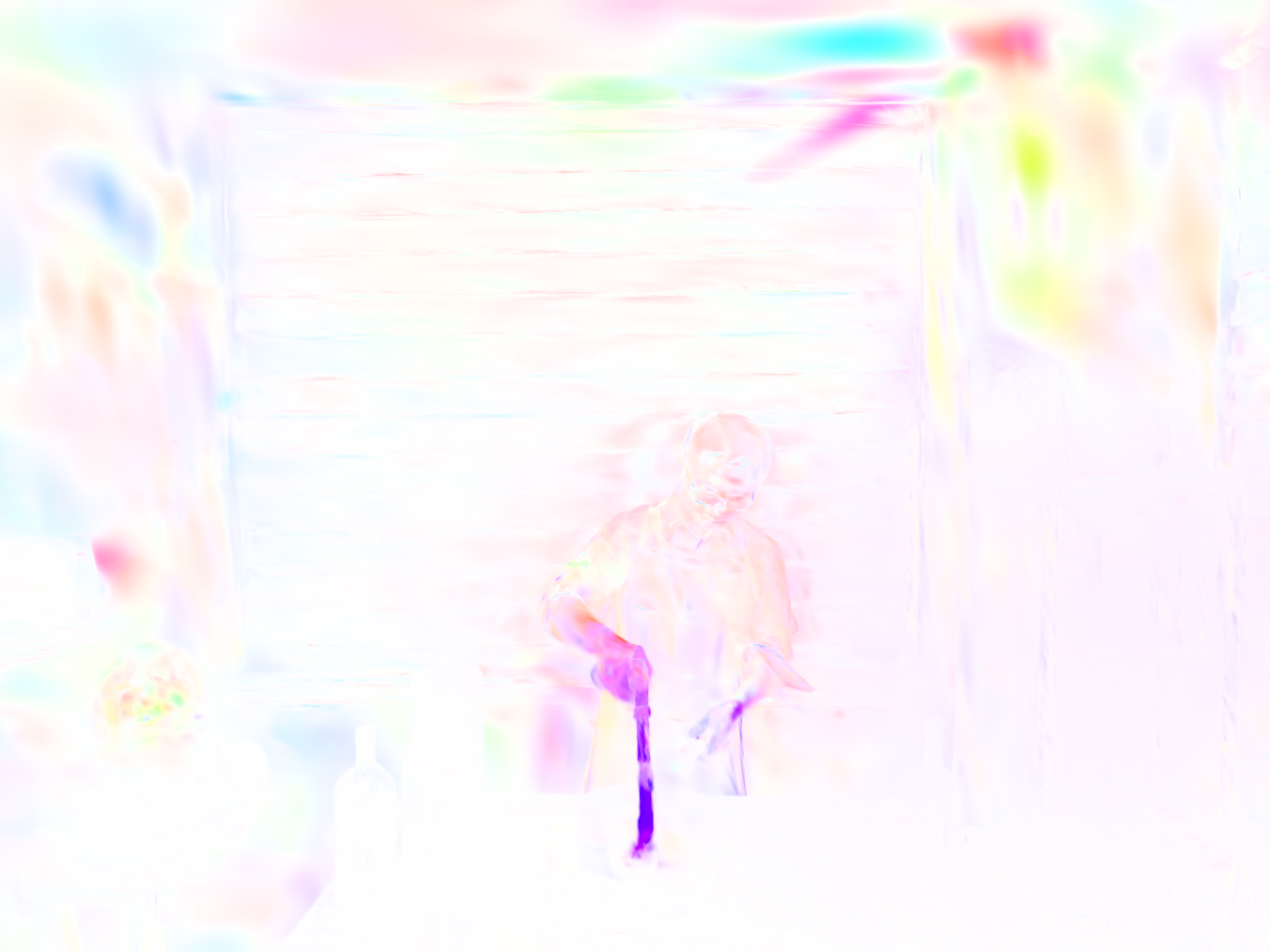}}
 \\
    \rotatebox{90}{\scriptsize{\makecell{GT Flow\\ }}} & \raisebox{-0.2\height}{\includegraphics[width=.16\textwidth,clip]{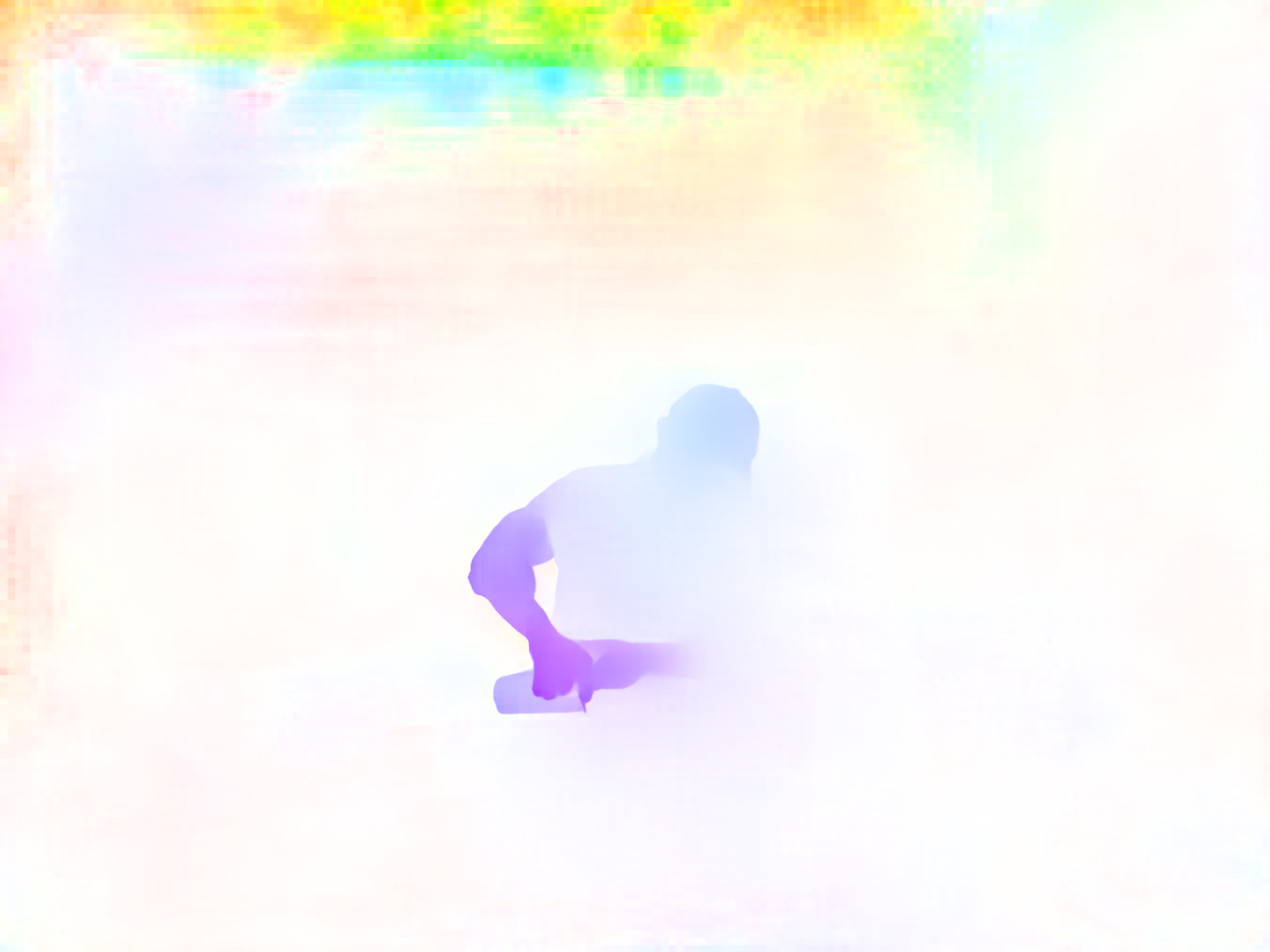}}
    &
    \raisebox{-0.2\height}{\includegraphics[width=.16\textwidth,clip]{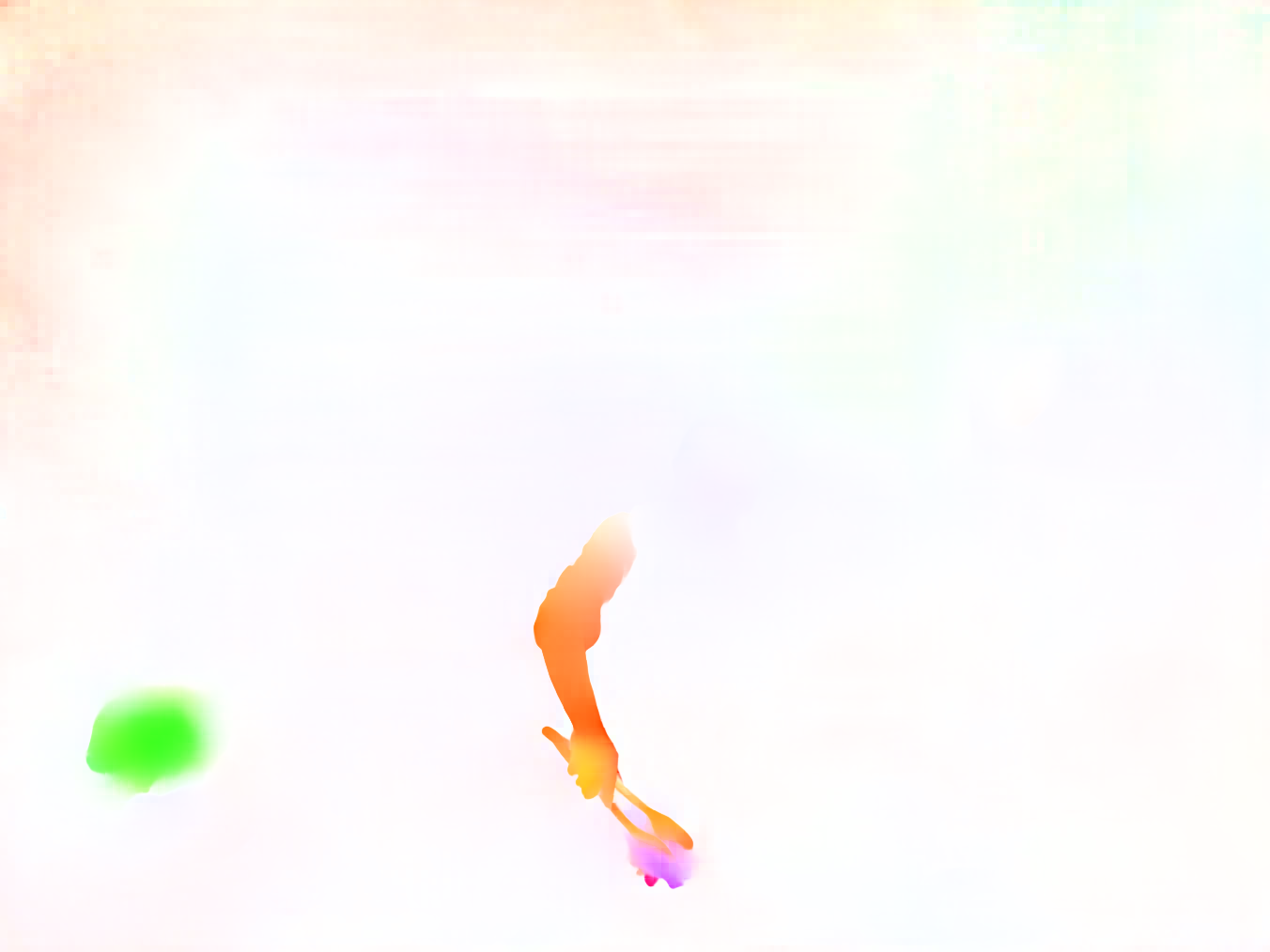}}
    &
    \raisebox{-0.2\height}{\includegraphics[width=.16\textwidth,clip]{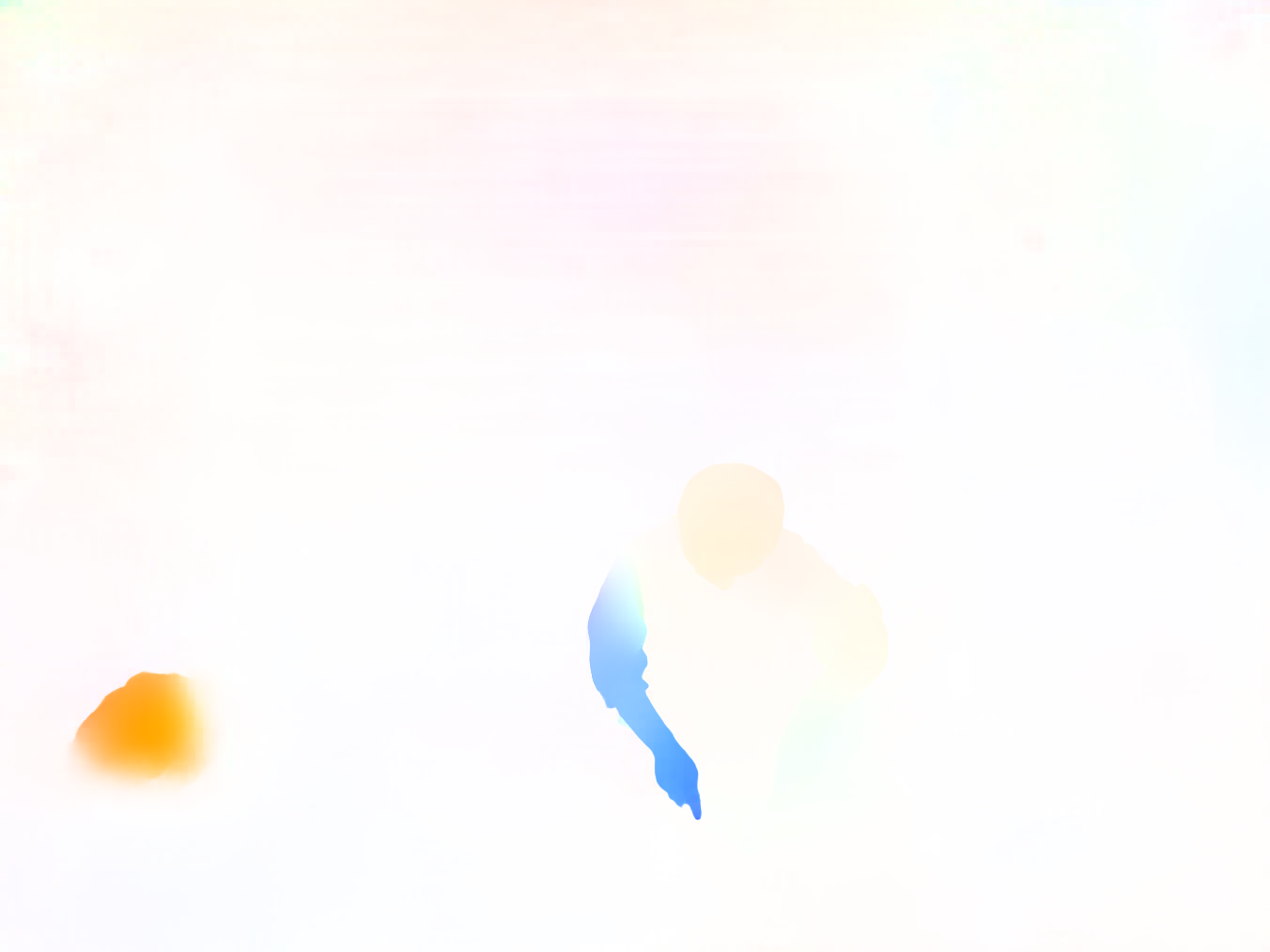}}
    &
    \raisebox{-0.2\height}{\includegraphics[width=.16\textwidth,clip]{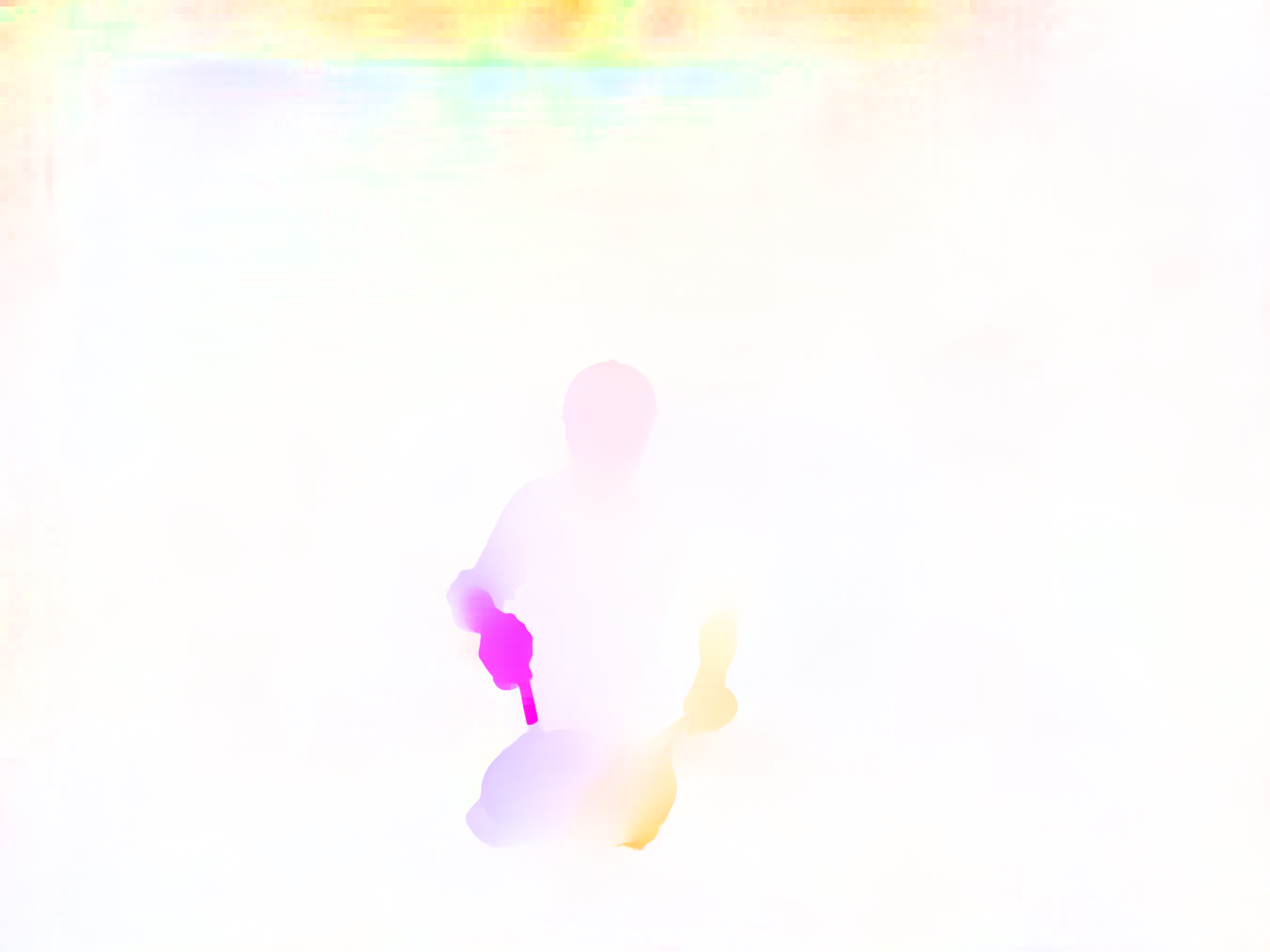}}
    &
    \raisebox{-0.2\height}{\includegraphics[width=.16\textwidth,clip]{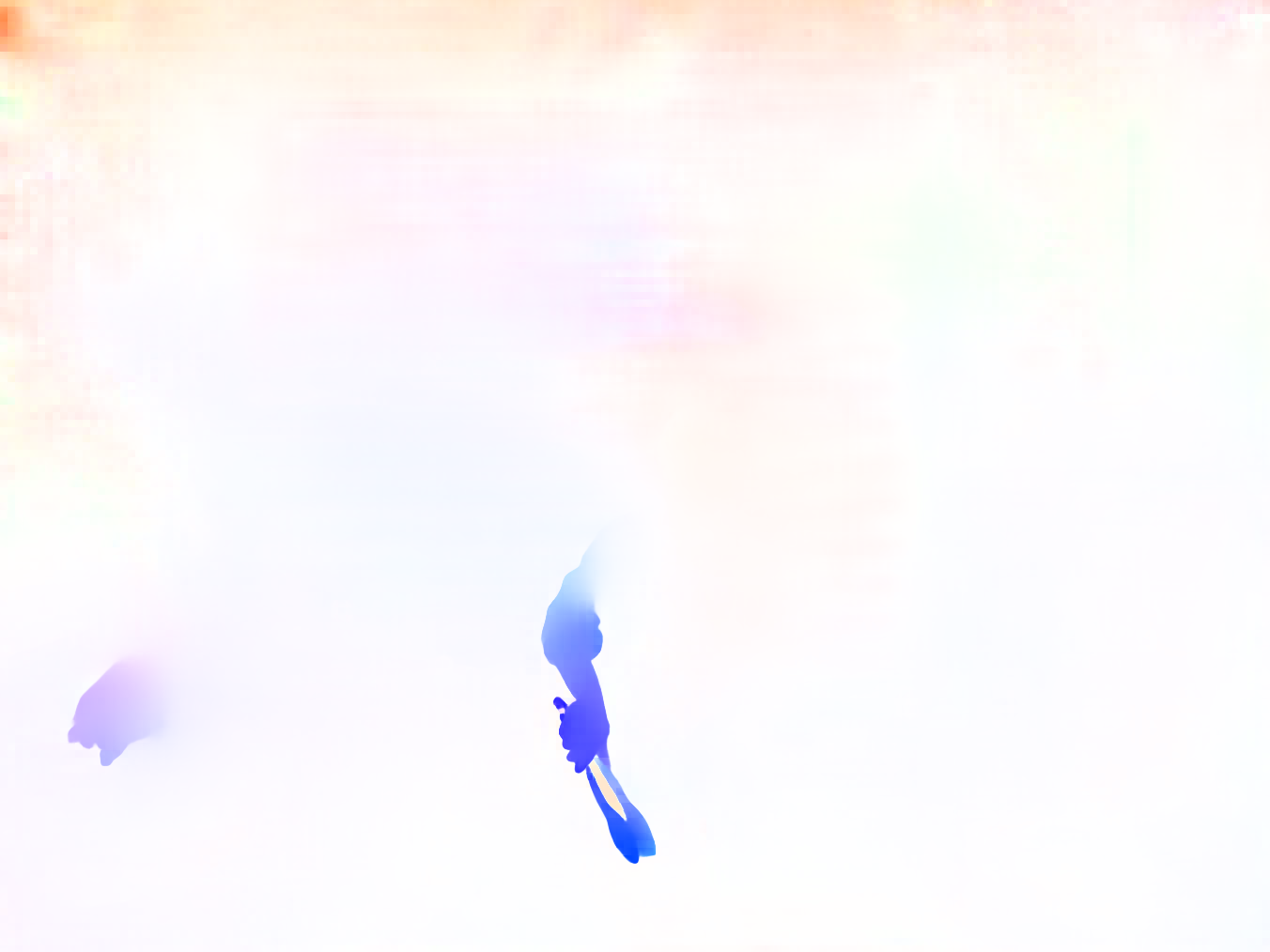}}
    &
    \raisebox{-0.2\height}{\includegraphics[width=.16\textwidth,clip]{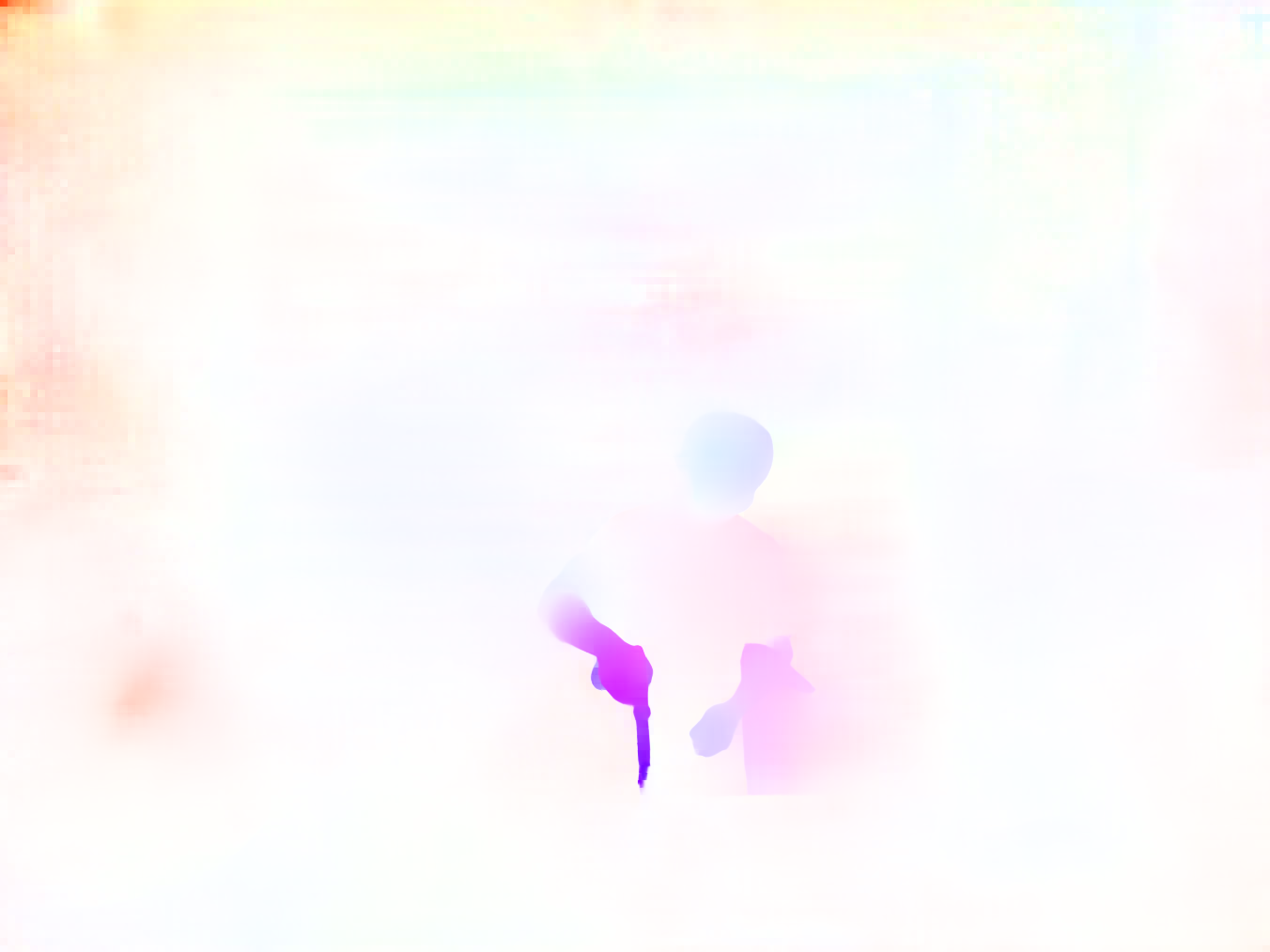}}
 \\
    \rotatebox{90}{\scriptsize{\makecell{Render Image\\ }}}  & \raisebox{-0.1\height}{\includegraphics[width=.16\textwidth,clip]{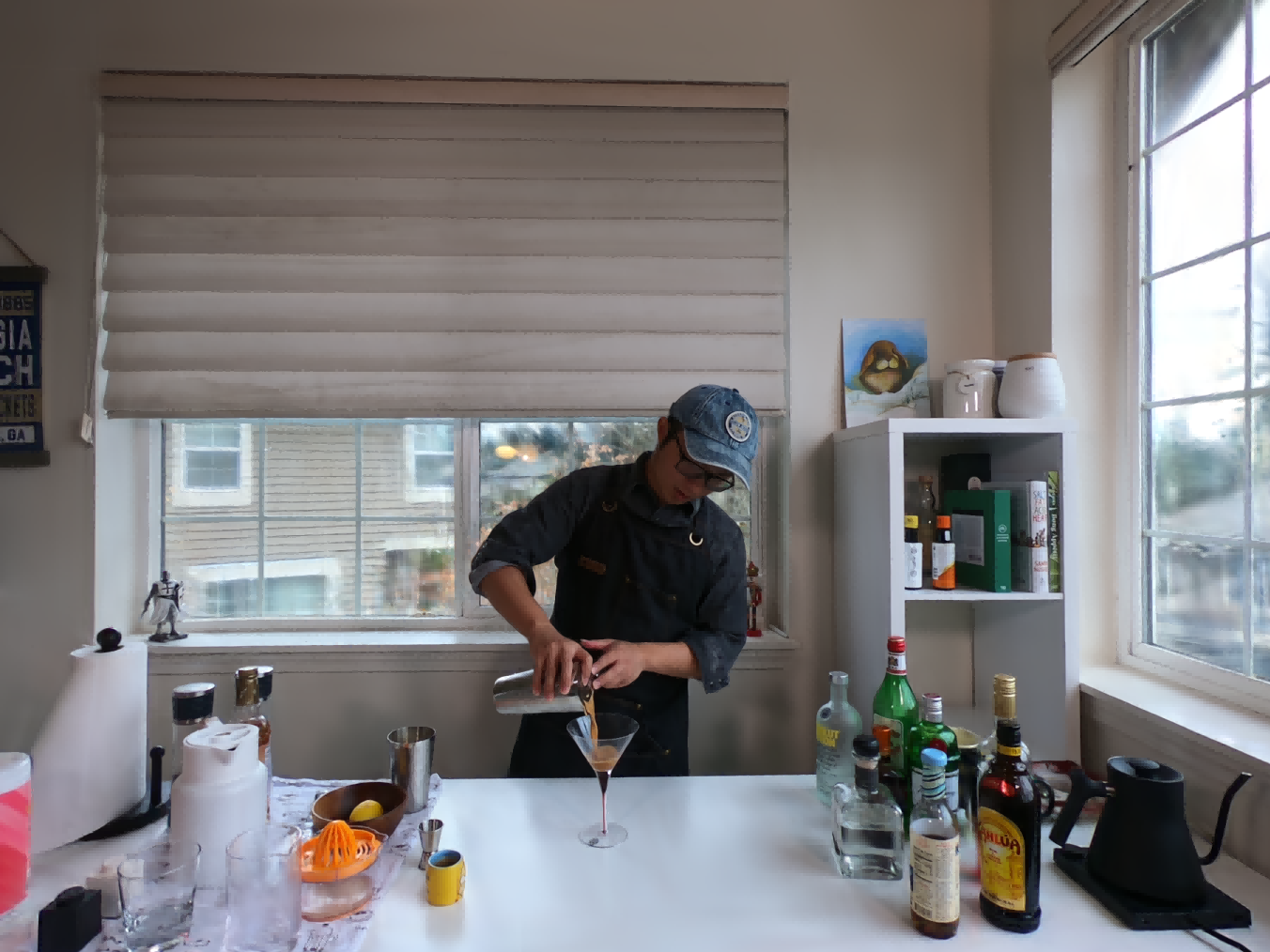}}
    &
    \raisebox{-0.1\height}{\includegraphics[width=.16\textwidth,clip]{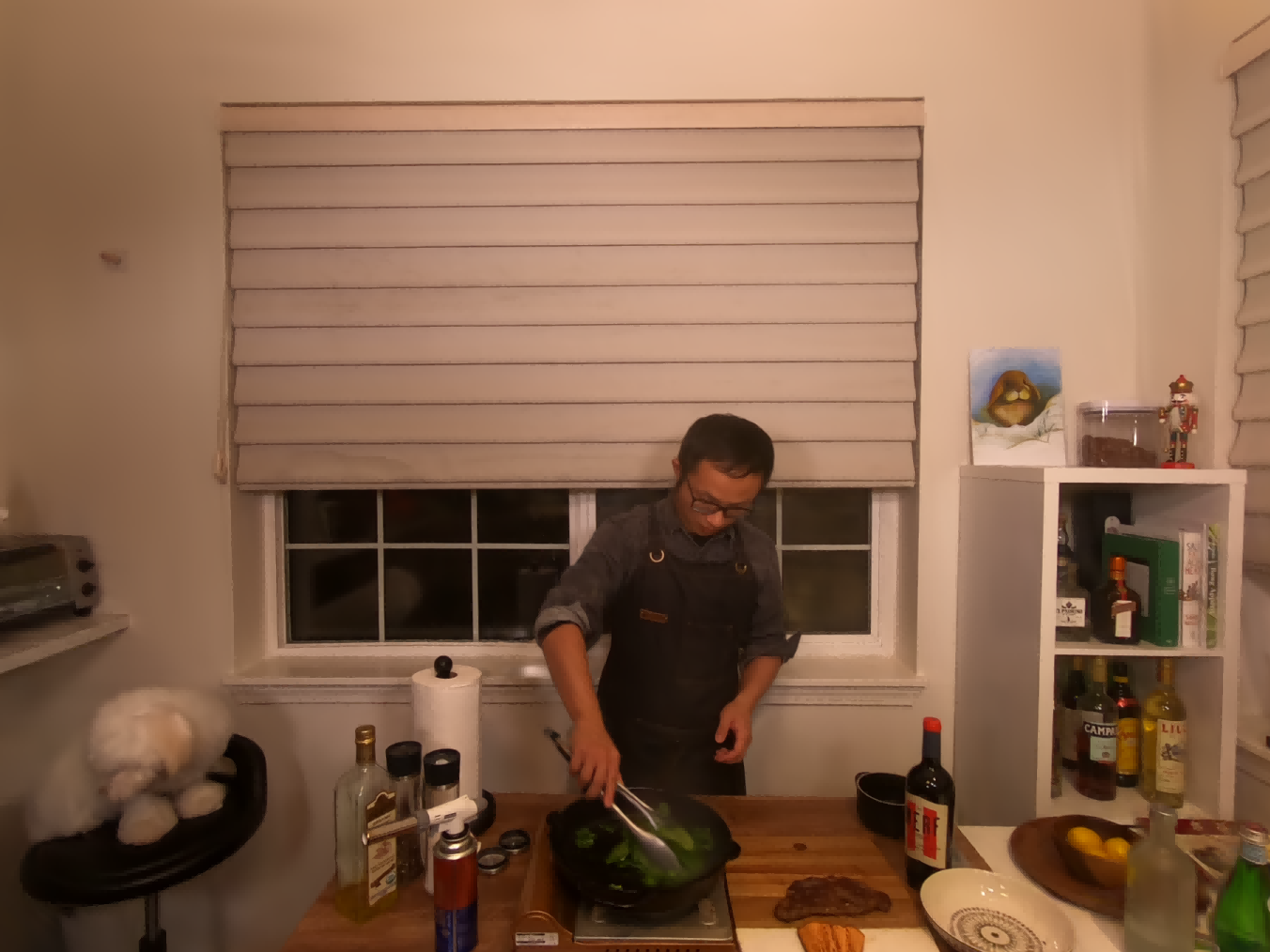}}
    &
    \raisebox{-0.1\height}{\includegraphics[width=.16\textwidth,clip]{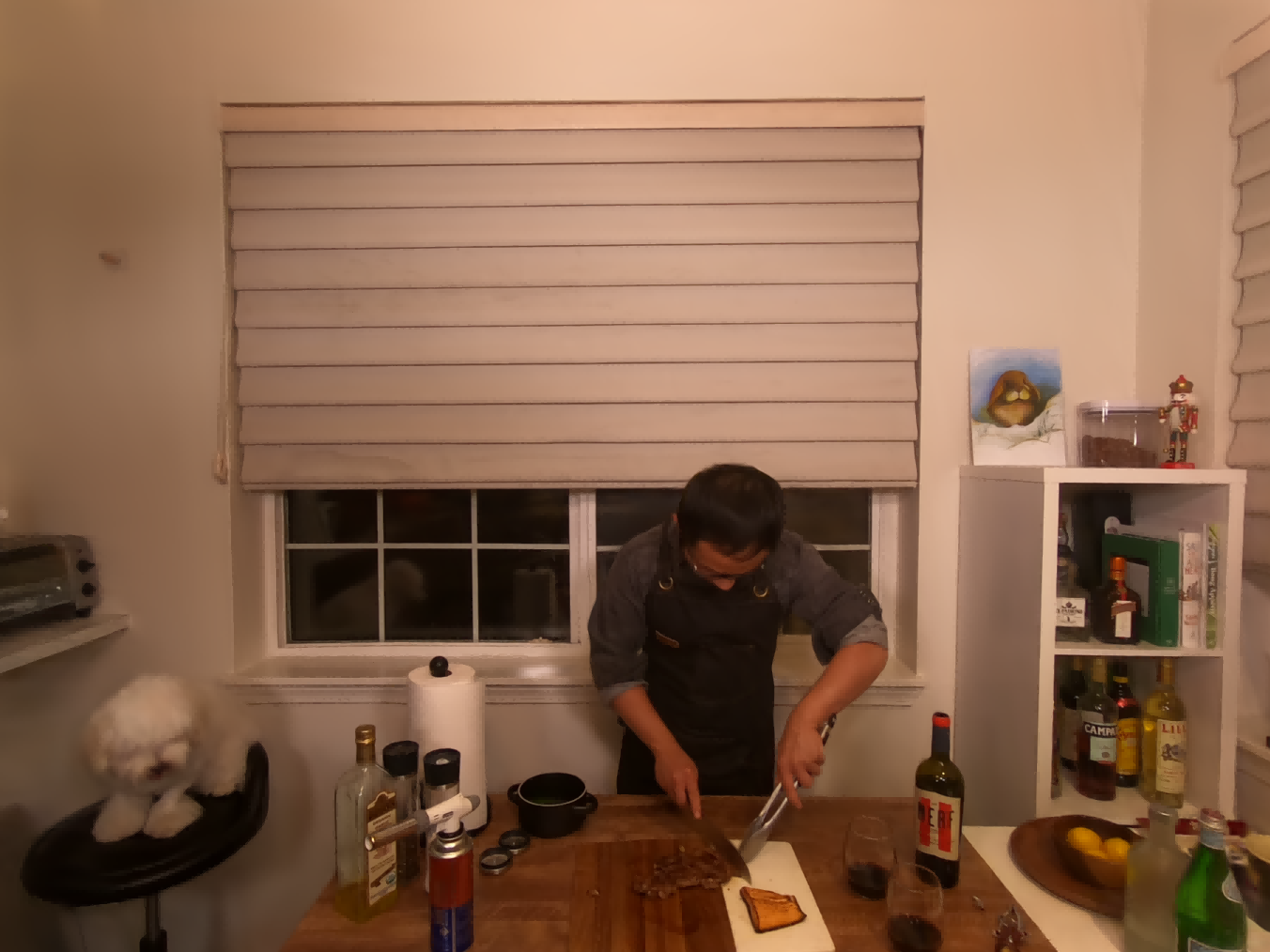}}
    &
    \raisebox{-0.1\height}{\includegraphics[width=.16\textwidth,clip]{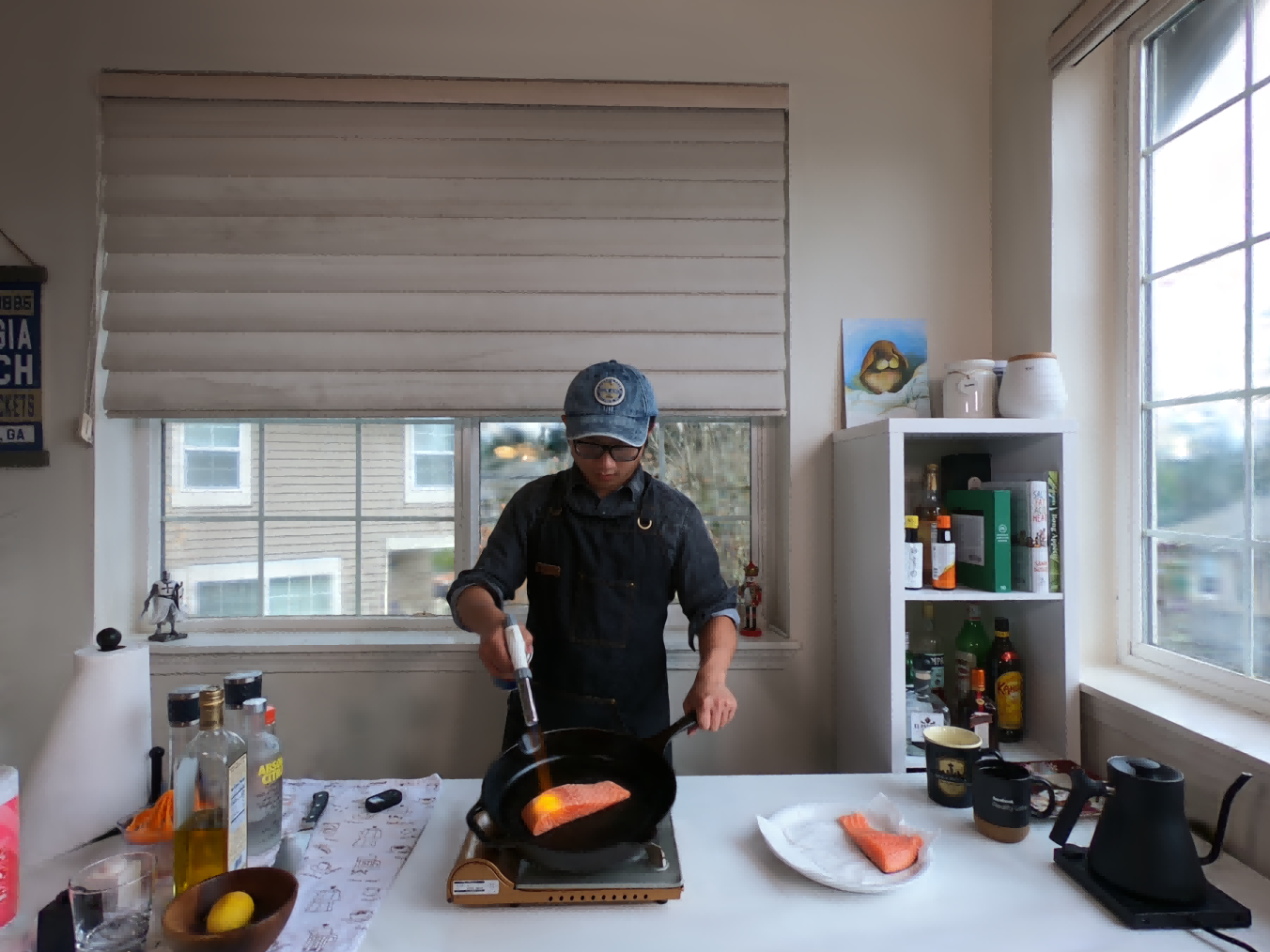}}
    &
    \raisebox{-0.1\height}{\includegraphics[width=.16\textwidth,clip]{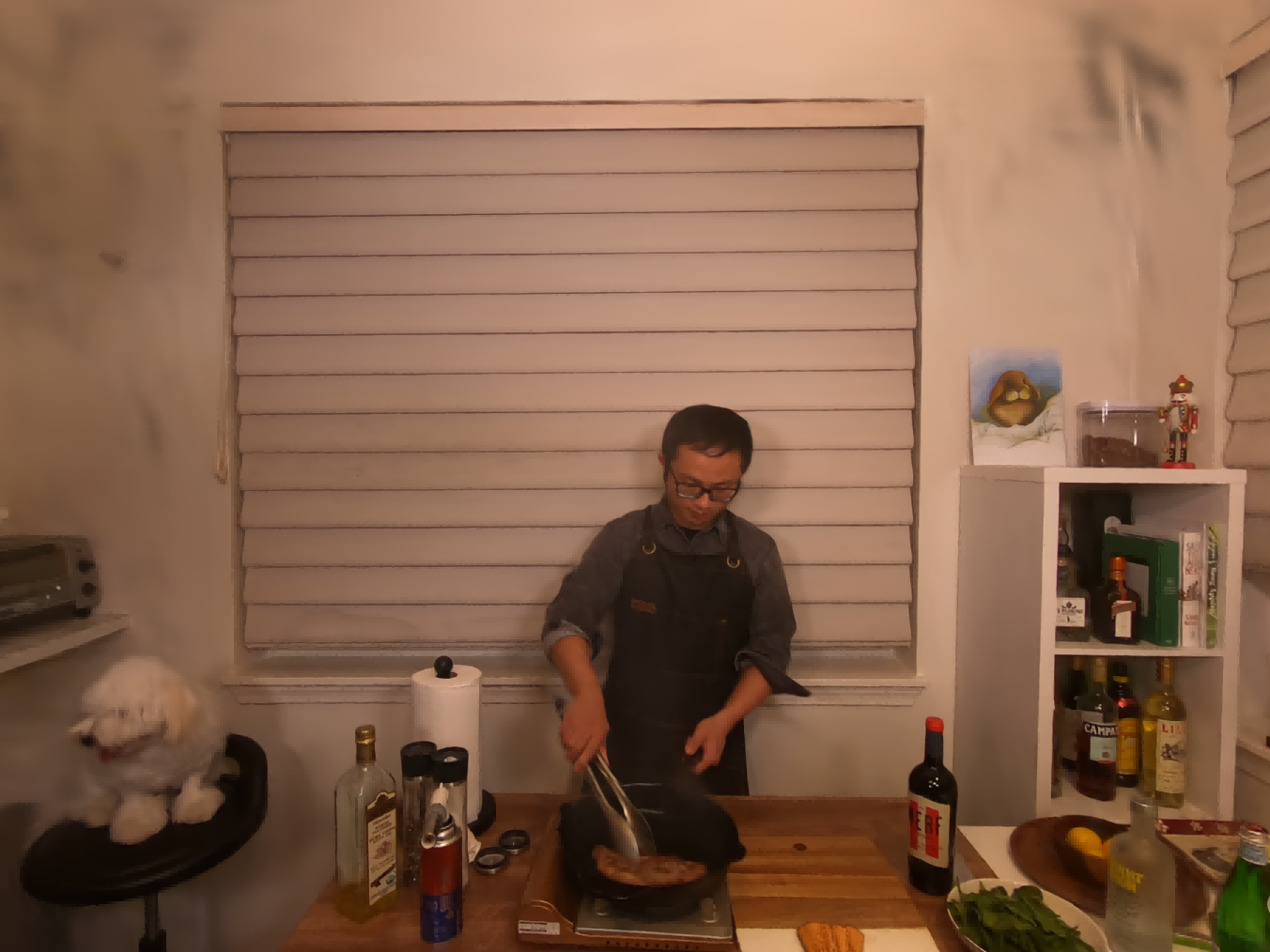}}
    &
    \raisebox{-0.1\height}{\includegraphics[width=.16\textwidth,clip]{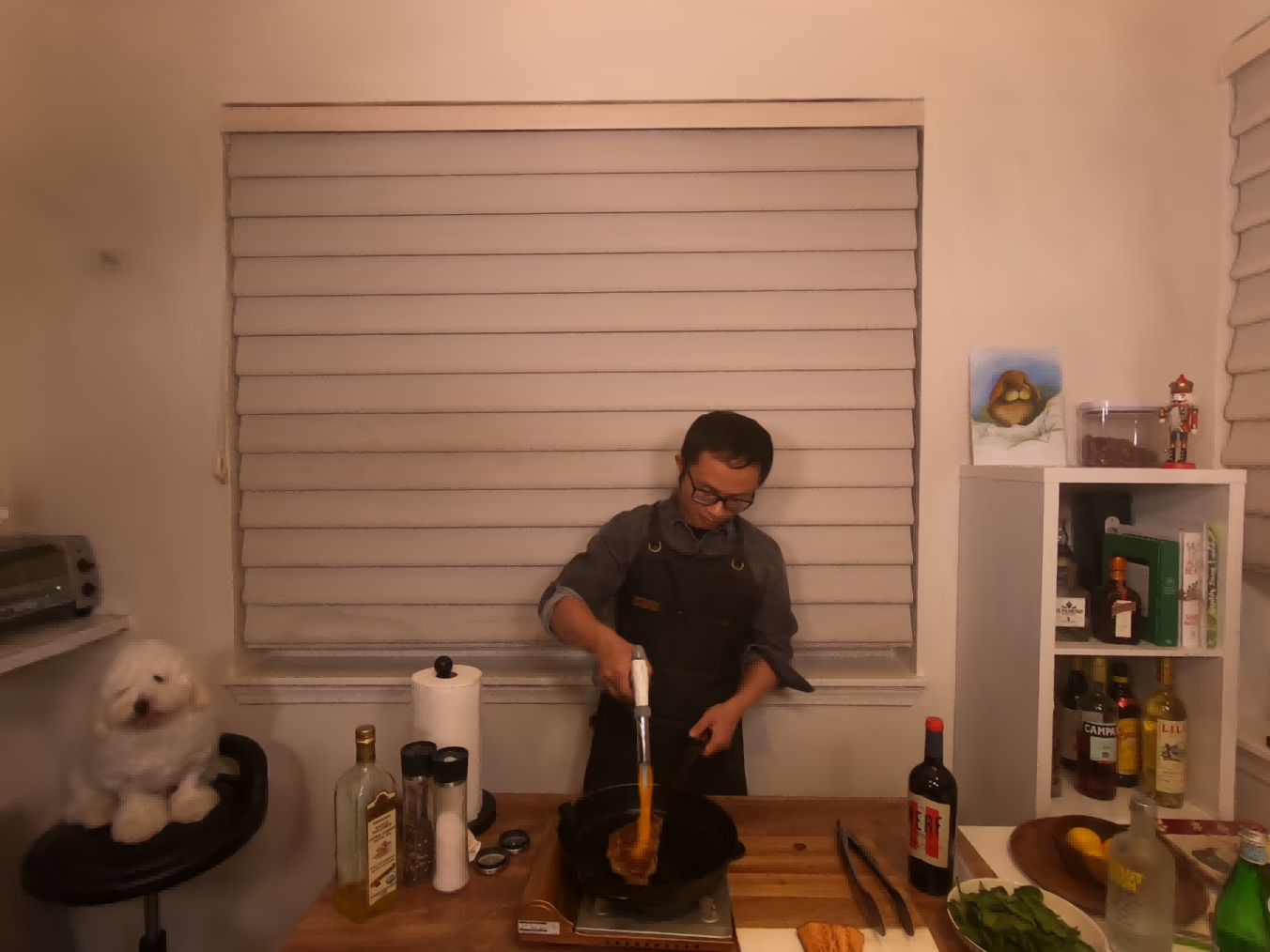}}
 \\
    \rotatebox{90}{\scriptsize{\makecell{GT Image\\ }}} & \raisebox{-0.17\height}{\includegraphics[width=.16\textwidth,clip]{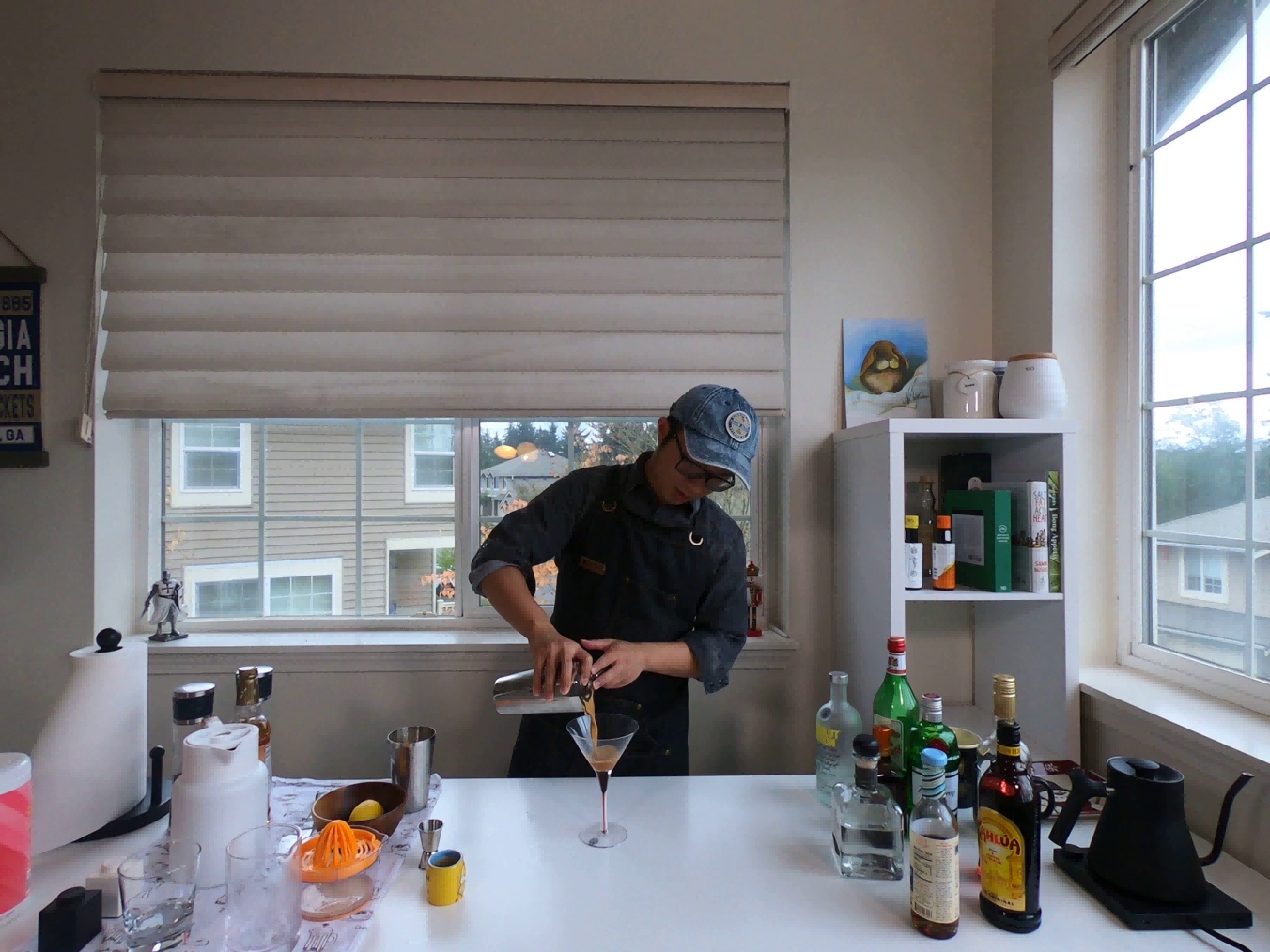}}
    &
    \raisebox{-0.17\height}{\includegraphics[width=.16\textwidth,clip]{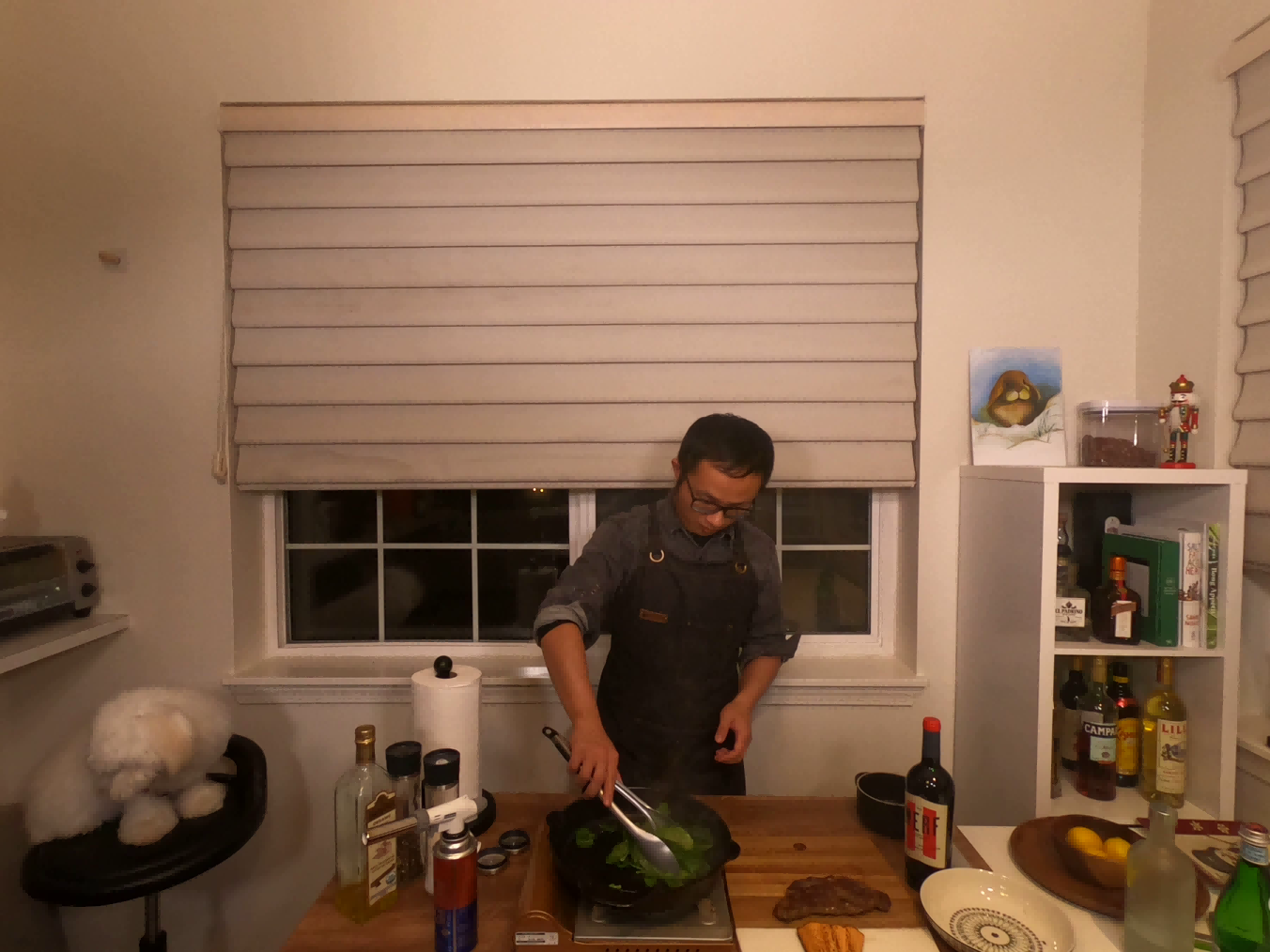}}
    &
    \raisebox{-0.17\height}{\includegraphics[width=.16\textwidth,clip]{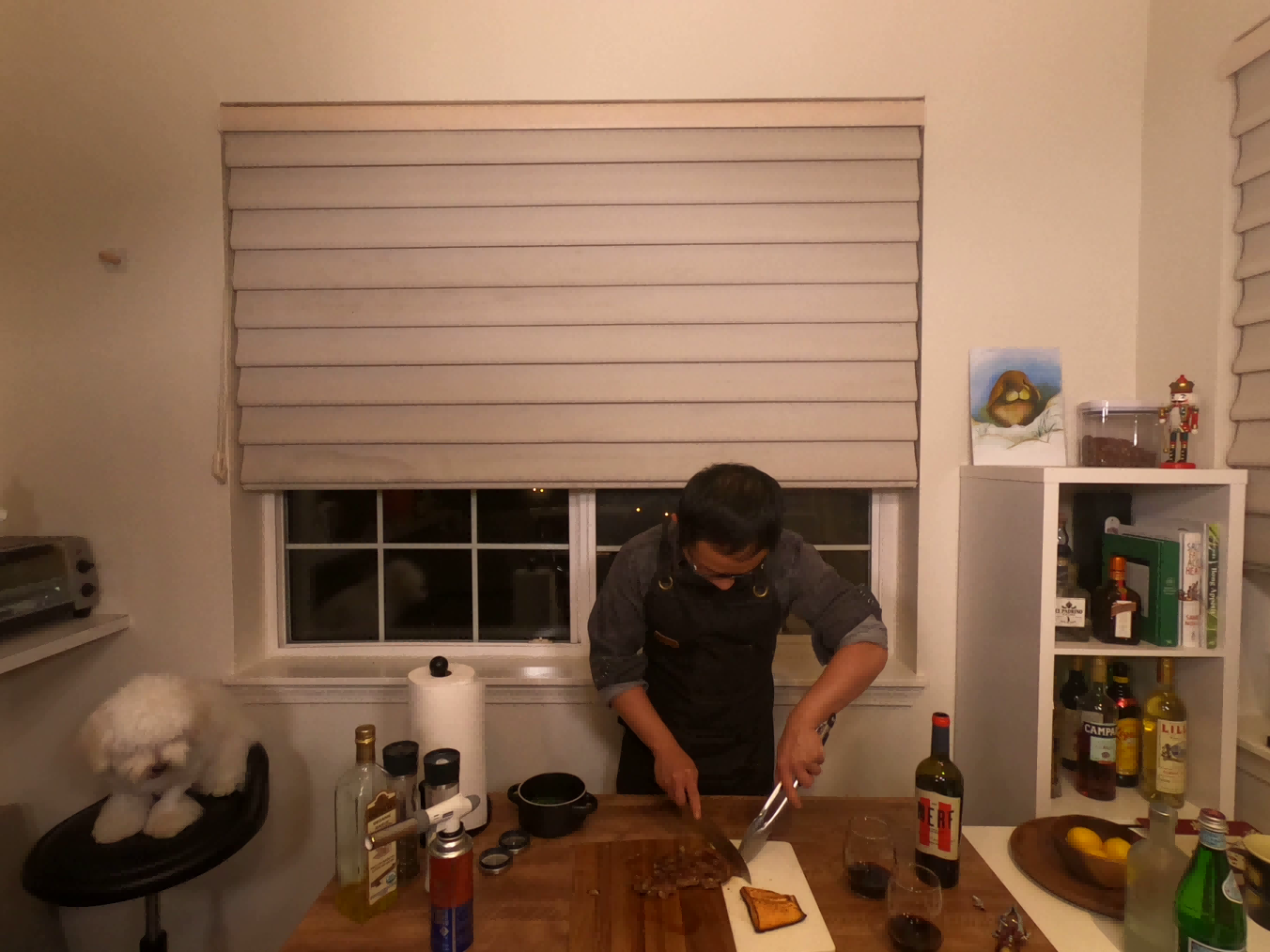}}
    &
    \raisebox{-0.17\height}{\includegraphics[width=.16\textwidth,clip]{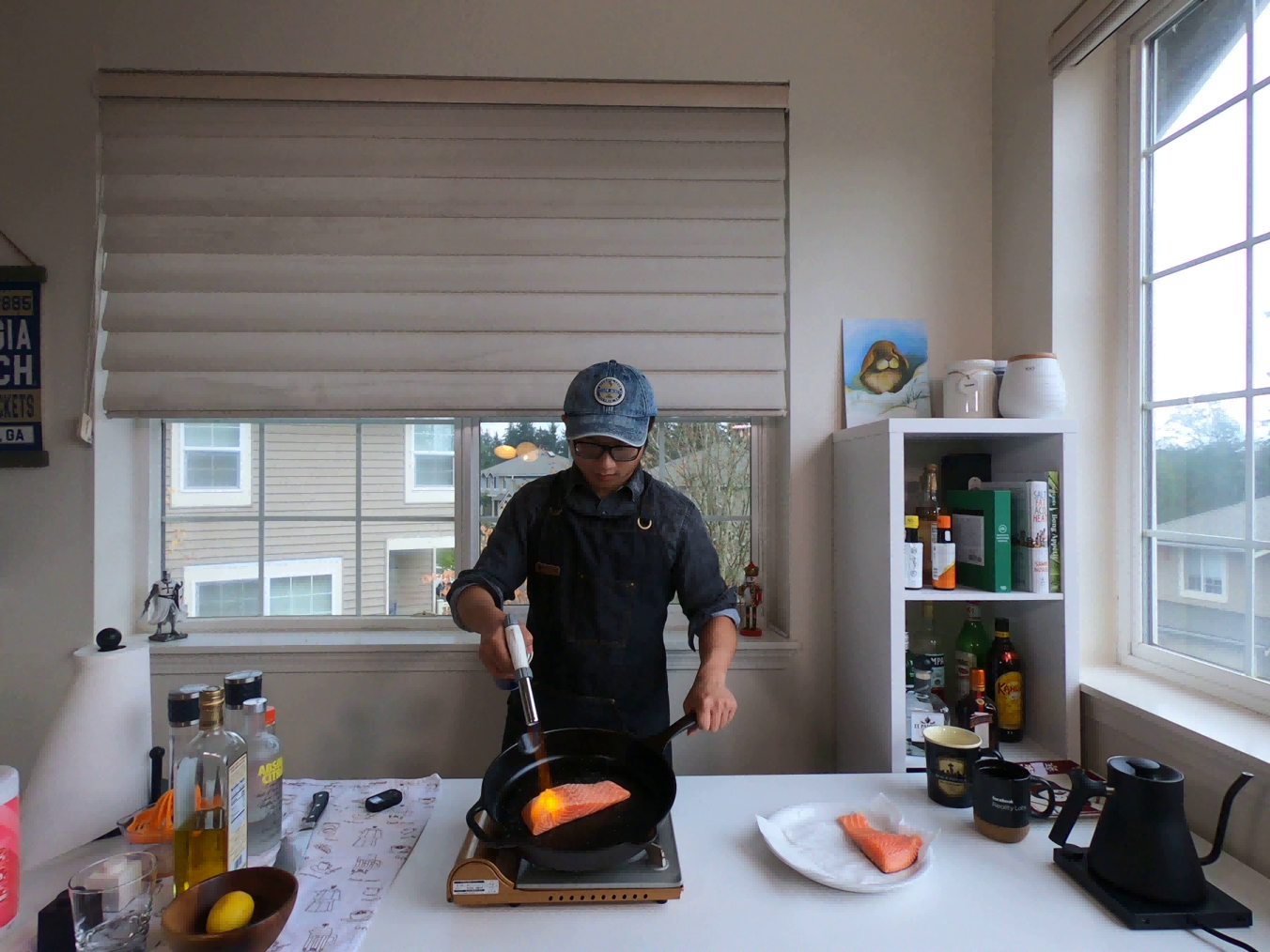}}
    &
    \raisebox{-0.17\height}{\includegraphics[width=.16\textwidth,clip]{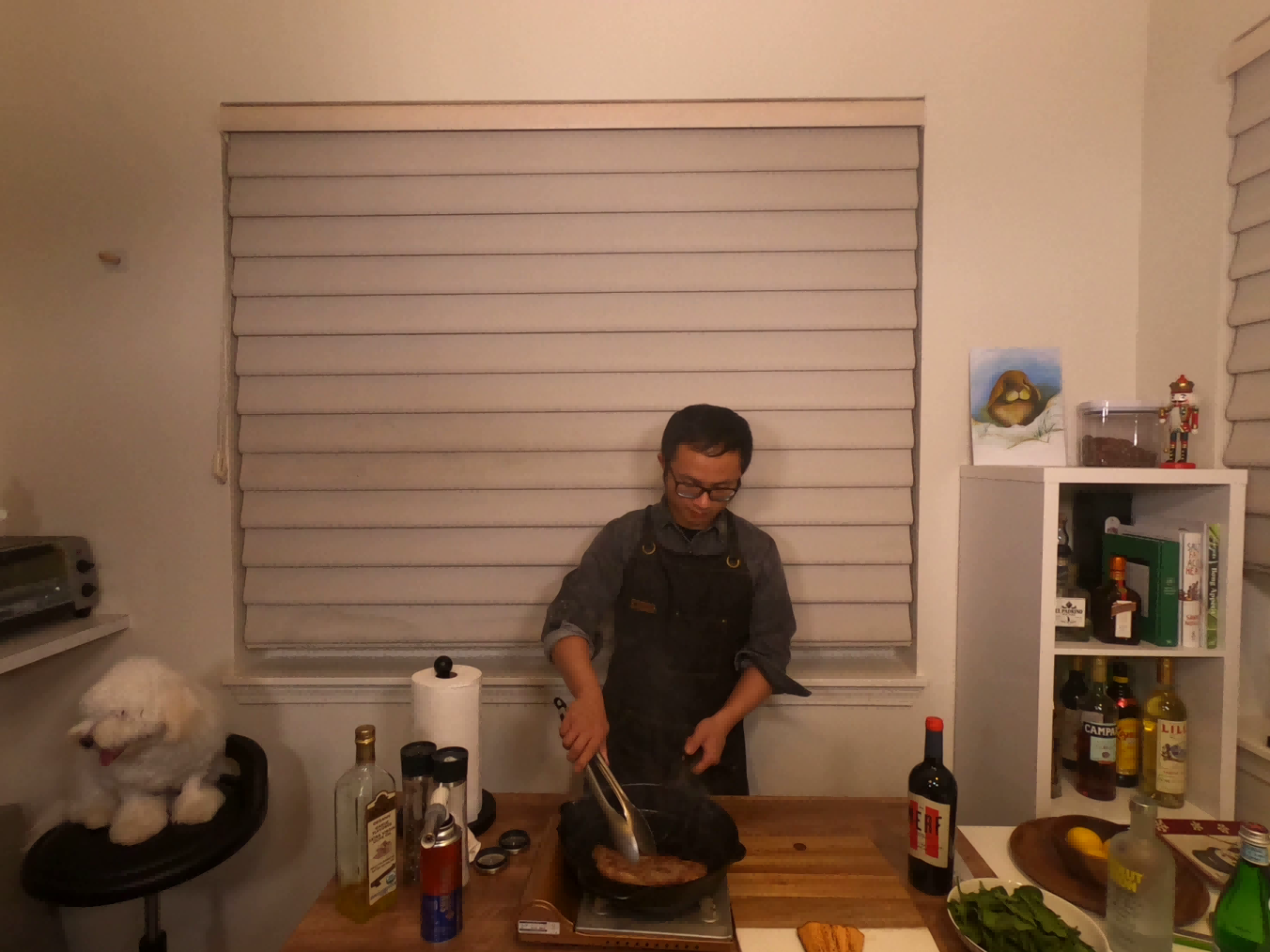}}
    &
    \raisebox{-0.17\height}{\includegraphics[width=.16\textwidth,clip]{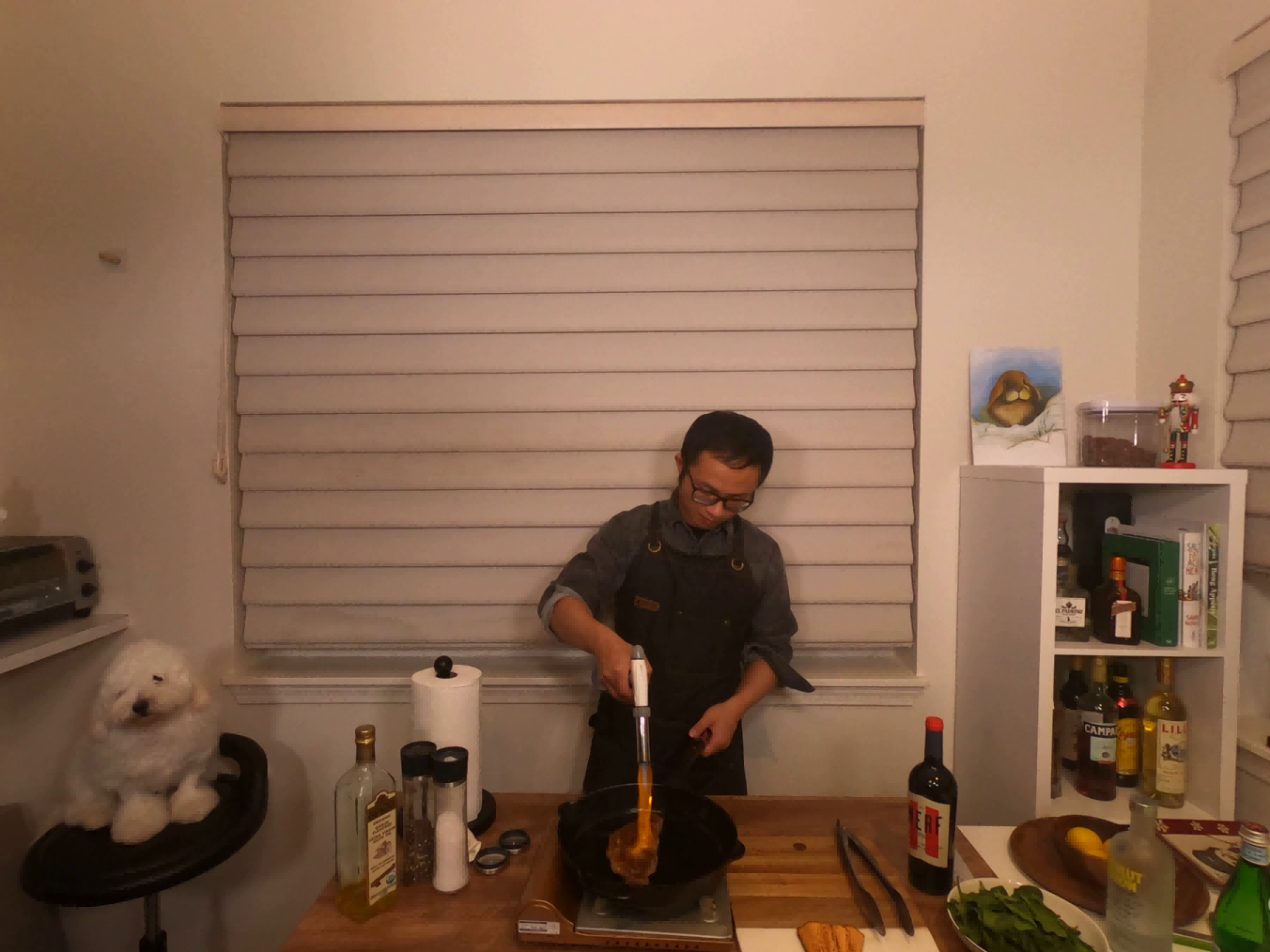}}
 \\
    \end{tabular}
    \caption{\textbf{Emerged dynamics from our 4D Gaussian.} The displayed views are selected from the test view of the Plenoptic Video dataset. The ground truth optical flows are extracted by VideoFlow~\cite{shi2023videoflow} for reference. }
\label{fig:flow}
\end{figure*} Incorporation of 4D rotation to our 4D Gaussian equips it with the ability to model the motion.
Note that 4D rotation can result in a 3D displacement. 
To assess this scene motion capture ability, we conduct a thorough evaluation.
For each Gaussian, we test the trajectory in space formed by 
the expectation of its conditional distribution $\mu_{xyz|t}$.
Then, we project its 3D displacement between consecutive two frames to the image plane and render it using equation~\eqref{eq:4dblending} as the estimated optical flow.
In Figure~\ref{fig:flow}, we select one frame from each scene in the Plenoptic Video dataset to exhibit the rendered optical flow.
The result reveals that without explicit motion supervision or regularization, 
optimizing the rendering loss alone can lead to the emergence of coarse scene dynamics.

\noindent \textbf{The temporal characteristic of 4D Gaussians}
\begin{figure*}[t]
    \centering 
    \setlength{\tabcolsep}{0.0pt}
    \begin{tabular}{cccccc} 
    \rotatebox{90}{\small{\makecell{Ref Image\\ }}}  & \raisebox{-0.20\height}{\includegraphics[width=.19\textwidth,clip]{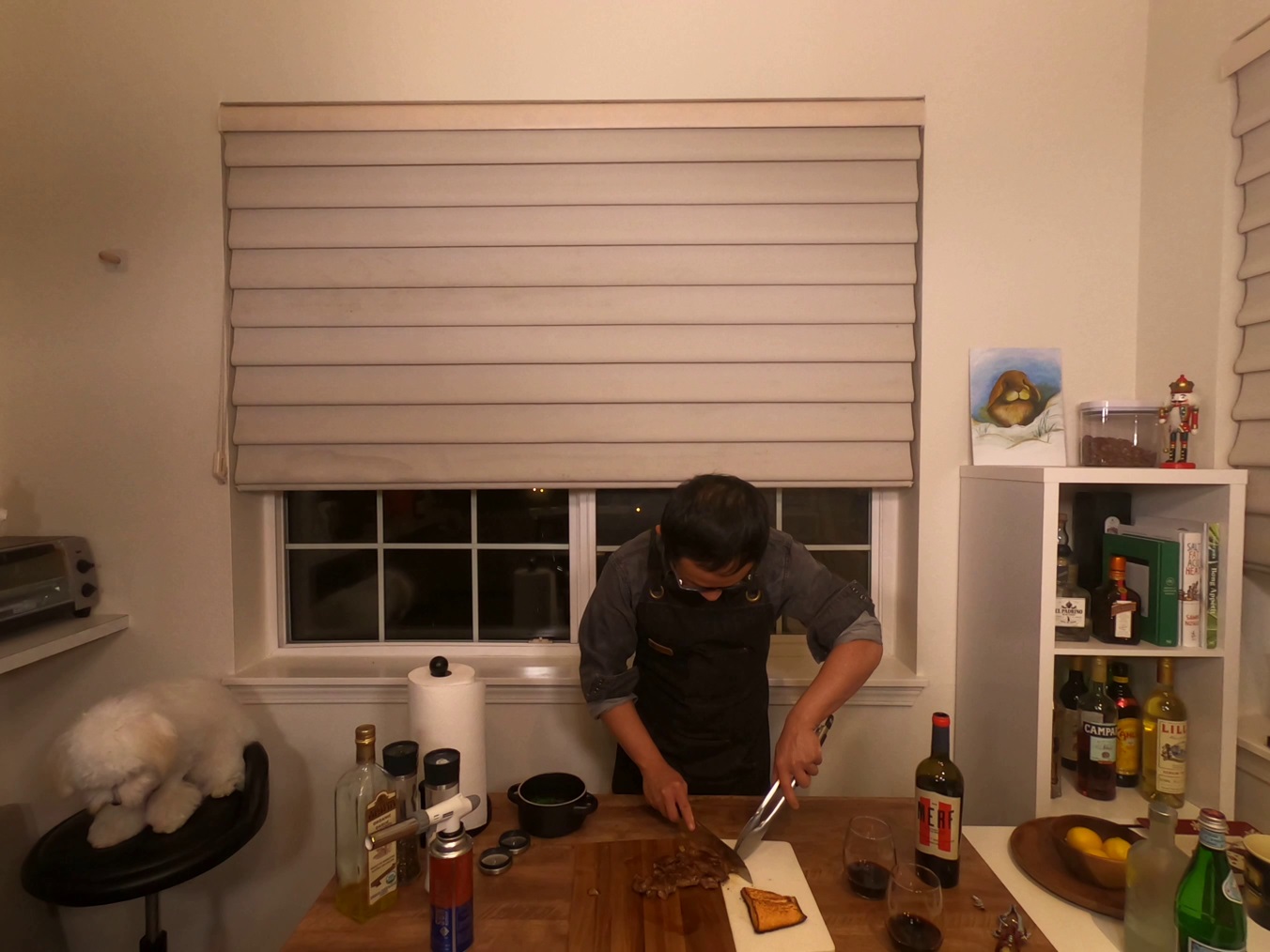}}
    &
    \raisebox{-0.20\height}{\includegraphics[width=.19\textwidth,clip]{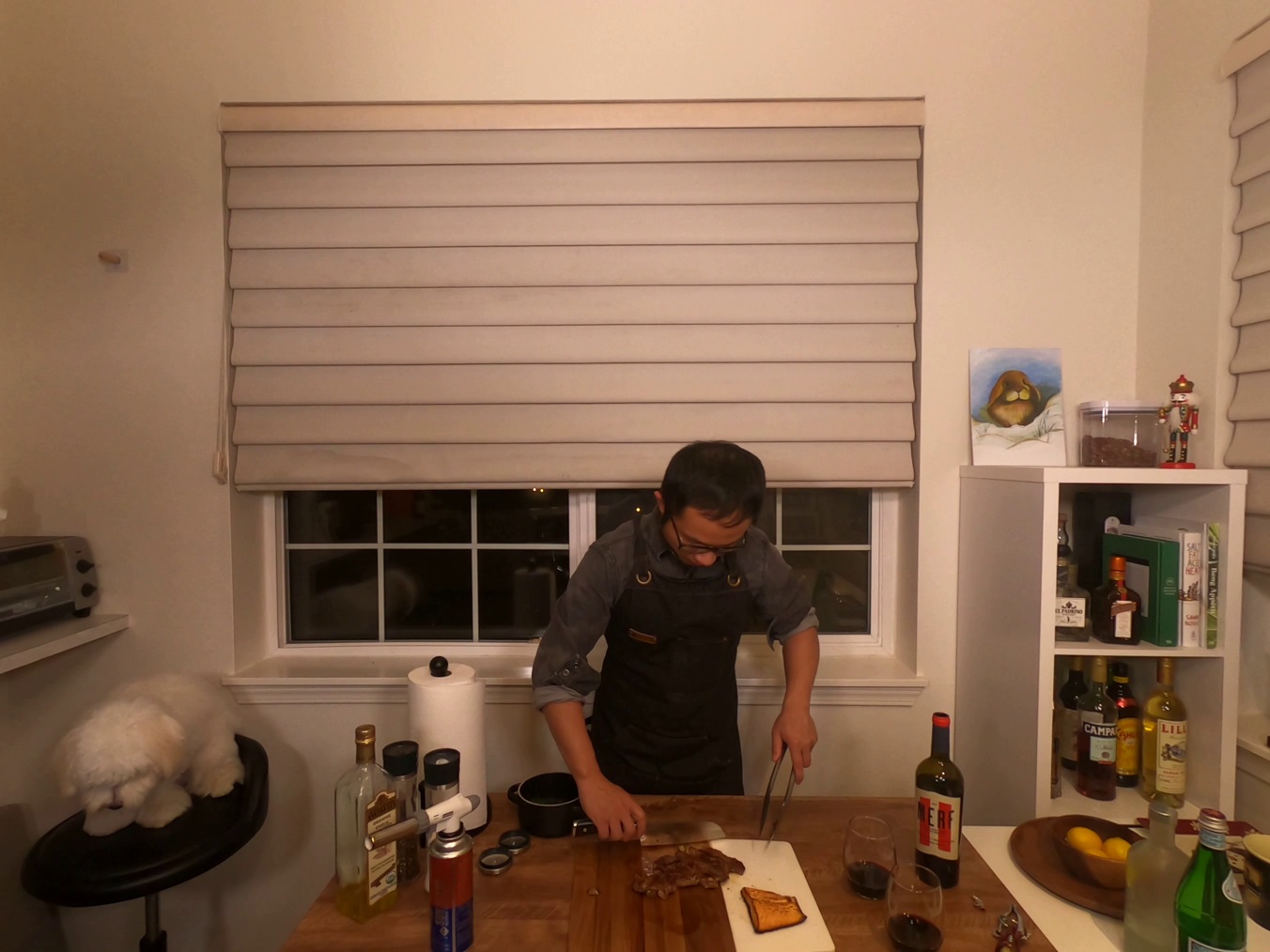}}
    &
    \raisebox{-0.20\height}{\includegraphics[width=.19\textwidth,clip]{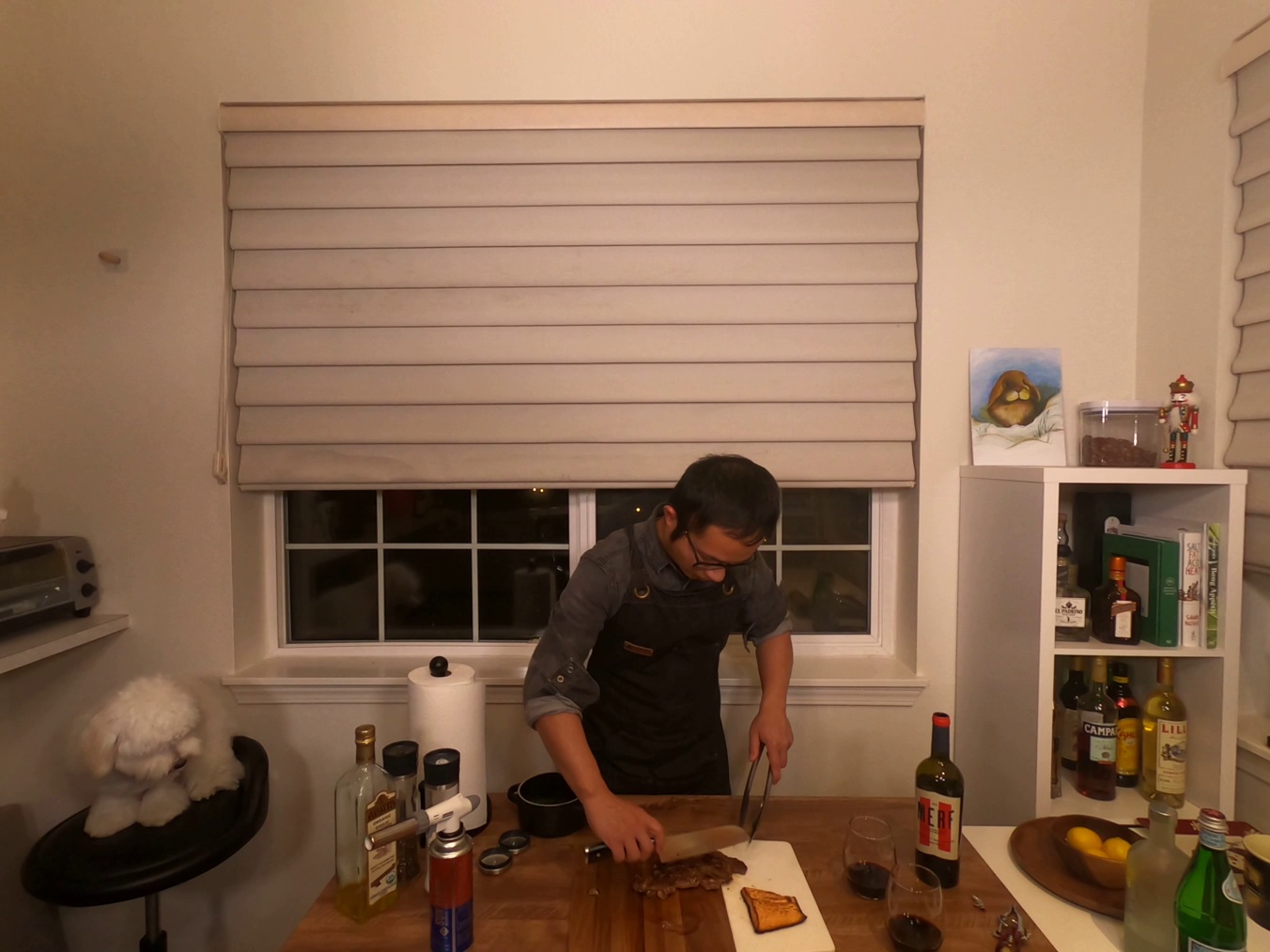}}
    &
    \raisebox{-0.20\height}{\includegraphics[width=.19\textwidth,clip]{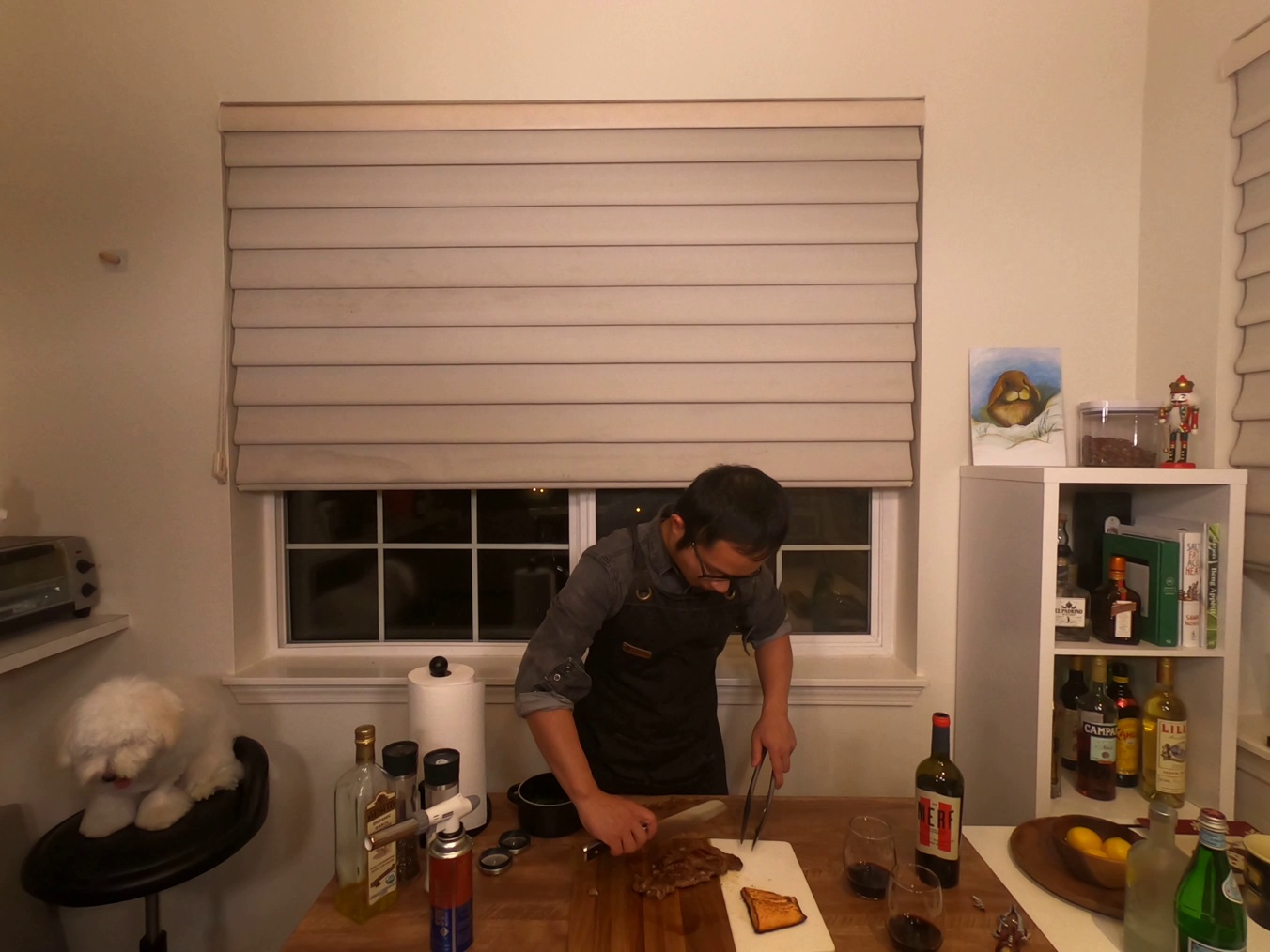}}
    &
    \raisebox{-0.20\height}{\includegraphics[width=.19\textwidth,clip]{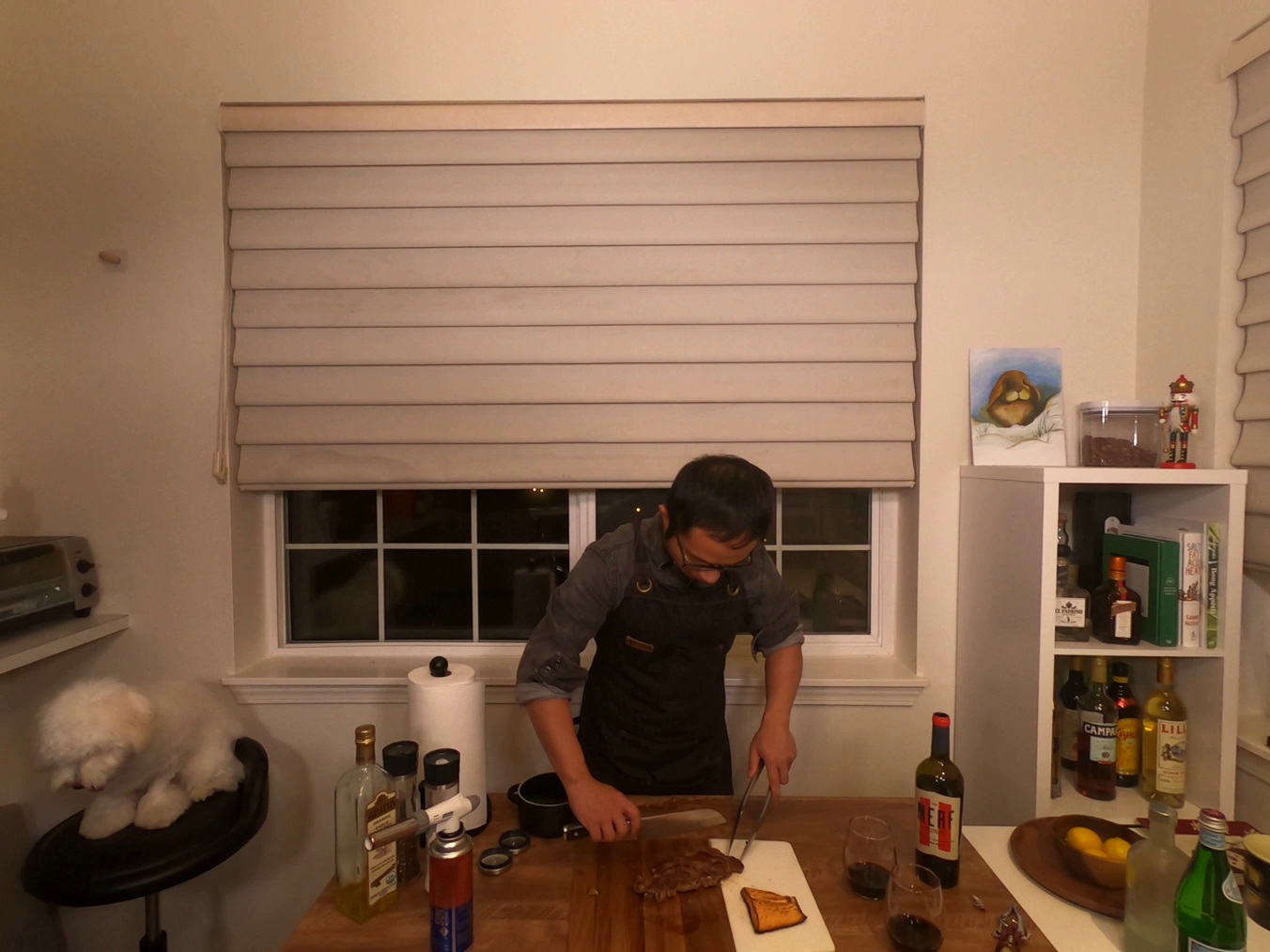}}
 \\
    \rotatebox{90}{\small{\makecell{Mean\\ }}} & \raisebox{-0.20\height}{\includegraphics[width=.19\textwidth,clip]{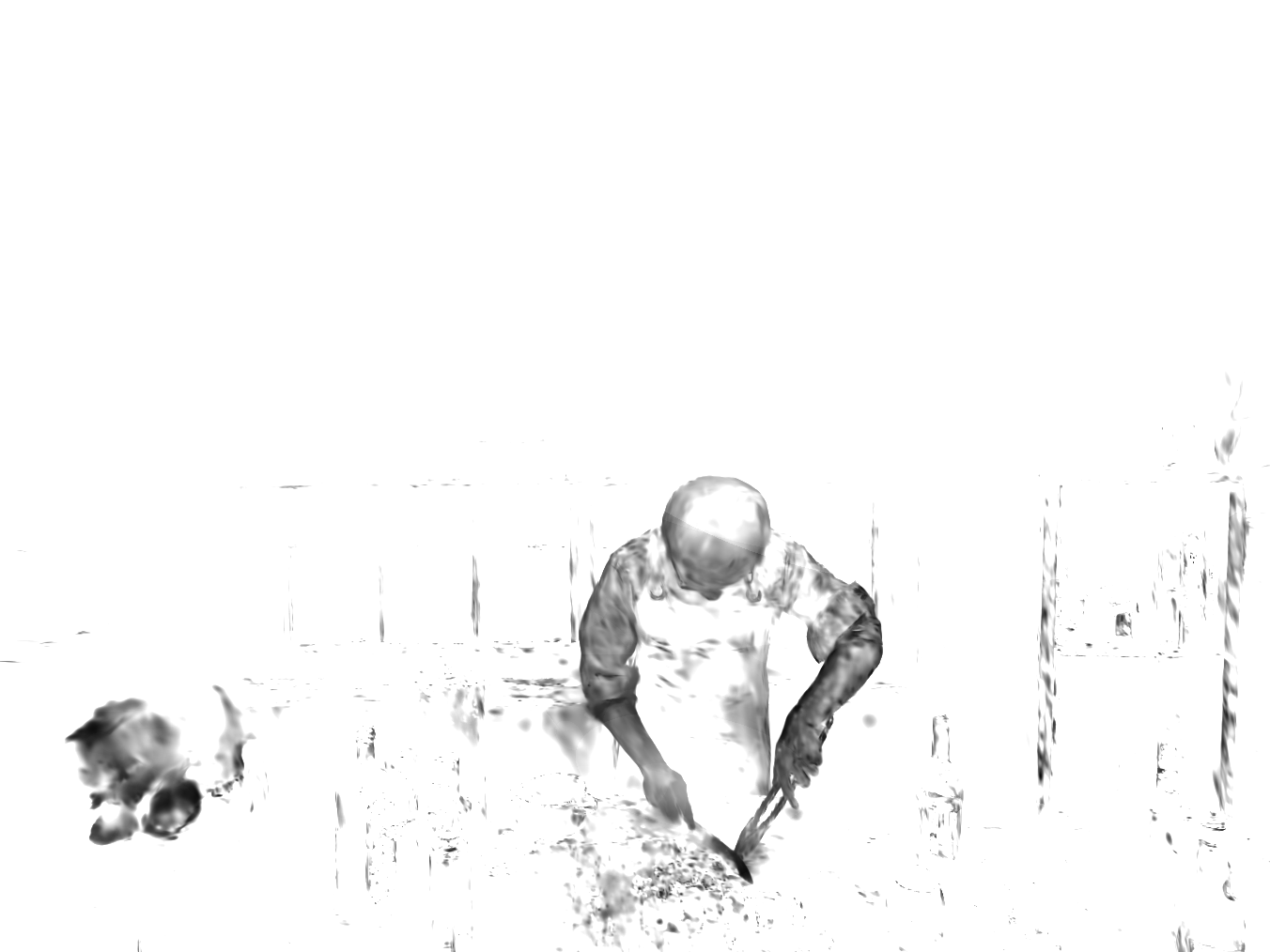}}
    &
    \raisebox{-0.20\height}{\includegraphics[width=.19\textwidth,clip]{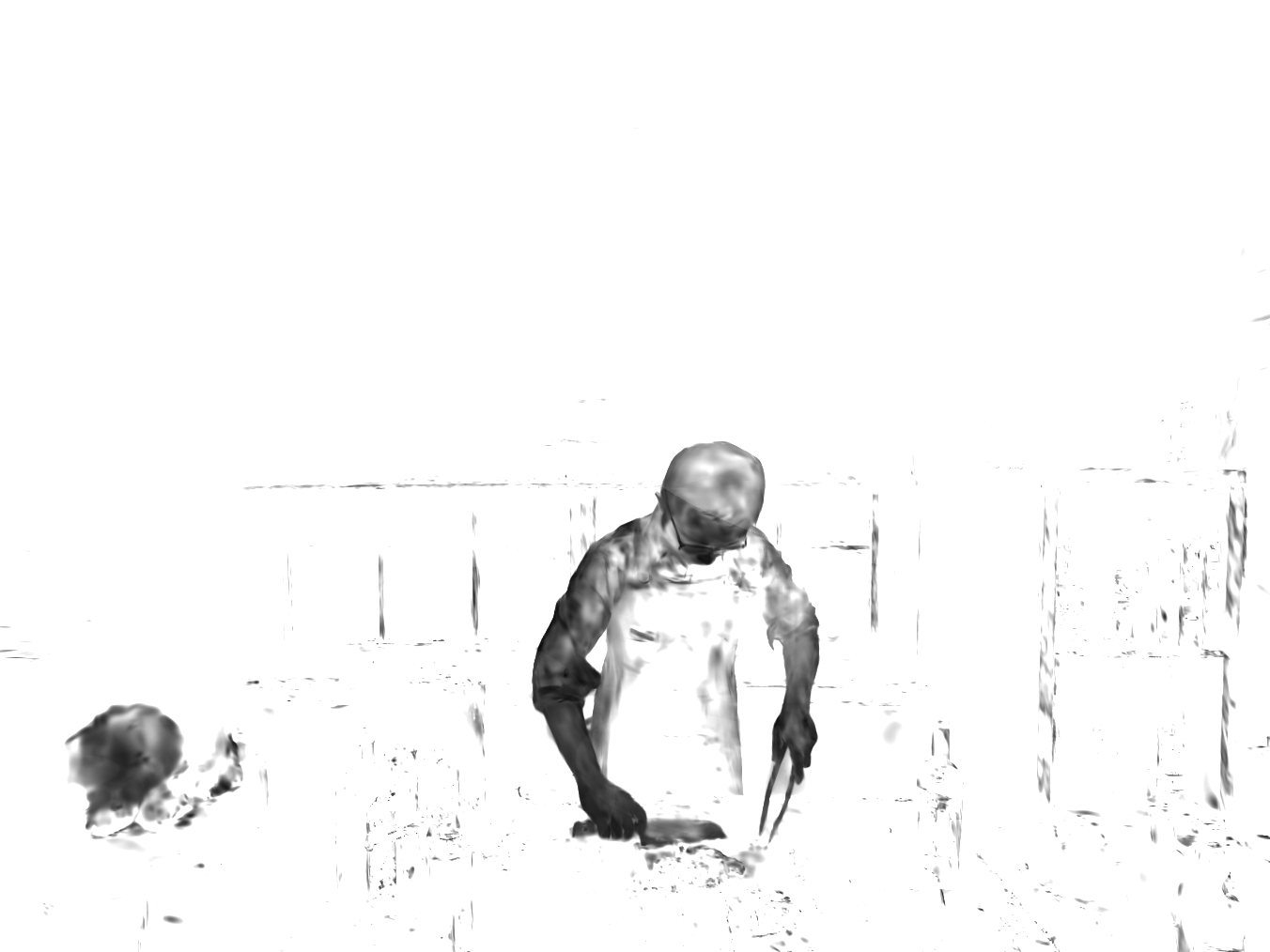}}
    &
    \raisebox{-0.20\height}{\includegraphics[width=.19\textwidth,clip]{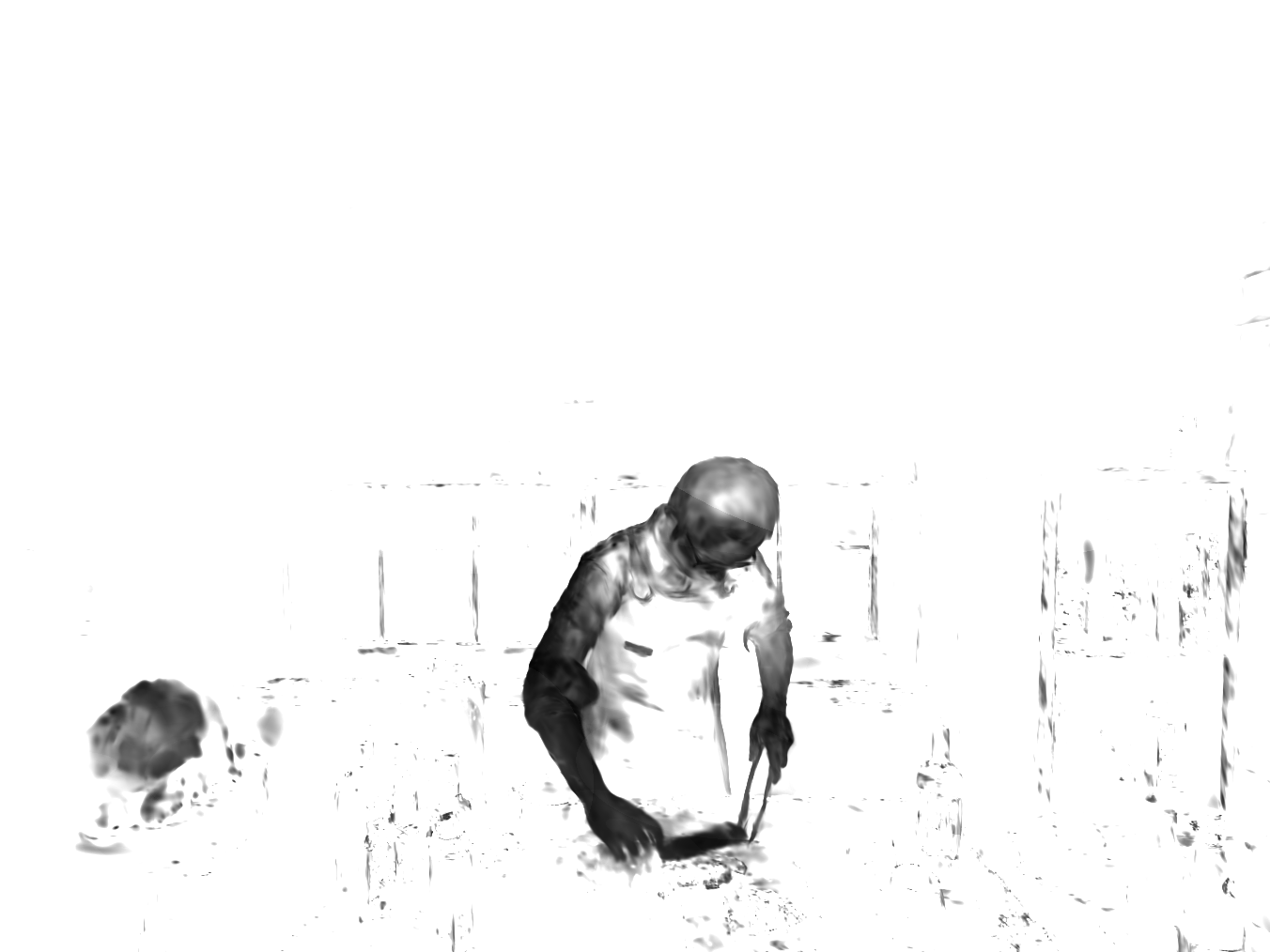}}
    &
    \raisebox{-0.20\height}{\includegraphics[width=.19\textwidth,clip]{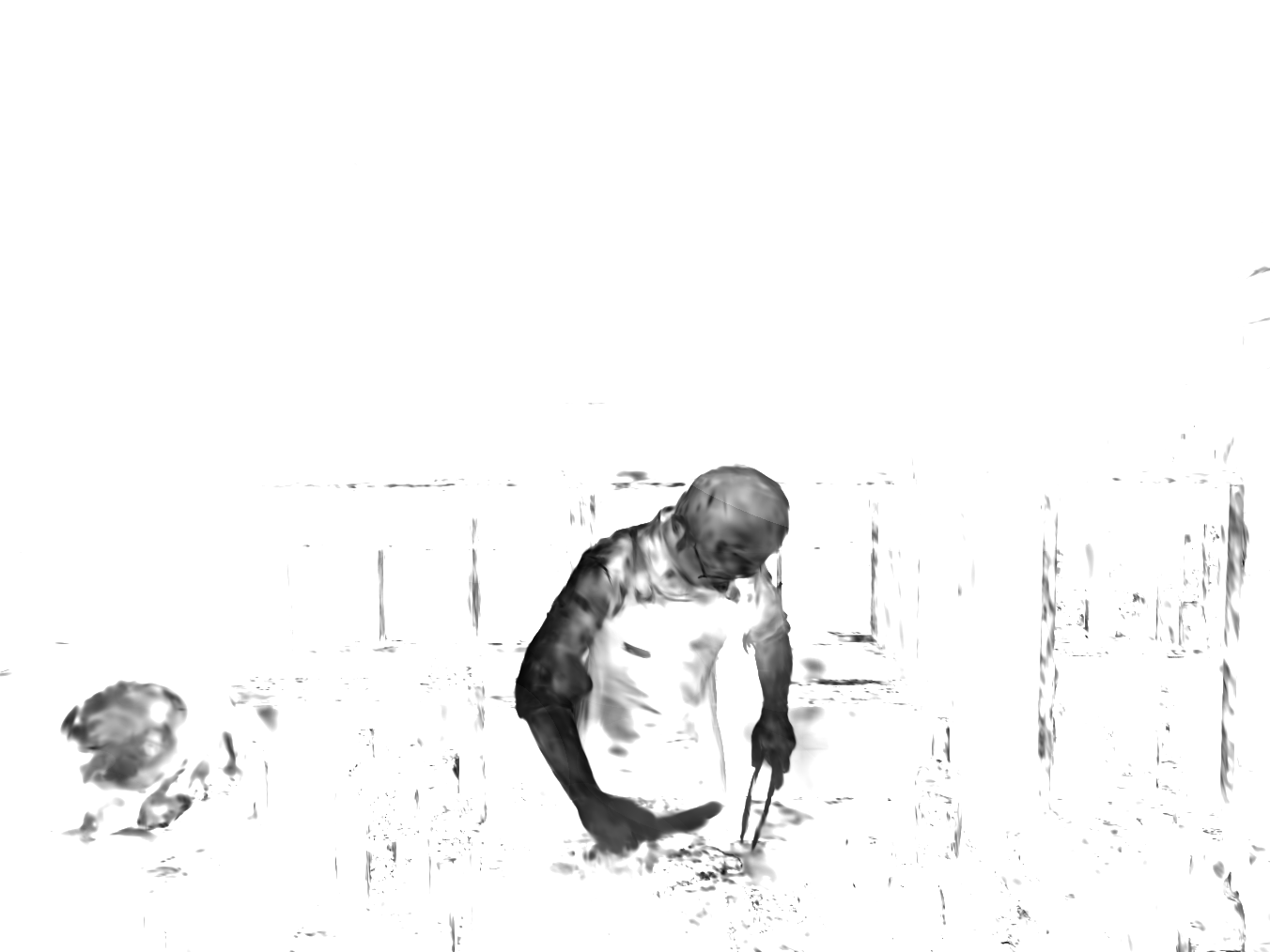}}
    &
    \raisebox{-0.20\height}{\includegraphics[width=.19\textwidth,clip]{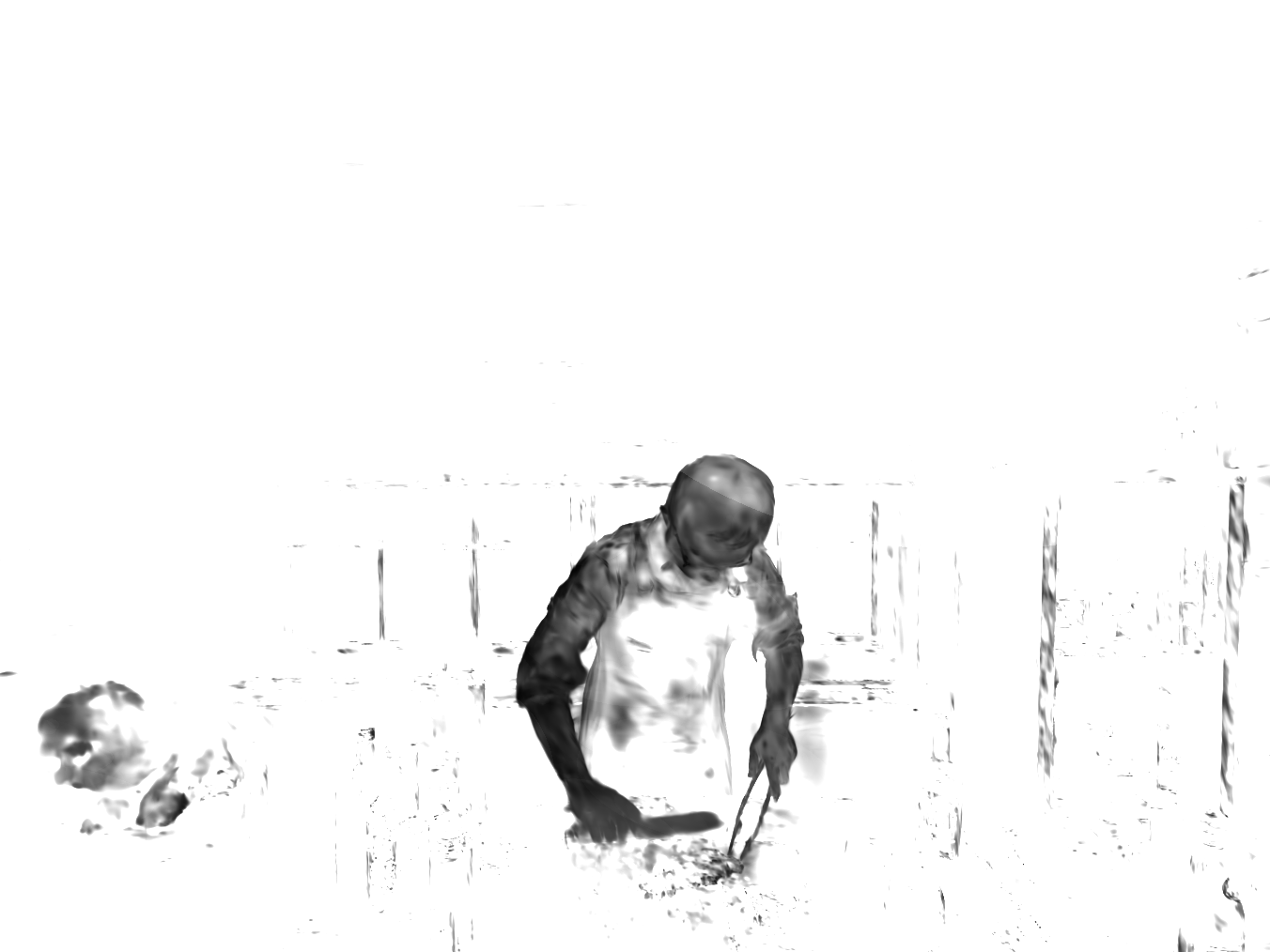}}
 \\
    \rotatebox{90}{\small{\makecell{Variance\\ }}} & \raisebox{-0.20\height}{\includegraphics[width=.19\textwidth,clip]{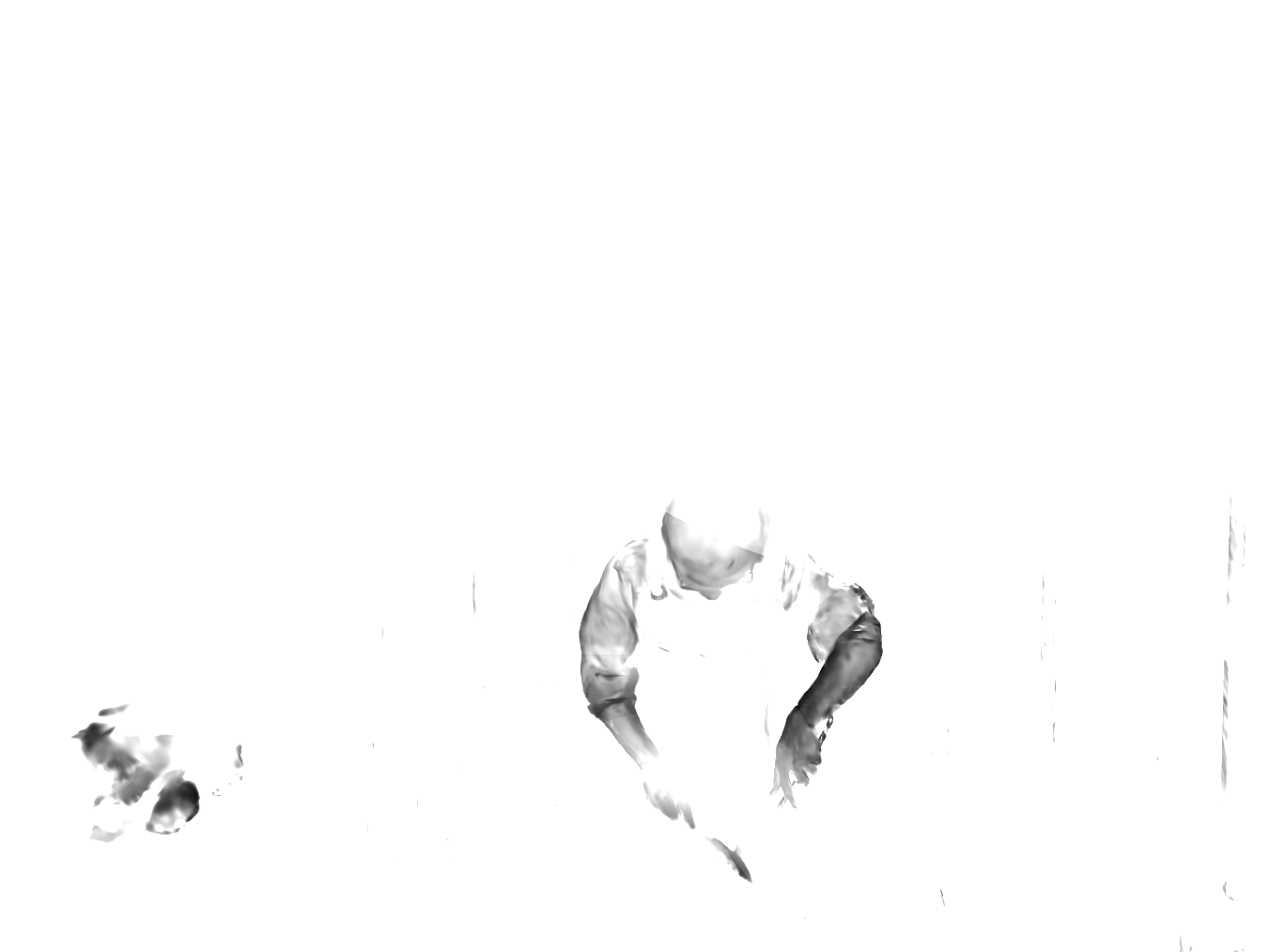}}
    &
    \raisebox{-0.20\height}{\includegraphics[width=.19\textwidth,clip]{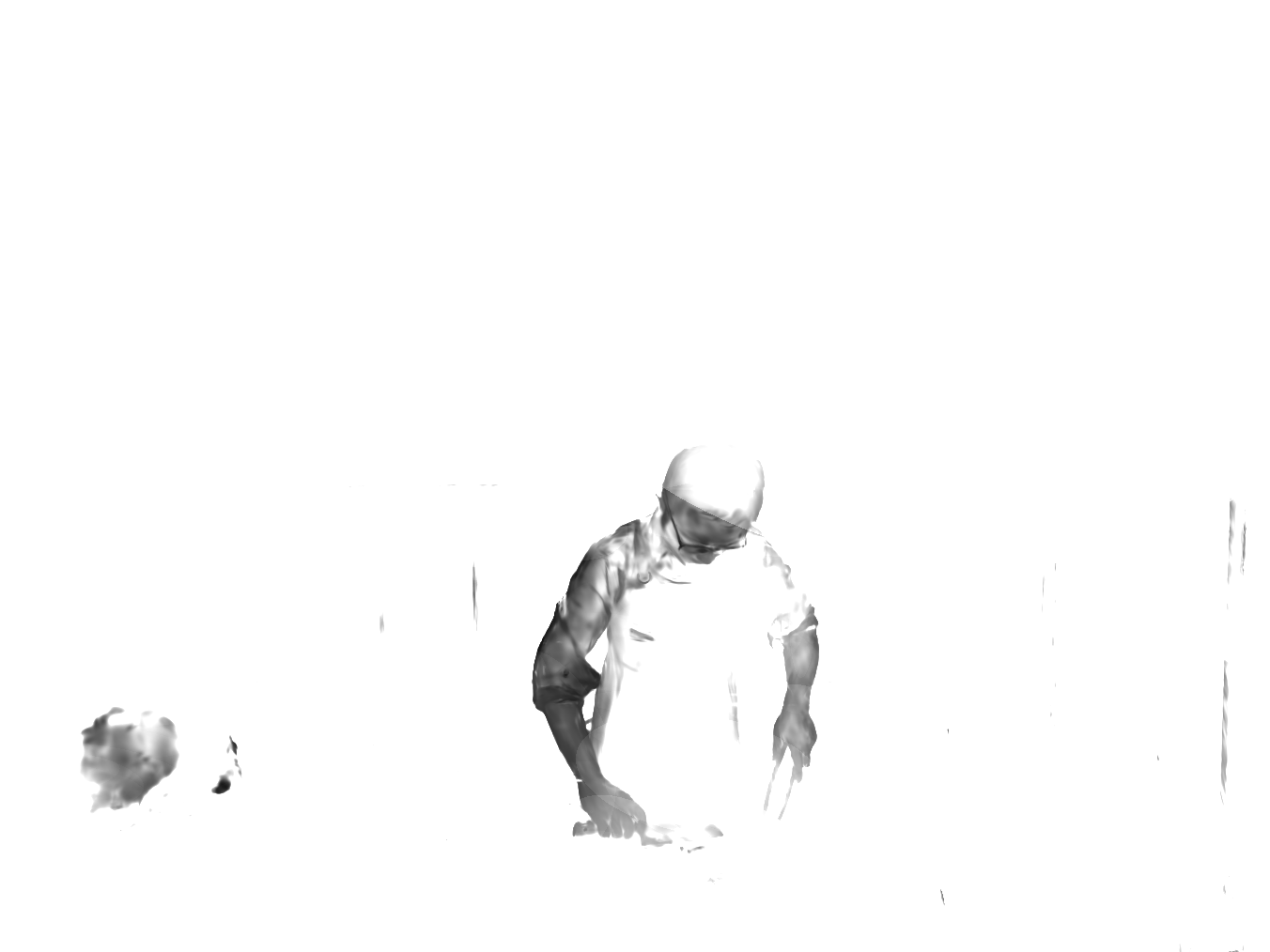}}
    &
    \raisebox{-0.20\height}{\includegraphics[width=.19\textwidth,clip]{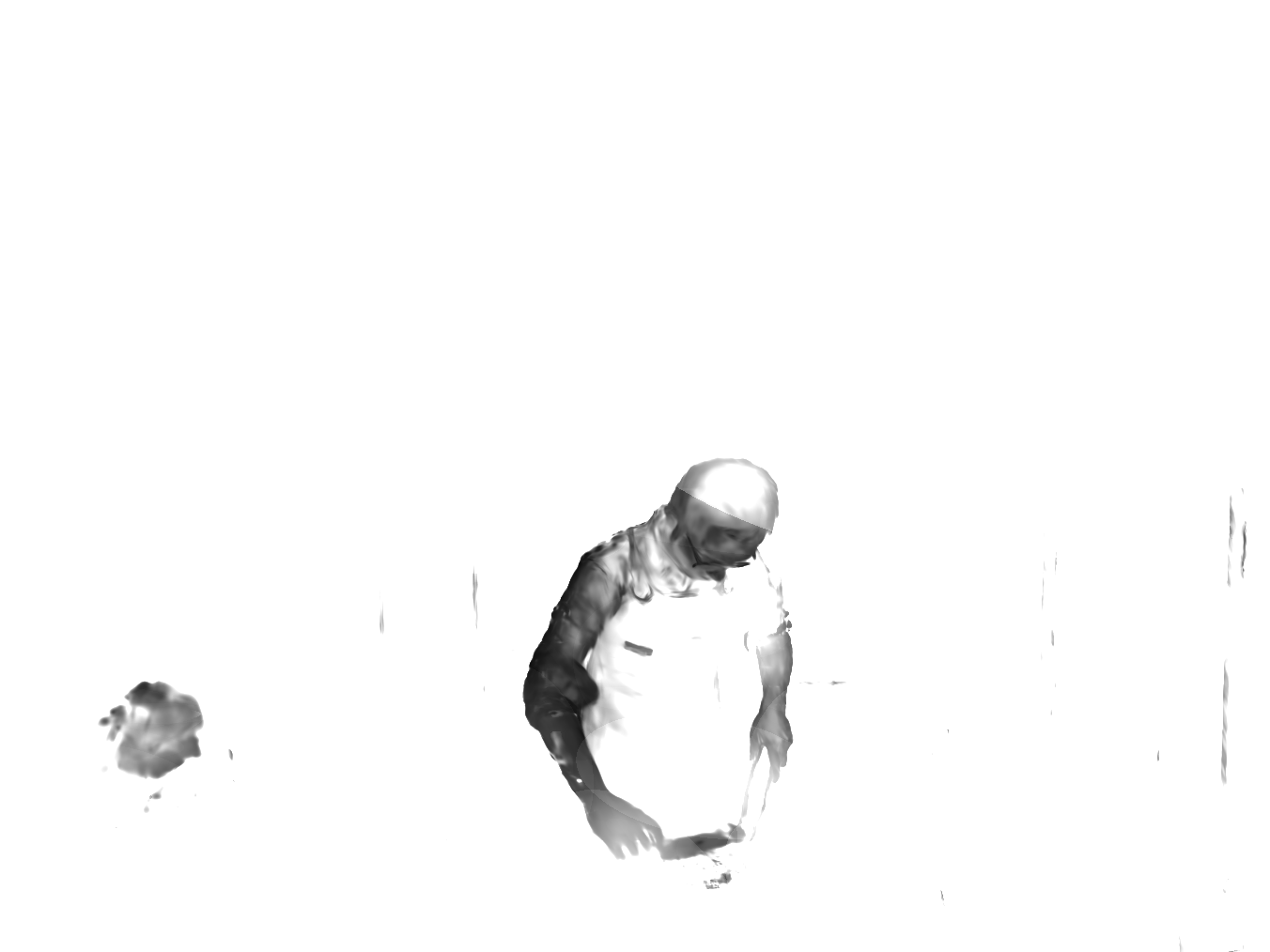}}
    &
    \raisebox{-0.20\height}{\includegraphics[width=.19\textwidth,clip]{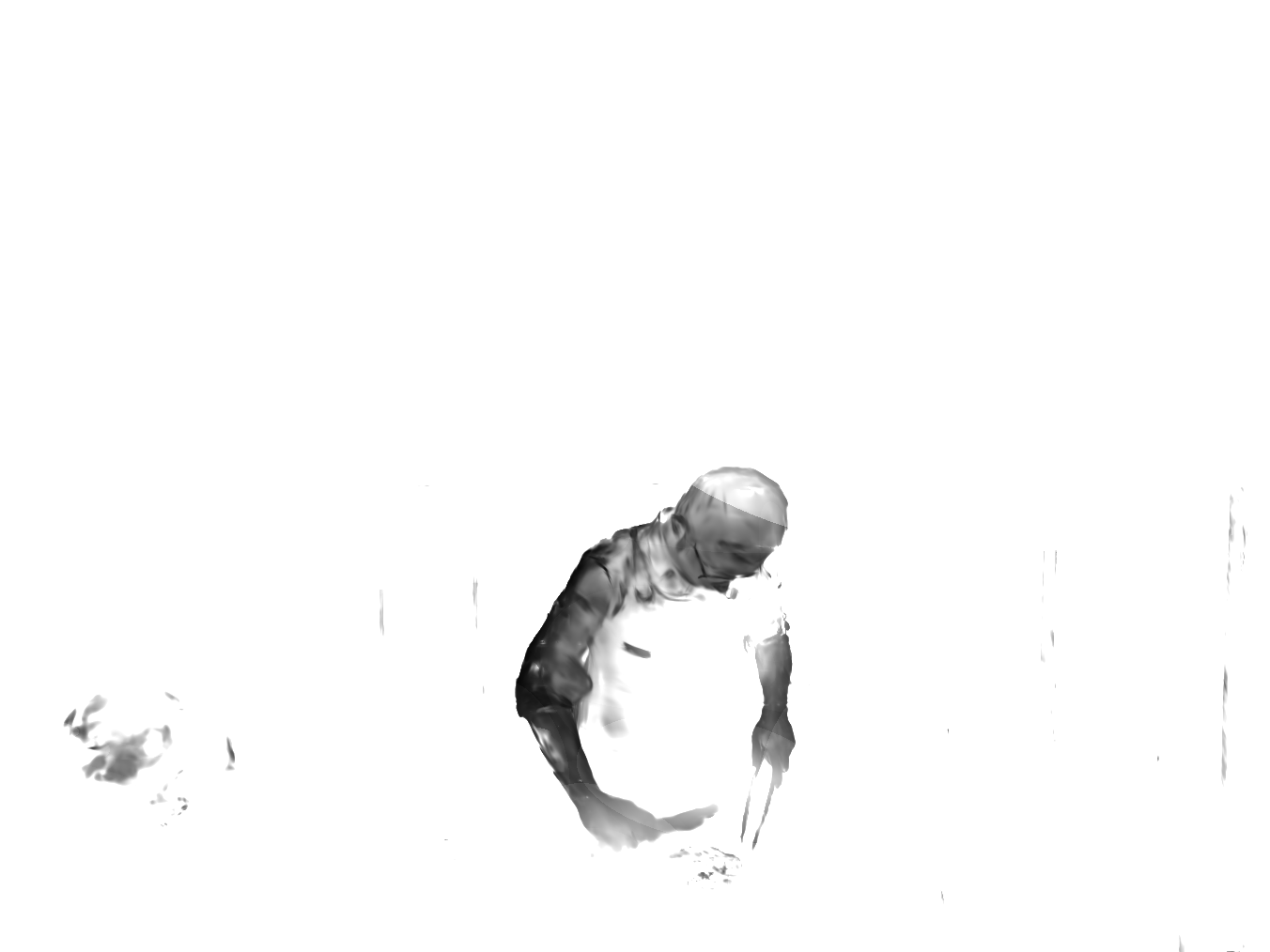}}
    &
    \raisebox{-0.20\height}{\includegraphics[width=.19\textwidth,clip]{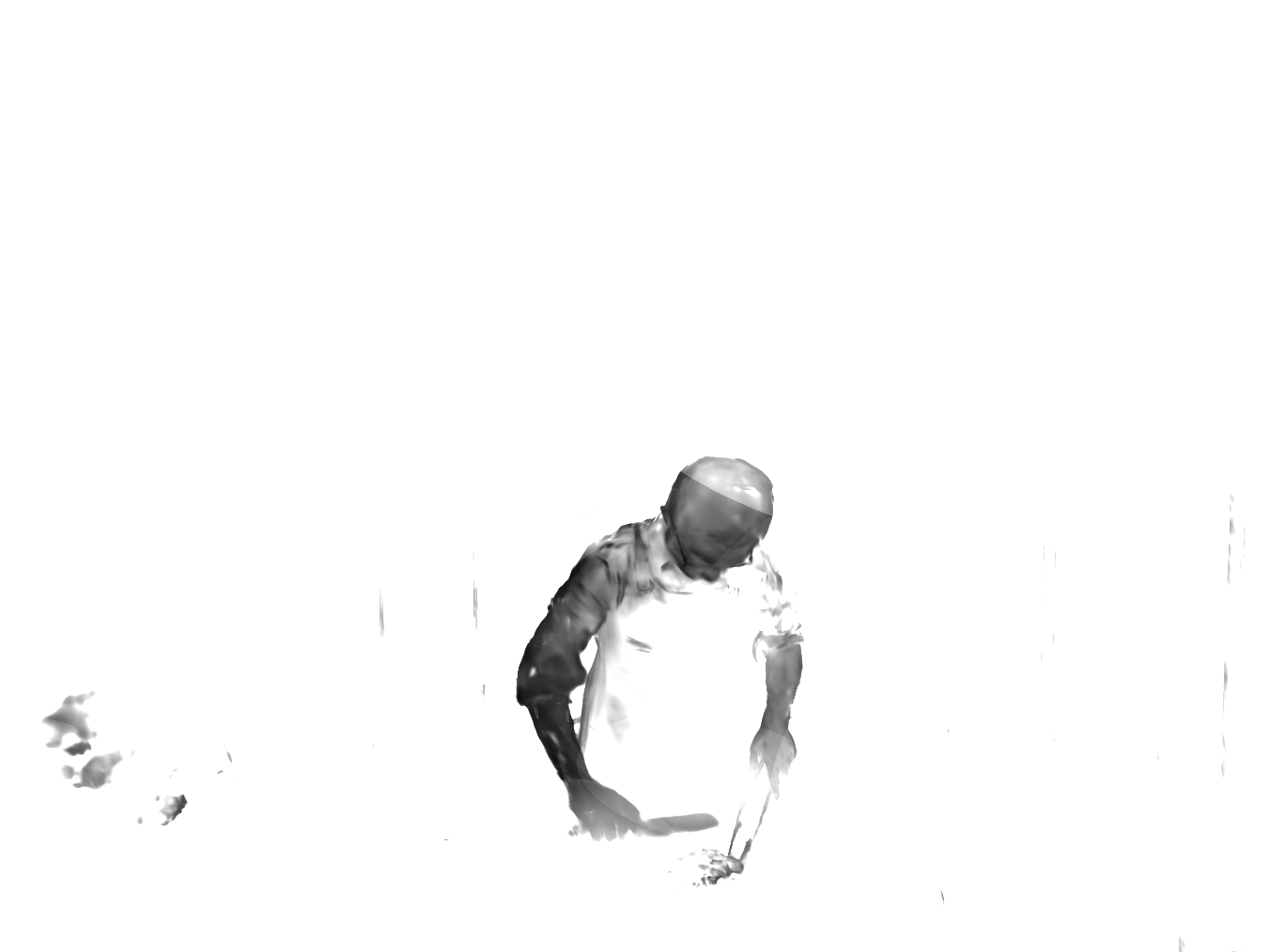}}
 \\
    \end{tabular}
    \caption{\textbf{Temporal distribution of the fitted 4D Gaussian on the \textit{cut roasted beef}.  
    } For better visualization, we show the distance between the mean and the rendered timestamp. }
\label{fig:temporal}
\end{figure*} \begin{figure}
    \centering
    \includegraphics[width=1.0\linewidth]{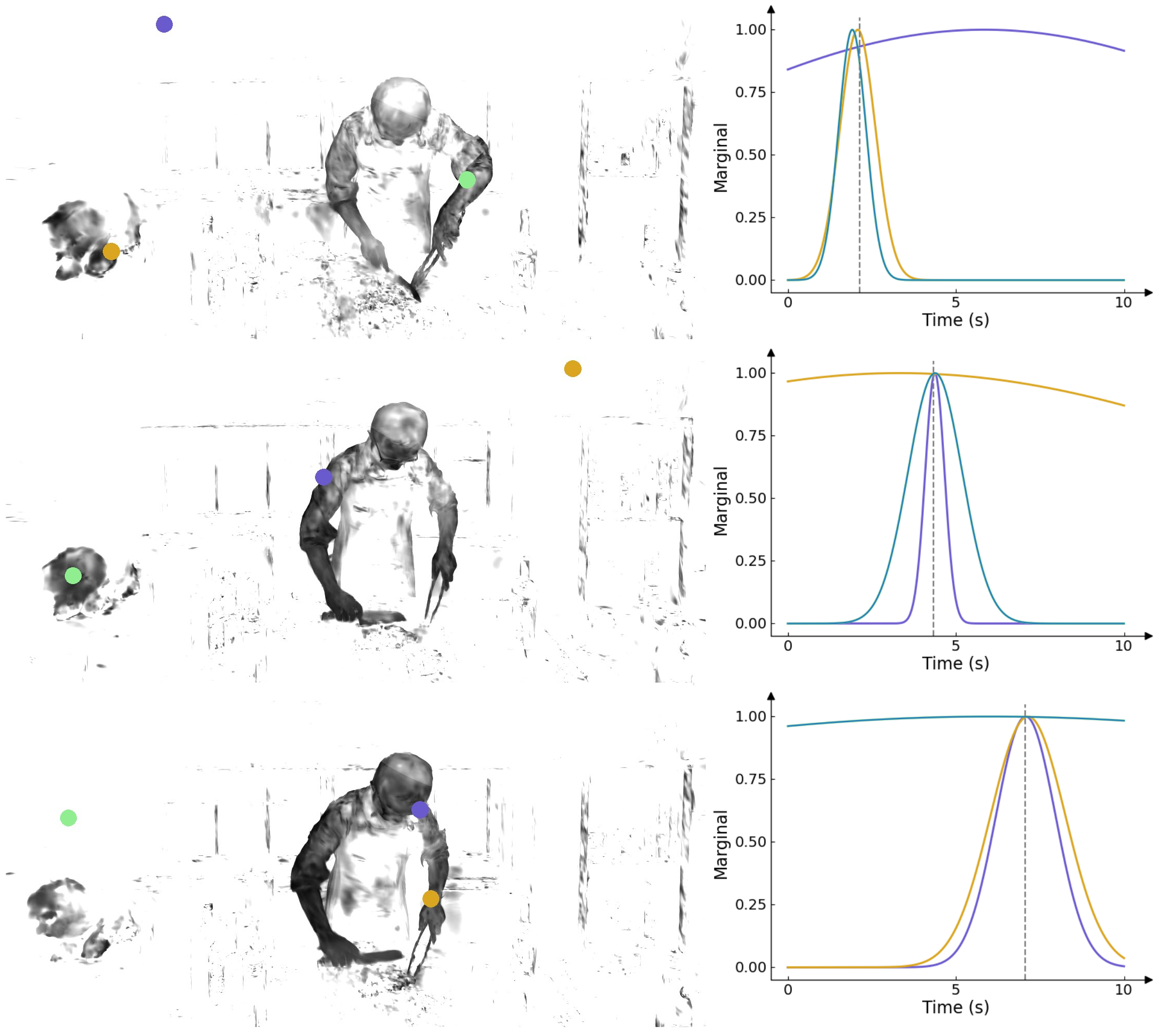}
    \caption{
    \textbf{Marginal distribution of the fitted 4D Gaussian in time dimension.}
}
    \label{fig:margnal_t}
\end{figure} If the 4D Gaussian has only local support in time, as the 3D Gaussian does in space, the number of 4D Gaussians may become very intractable as the video length increases. Fortunately, the anisotropic characteristic of Gaussian offers a prospect of avoiding this predicament. To further unleash the potential of this characteristic, we set the initial time scaling to half of the scene's duration as mentioned in Section~\ref{sec:implement_details}.

In order to more intuitively comprehend the temporal distribution of the fitted 4D Gaussian, Figure~\ref{fig:temporal} presents a visualization of mean and variance in the time dimension, by which the marginal distribution on $t$ of 4D Gaussians can be completely described.
It can be observed that these statistics naturally form a mask to delineate dynamic and static regions, where the background Gaussians have a large variance in the time dimension, which means they are able to cover a long time period. Actually, as shown in Figure~\ref{fig:margnal_t}, the background Gaussians are able to be active throughout the entire time span of the scene, which allows the total number of Gaussians to grow very restrained with the video length extending. 

Moreover, considering that we filter the Gaussians according to the marginal probability $p_i(t)$ at a negligible time cost before the frustum culling, the number of Gaussians actually participated in the rendering of each frame is nearly constant, and thus the rendering speed tends to remain stable with the increase of the video length. This locality instead makes our approach friendly to long videos in terms of rendering speed. 

\noindent \textbf{More ablations}
To examine whether modeling the spatiotemporal evolution of Gaussian's appearance is helpful, we ablate 4DSH in the second row of Table~\ref{tab:ablation_main}. Compared to the result of our default setting, we can find there is indeed a decline in the rendering quality. 
Moreover, when turning our attention to 4D spacetime, we realize that over-reconstruction may occur in more than just space. Thus, we allow the Gaussian to split in time by sampling new positions using a complete 4D Gaussian as PDF. The last two rows in Table~\ref{tab:ablation_main} verified the effectiveness of the densification in time.

\subsection{4D from monocular video}
\begin{table}[t]
\caption{
\textbf{Qualitative comparison on monocular dynamic scenes.} The LPIPS are computed using VGGNet~\cite{simonyan2014very}. }
\label{tab:dnerf}
\renewcommand\tabcolsep{4pt}
\renewcommand\arraystretch{1.2}
\scriptsize
 
\begin{tabular}{l|ccc}
\hline

\hline

\hline

\hline
Method & PSNR $\uparrow$ & SSIM $\uparrow$ & LPIPS $\downarrow$ \\ 
\hline

\hline

\multicolumn{3}{l}{- \textit{D-NeRF (synthetic, monocular)}}  \\
\hline

T-NeRF~\cite{pumarola2021d} & 29.51 & 0.95 & 0.08 \\
D-NeRF~\cite{pumarola2021d} & 29.67 & 0.95 & 0.07 \\
TiNeuVox~\cite{fang2022fast} & 32.67 & 0.97 & 0.04 \\
HexPlanes~\cite{cao2023hexplane} & 31.04 & 0.97 & 0.04 \\
K-Planes-explicit~\cite{fridovich2023kplane} & 31.05 & 0.97 & - \\
K-Planes-hybrid~\cite{fridovich2023kplane} & 31.61 & 0.97 & - \\
V4D~\cite{gan2023v4d} & 33.72 & 0.98 & 0.02 \\
4DGS-HexPlanes~\cite{wu20234dgaussians}\footnotemark[1] & 33.30 & 0.98 & 0.03 \\
\rowcolor[gray]{.9}
\textbf{4DGS (Ours)} & \textbf{34.09} & \textbf{0.98} & \textbf{0.02} \\

\hline

\hline
\end{tabular}
\footnotetext[1]{Rendered at 800$\times$800, otherwise downsampled 2x by default.}
\end{table} \subsubsection{Results of 4D reconstruction}

Novel view synthesis for dynamic scenes from monocular video is an inherently under-constrained problem. 
An ideal solution often requires incorporating proper architectural priors. 
Although our method, designed as a general-purpose representation for dynamic scenes, does not directly introduce such priors, it still achieves comparable or even better performance than prior art.
To validate that, we evaluate our approach on the commonly used monocular dataset.
Since the real-world monocular videos for dynamic scenes usually suffer from inaccurate pose estimation, and the effectiveness in complex real-world scenes has been thoroughly validated in the previous section, we test the monocular dynamic reconstruction 
D-NeRF dataset~\cite{pumarola2021d}, which comprises 8 object-centric videos synthesized using graphic engine with accurate camera poses accessible.
For experiments on this dataset, we initialize the mean position of 4D Gaussians with 100,000 randomly selected points evenly distributed within the cubic volume defined by $[-1.2, 1.2]^3$.
suggest that the proposed general representation can also work well in such an ill-posed task, highlighting its versatility and potential.

{\color{cyan}
\subsubsection{4D from monocular video with generative prior}
\label{sec:exp_generation}

\begin{table}[t]
\caption{
{\color{cyan}
\textbf{
Comparison of 4D representations in video-to-4D generation.}
To perform fair comparison, we only use anchored images and SDS loss as supervision without any other techniques introduced in representative methods.
}
}
\label{tab:generation}
\renewcommand\tabcolsep{1.2pt}
\renewcommand\arraystretch{1.2}
\tiny
\begin{tabular}{c|c|ccc}
\hline

\hline

\hline

\hline
Representation & 
\makecell{Representative \\methods} & \makecell{Total time$\downarrow$ \\ (time per iter$\downarrow$)} & CLIP$\uparrow$ & 
LPIPS$\downarrow$ \\
\hline

\hline
C-DyNeRF~\cite{jiang2023consistent4d} & Consistent4D~\cite{jiang2023consistent4d} & 3h (11s) & 0.875 & 0.147 \\
\hline

\hline
Deform-GS~\cite{wu20234dgaussians} & \makecell{DG4D~\cite{ren2023dreamgaussian4d}\\ 4DGen~\cite{yin20234dgen} \\ STAG4D~\cite{zeng2024stag4d}} & 22m (2.6s) & 0.898 & 0.133 \\
\hline

\hline
\rowcolor[gray]{.9}
\bf 4DGS (Ours) & Efficient4D~\cite{pan2024efficient4d} & \bf 9m30s (1.1s) & \bf 0.905 & \bf 0.129 \\

\hline

\hline
\end{tabular}
\end{table} \begin{figure}[t]
    \centering
    \includegraphics[width=0.9\linewidth]{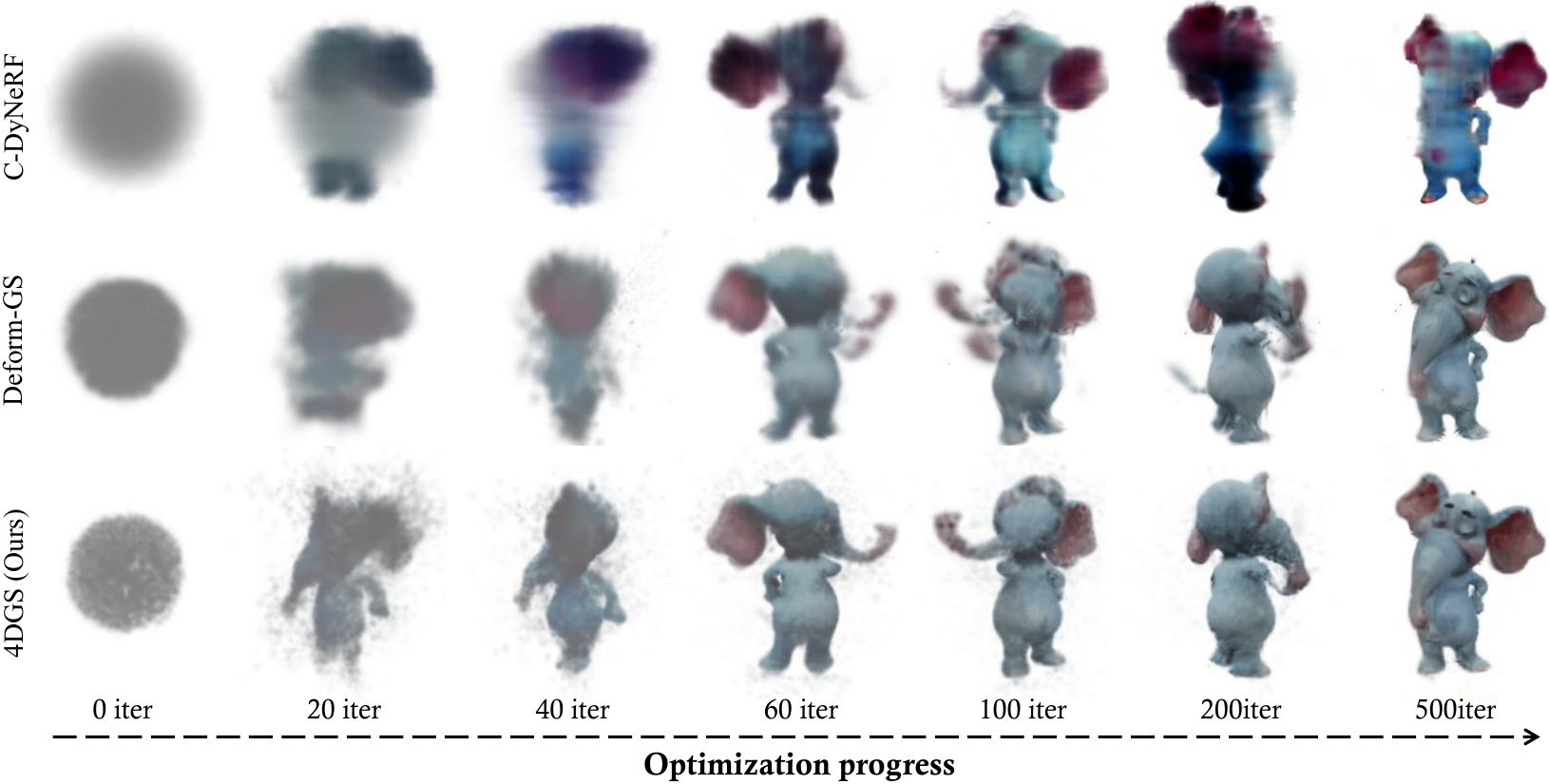}
    \caption{{\color{cyan}
    \textbf{
    Training progress with different 4D representations under the same objective.} Our 4DGS achieves the fastest convergence, while exhibiting the best robustness on the inconsistency between reference views.}
}
    \label{fig:generation}

\end{figure} 
\begin{figure*}[t]
    \centering
    \includegraphics[width=0.98\linewidth]{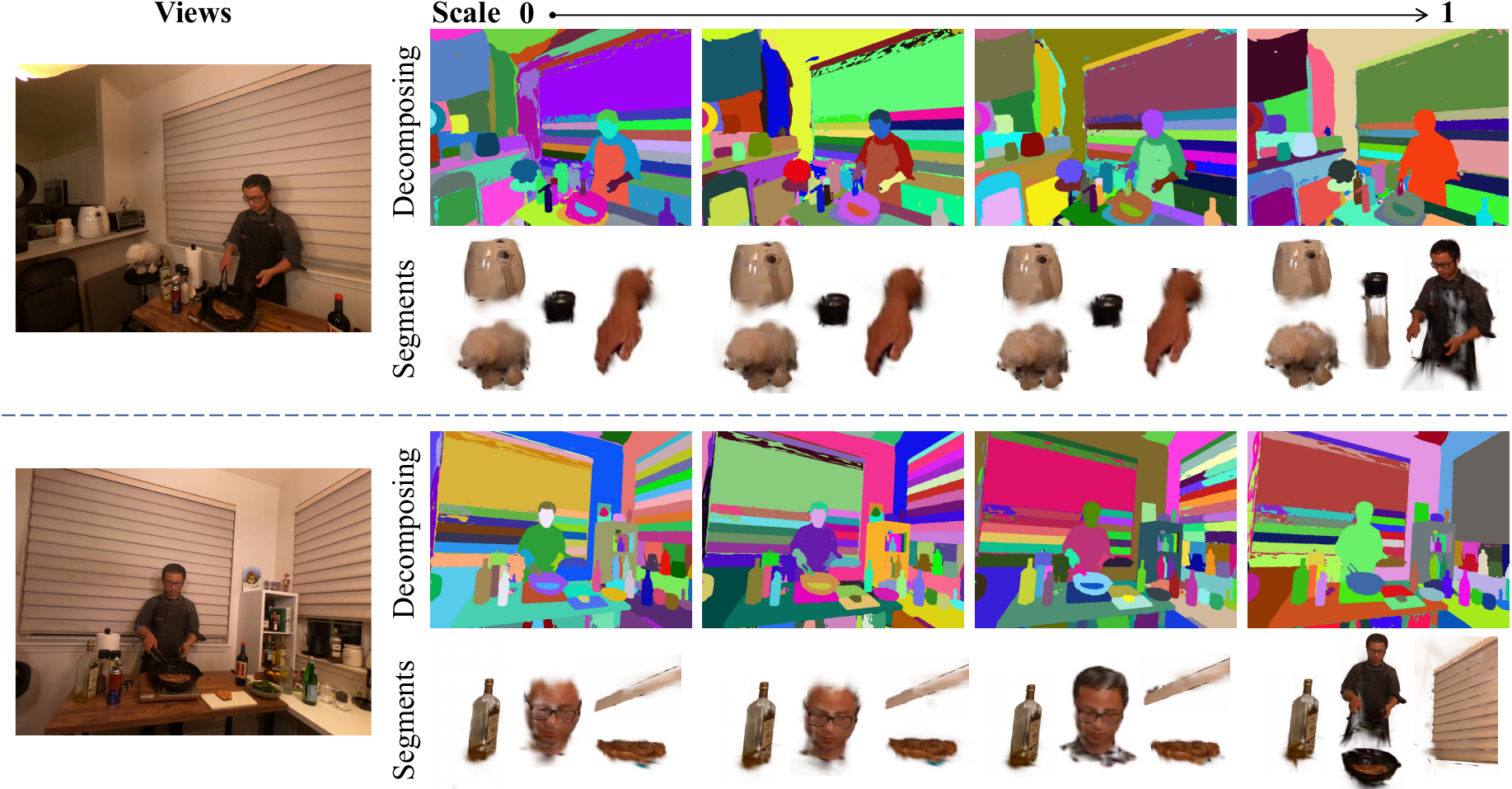}
    \caption{{\color{cyan}
    \textbf{Segmentation results.} We show two views of the scene decomposition and object segments at different scales. Please refer to our \href{https://fudan-zvg.github.io/4d-gaussian-splatting/}{project page} for the corresponding videos.
    }}
    \label{fig:segment}
\end{figure*} 
\noindent \textbf{Dataset} 
Consistent4D test dataset~\cite{jiang2023consistent4d}: comprises seven videos with 32 frames, each containing $32\times4$ ground truth images from four viewpoints at every timestamp. In this dataset, we initialize 50,000 random points inside a sphere of radius 0.5 with uniformly distributed timestamps.

\noindent \textbf{Implementation details}
By default, we conduct training with a total of
500 iterations and halted densification at the 100$th$ iteration. The weights in equation~(\ref{eq:loss_generation}) are set as $\lambda_{\text{img}}=10,000$ and $\lambda_{\text{sds}}=1$. The other settings for hyperparameters are the same as reconstruction in Section~\ref{sec:implement_details}.

\noindent \textbf{Baselines}
We include cascade dynamic NeRF (C-DyNeRF) introduced in \cite{jiang2023consistent4d} and Deform-GS~\cite{wu20234dgaussians} as alternative representations for comparison. Due to different convergence speeds, we train C-DyNeRF for 1000 iterations and Deform-GS for 500 iterations. The other training settings follow those specified in the respective papers.

\noindent \textbf{Metrics} 
For each 4D representation, we report the total generation time needed for complete convergence and the time required for a single training iteration. We report the CLIP similarity and LPIPS between rendered images and ground-truth images.

\noindent \textbf{Results}
Table~\ref{tab:generation} reports the results on the Consistent4D test dataset. Our 4DGS significantly surpasses all alternatives in terms of both generation time and quality. 
Figure~\ref{fig:generation} shows that 4DGS is fastest to converge.
We note C-DyNeRF fails to converge in 500 iterations, whereas Gaussian-based representations succeed. 

\subsection{Experiments for the compact 4D Gaussian}
\label{sec:exp_compact}

\subsubsection{Gaussian attributes compression}
\label{sec:exp_gaussian_attributes_comp}

\begin{table*}[t]
\caption{{\color{cyan}
\textbf{The impact of parameter compression.} Metrics are evaluated on the cut roasted beef. All differences are compared to the baseline. 
For each attribute, we report the compression ratio with respect to the storage of itself.
}}
\label{tab:compact4dgs}
\renewcommand\tabcolsep{4pt}
\renewcommand\arraystretch{1.2}
\tiny
\begin{tabular}{c|ccc|ccc|ccc|ccc}
\hline

\hline

\hline

\hline
& \multicolumn{3}{c}{Scaling \& Rotation} \vline &
                        \multicolumn{3}{c}{Position}\vline &
                        \multicolumn{3}{c}{Opacity}\vline &
                        \multicolumn{3}{c}{Color} 
\\ 
 & PSNR$\uparrow$ & SSIM$\uparrow$ & Storge$\downarrow$
 & PSNR$\uparrow$ & SSIM$\uparrow$ & Storge$\downarrow$
 & PSNR$\uparrow$ & SSIM$\uparrow$ & Storge$\downarrow$
 & PSNR$\uparrow$ & SSIM$\uparrow$ & Storge$\downarrow$
\\ 
\hline

\hline
4DGS & \multicolumn{4}{c}{PSNR: 33.96} & \multicolumn{4}{c}{SSIM: 0.9797} & \multicolumn{4}{c}{Storge: 1183.21MB} \\
\hline
+R-VQ & \textcolor{blue}{-0.49} & \textcolor{blue}{-0.0020} & \textcolor{red}{-71.88\%} & \multicolumn{3}{c}{-}\vline & \multicolumn{3}{c}{-}\vline & \textcolor{blue}{-0.44} & \textcolor{blue}{-0.0015} & \textcolor{red}{-96.87\%} \\
+8-bit quant. & \multicolumn{3}{c}{-}\vline & \multicolumn{3}{c}{-}\vline & \textcolor{black}{-0.00} & \textcolor{black}{-0.0000} & \textcolor{red}{-75.00\%} & \multicolumn{3}{c}{-}\\
+16-bit quant. & \multicolumn{3}{c}{-}\vline & \textcolor{blue}{-0.01} & \textcolor{blue}{-0.0001} & \textcolor{red}{-50.00\%} & \multicolumn{3}{c}{-}\vline  & \multicolumn{3}{c}{-} \\
+fine-tune & \textcolor{red}{+0.02} & \textcolor{red}{+0.0003} & \textcolor{red}{-71.88\%} & \textcolor{red}{+0.10} & \textcolor{red}{+0.0007} & \textcolor{red}{-50.00\%} & \textcolor{red}{+0.11} & \textcolor{red}{+0.0007} & \textcolor{red}{-75.00\%} & \textcolor{red}{+0.08} & \textcolor{red}{+0.0007} & \textcolor{red}{-96.87\%} \\
+encode & \textcolor{red}{+0.02} & \textcolor{red}{+0.0003} & \textcolor{red}{-72.56\%} & \multicolumn{3}{c}{-}\vline & \textcolor{red}{+0.11} & \textcolor{red}{+0.0007} & \textcolor{red}{-76.33\%} & \textcolor{red}{+0.08} & \textcolor{red}{+0.0007} & \textcolor{red}{-98.93\%} \\
\hline
+All & \multicolumn{4}{c}{PSNR: 33.86 \textcolor{blue}{(-0.1)}} & \multicolumn{4}{c}{SSIM: 0.9795 \textcolor{blue}{(-0.0002)}} & \multicolumn{4}{c}{Storge: 56.74MB \textcolor{red}{(-95.20\%)}} \\
\hline

\hline
\end{tabular}
\end{table*} \noindent \textbf{Implementation details}
The baseline model adopts the default settings except for increasing the opacity pruning threshold to 0.05. For scaling, we employ an R-VQ model with $6$ codebooks each containing $64$ entries. The codebook parameters are optimized during $512$ extra steps with a learning rate $4 \times 10^{-4}$ on the pre-trained baseline model. Both rotation parameters share the same R-VQ model also consisting of $6$ codebooks with 64 entries, optimized for $1024$ steps at $8 \times 10^{-4}$ learning rate.
For appearance modeling, the direct and rest components adopt $6$ codebooks with $64$ entries and $8$ codebooks with $256$ entries, respectively.
For opacity, we use quantization-aware fine-tuning with $k$-bit Min-Max Quantization~\cite{niedermayr2024compressed,rastegari2016xnor} to fine-tune for $1000$ iterations with $0.1 \times$ learning rate.
All experiments are conducted on the \textit{Cut Roasted Beef} from~\cite{li2022neural}.

\noindent \textbf{Results}
In Table~\ref{tab:compact4dgs}, we validate the effects of the parameter compression strategy introduced in Section~\ref{sec:paramcompression}. For the shape-related attributes, \ie, rotation and scaling, the R-VQ reduces their memory from $12 \times 32 \times N$ bits to $3 \times 6 \times 6 \times N + 2 \times 6 \times 64 \times 4 \times 32$ bits for $N$ Gaussians. 
Since the real scenes could contain millions of Gaussians, this process can achieve around $3.5\times$ compression rates with negligible damage to quality.
After VQ and Huffman encoding on the codebook indices, their storage consumption decreases from 88.19 MB to 24.20 MB. Although it leads to a 0.49 dB PSNR drop, the post-finetuning of the codebook entries can recover the PSNR by 0.51 dB. 
Similarly, the size of color-related parameters decreases by nearly 1000 MB with little PSNR drop.
Quantization of position and opacity parameters reduces storage overhead by 50\% and 75\% respectively, with a negligible performance cost. And the post-finetuning even yields better performance than the original model.
By utilizing all of the above strategies together, the total memory overhead decreases from 1183.21 MB to 56.74 MB, while almost maintaining the same rendering quality.

\subsubsection{Culling redundant Gaussians}
\begin{figure}
    \centering
    \includegraphics[width=1.0\linewidth]{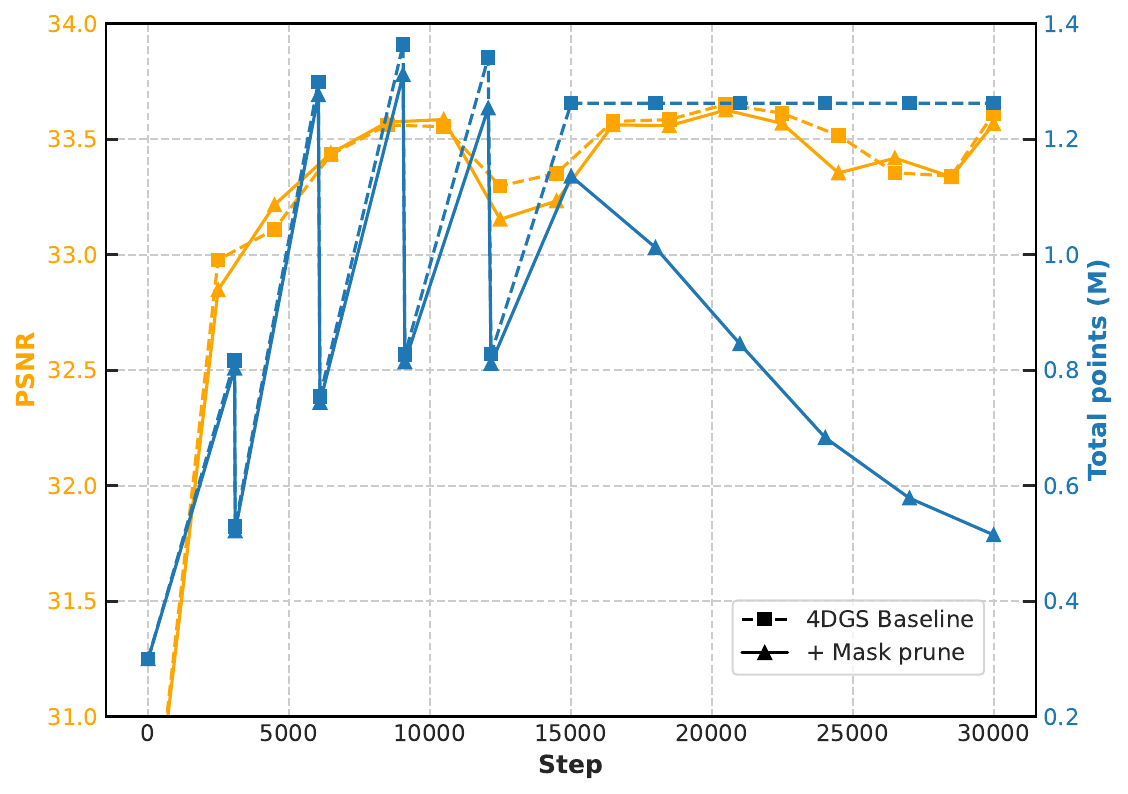}
    \caption{
    \textbf{The progress of testing PSNR and total points number during the training on the \textit{sear steak}.}
}
    \label{fig:mask_prune}
\end{figure} \begin{table}[t]
\caption{{\color{cyan}
\textbf{Impact of pruning thresholds of opacity.}
}}
\label{tab:ablation_prune}
\renewcommand\tabcolsep{4pt}
\renewcommand\arraystretch{1.4}
\scriptsize
\begin{tabular}{cc|ccc}
\hline

\hline

\hline

\hline

\makecell{Opacity\\threshold} & Scene & PSNR$\uparrow$ & SSIM$\uparrow$ & \#Points$\downarrow$ \\

\hline

\hline

\multirow{2}{*}{0.005} & Sear Steak  & 33.53 & 0.963 & 2.54M \\
 & Cut Beef & 33.87 & 0.980 & 3.05M \\

\hline

\multirow{2}{*}{0.05} & Sear Steak & 33.81 & 0.962 & 1.26M \\
 & Cut Beef & 34.02 & 0.980 & 1.49M \\

\hline

\hline
\end{tabular}
\end{table} For redundant primitive pruning, we compare two methods for eliminating Gaussians: the volume mask~\cite{lee2024compact} and vanilla opacity with varying thresholds from conservative to aggressive. This comparison clearly demonstrates the trade-off between performance and storage cost under different criteria. Specifically, we set $\varepsilon = 0.03$ in equation~(\ref{eq:mask_prune}) for the volume mask and use opacity thresholds of 0.005 and 0.05. The pruning experiments are conducted on the {\it Sear Steak} sequence from the Plenoptic Video dataset.

As shown in Figure~\ref{fig:mask_prune}, the total number of Gaussian points decreases rapidly after 
the stop of densification
when using the volume mask, while the PSNR remains stable, indicating that the mask pruning technique effectively removes redundant Gaussians. This results in a $2.47\times$ reduction in the total number of points, from 1.26M to 0.51M.
Table~\ref{tab:ablation_prune} provides the detailed quantitative results. A higher opacity threshold eliminates floaters in the scene, leading to a higher PSNR (33.46 vs. 30.48) and decreasing the number of Gaussian points from 2.62M to 1.26M. Additionally, applying mask pruning further reduces the number of redundant Gaussians while preserving similar rendering quality. Given that the mask pruning technique introduces minimal additional computational overhead, we choose to use it with a threshold of 0.05 as the default method for reducing the number of Gaussians.

\subsection{Experiments on dynamic urban scenes}
\label{sec:exp_urban}
\begin{table}[t]
\caption{
{\color{cyan}
\textbf{Qualitative results of tasks on the dynamic urban scenes.} Our results are measured on the Dynamic-32 split of the Waymo-NOTR dataset.}}
\label{tab:waymonotr}
\renewcommand\arraystretch{1.2}
\scriptsize
\begin{tabular}{l|ccc}
\hline

\hline

\hline

\hline
Method & PSNR $\uparrow$ & SSIM $\uparrow$ & LPIPS $\downarrow$ \\ 
\hline

\hline

\multicolumn{3}{l}{- \textit{Scene reconstruction}}  \\
\hline

D$^2$NeRF~\cite{wu2022d} & 24.35 & 0.645 & - \\
HyperNeRF~\cite{park2021hypernerf} & 25.17 & 0.688 & - \\
EmerNeRF~\cite{yang2023emernerf} & 28.87 & 0.814 & - \\
MARS~\cite{wu2023mars} & 28.24 & 0.866 & 0.252 \\
3DGS~\cite{kerbl3Dgaussians} & 28.47 & 0.876 & 0.136 \\
S$^3$Gaussian~\cite{huang2024textit} & 31.35 & 0.911 & 0.106 \\
StreetGaussian~\cite{yan2024street} & 29.89 & 0.906 & 0.125 \\
\rowcolor[gray]{.9}
\textbf{4DGS (Ours)} & \textbf{35.04} & \textbf{0.943} & \textbf{0.057} \\
\hline

\multicolumn{3}{l}{- \textit{Novel view synthesis}}  \\
\hline

D$^2$NeRF~\cite{wu2022d} & 24.17 & 0.642 & - \\
HyperNeRF~\cite{park2021hypernerf} & 24.71 & 0.682 & - \\
EmerNeRF~\cite{yang2023emernerf} & 27.62 & 0.792 & - \\
MARS~\cite{wu2023mars} & 26.61 & 0.796 & 0.305 \\
3DGS~\cite{kerbl3Dgaussians} & 25.14 & 0.813 & 0.165 \\
S$^3$Gaussian~\cite{huang2024textit} & 27.44 & 0.857 & 0.137 \\
StreetGaussian~\cite{yan2024street} & 28.19 & 0.867 & 0.133 \\
\rowcolor[gray]{.9}
\textbf{4DGS (Ours)} & \textbf{28.67} & \textbf{0.868} & \textbf{0.102} \\

\hline

\hline
\end{tabular}
\end{table} \begin{figure}[t]
    \centering 
    \setlength{\tabcolsep}{1.0pt}
    \begin{tabular}{cc} 
    {\includegraphics[width=.45\linewidth,clip]{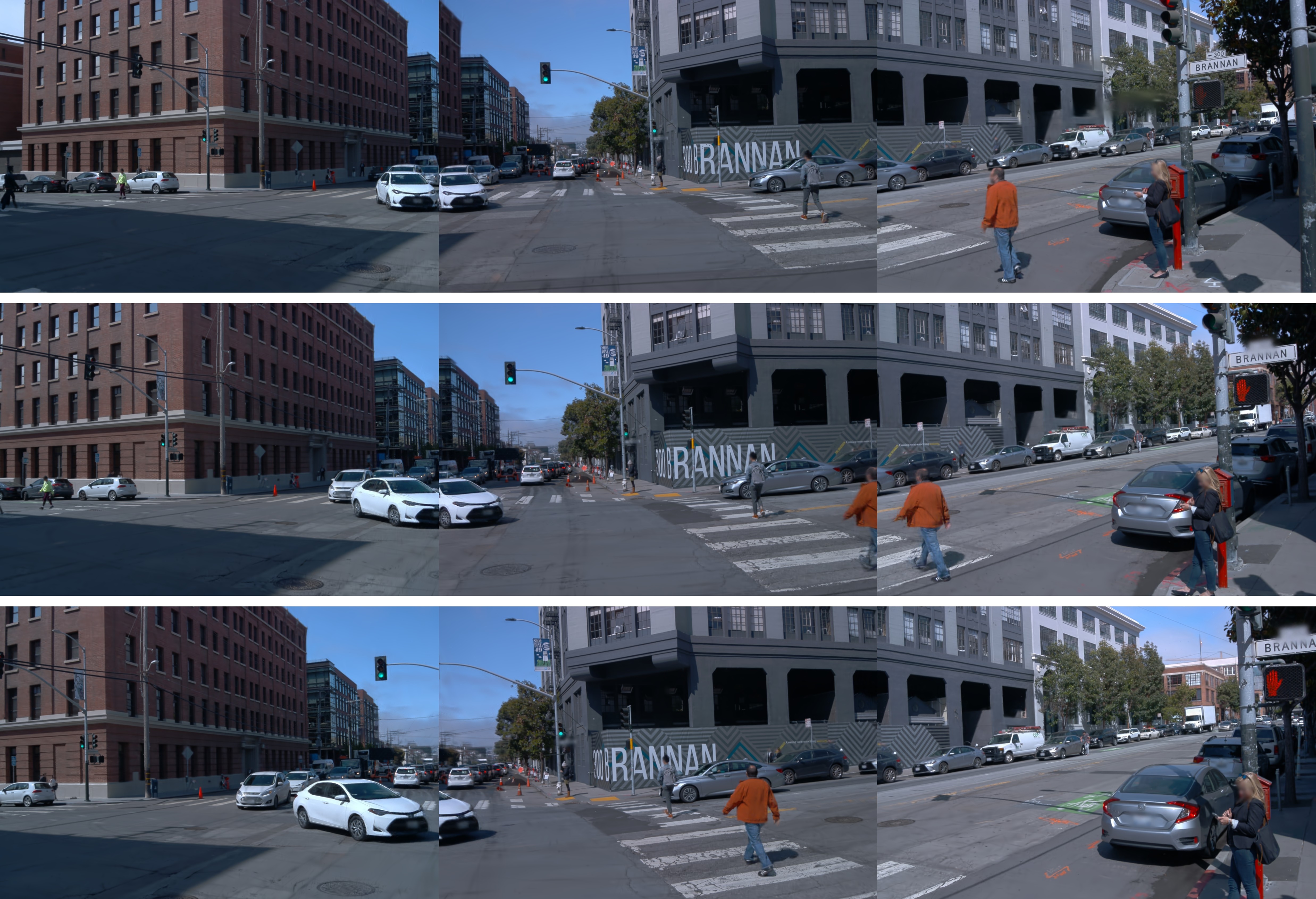}}
    &
    {\includegraphics[width=.45\linewidth,clip]{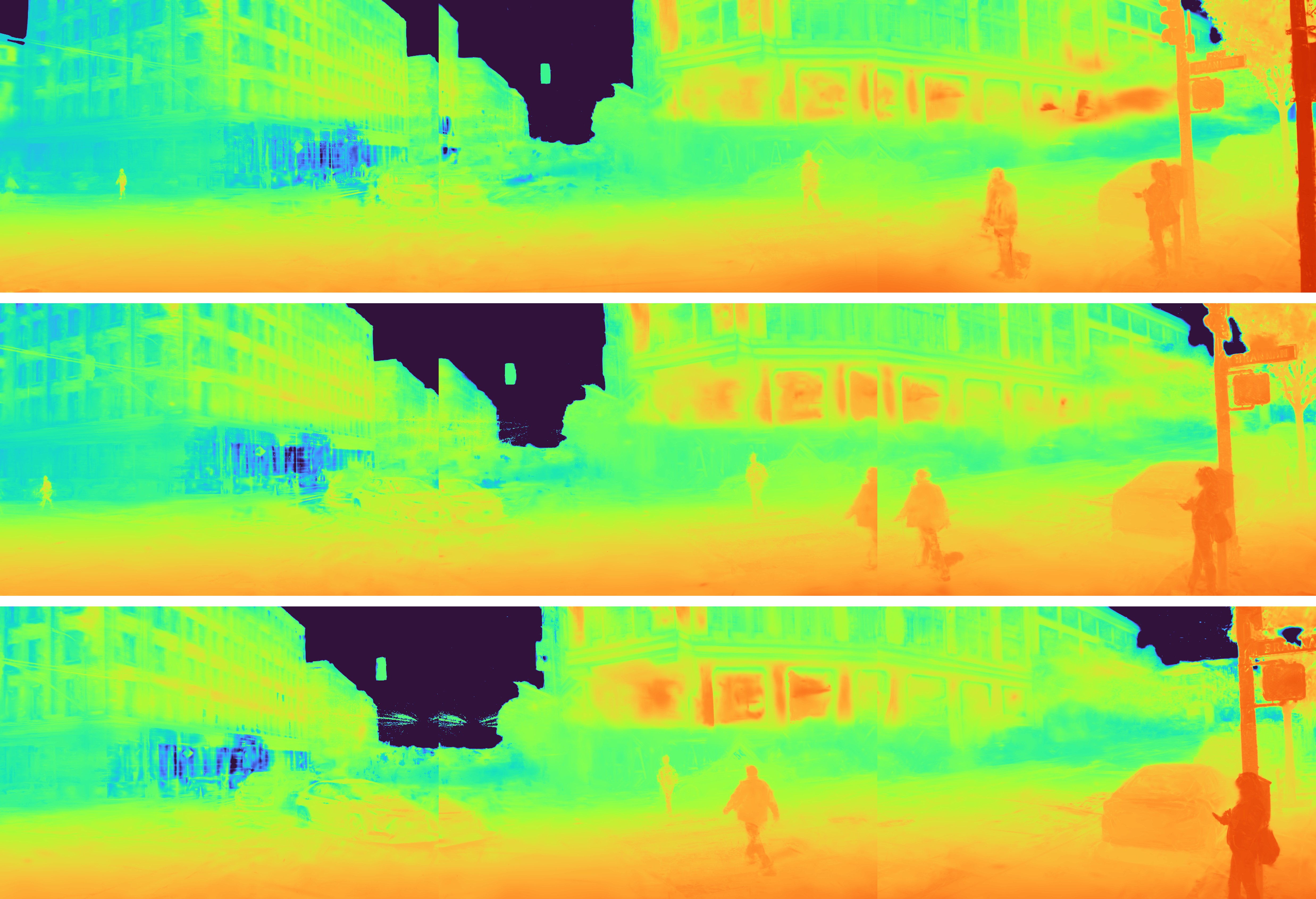}}
 \\
 \\
    {\includegraphics[width=.45\linewidth,clip]{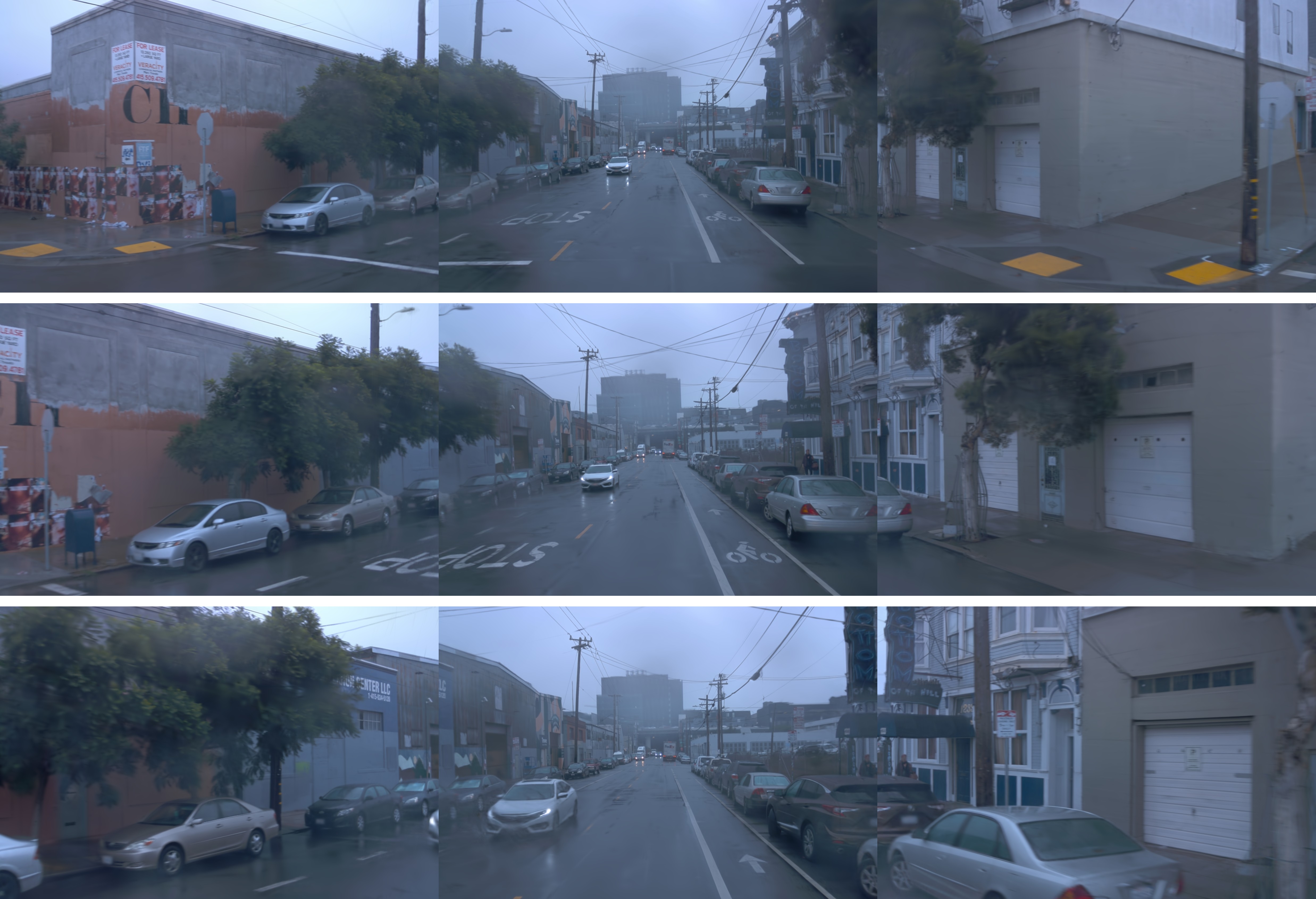}}
    &
    {\includegraphics[width=.45\linewidth,clip]{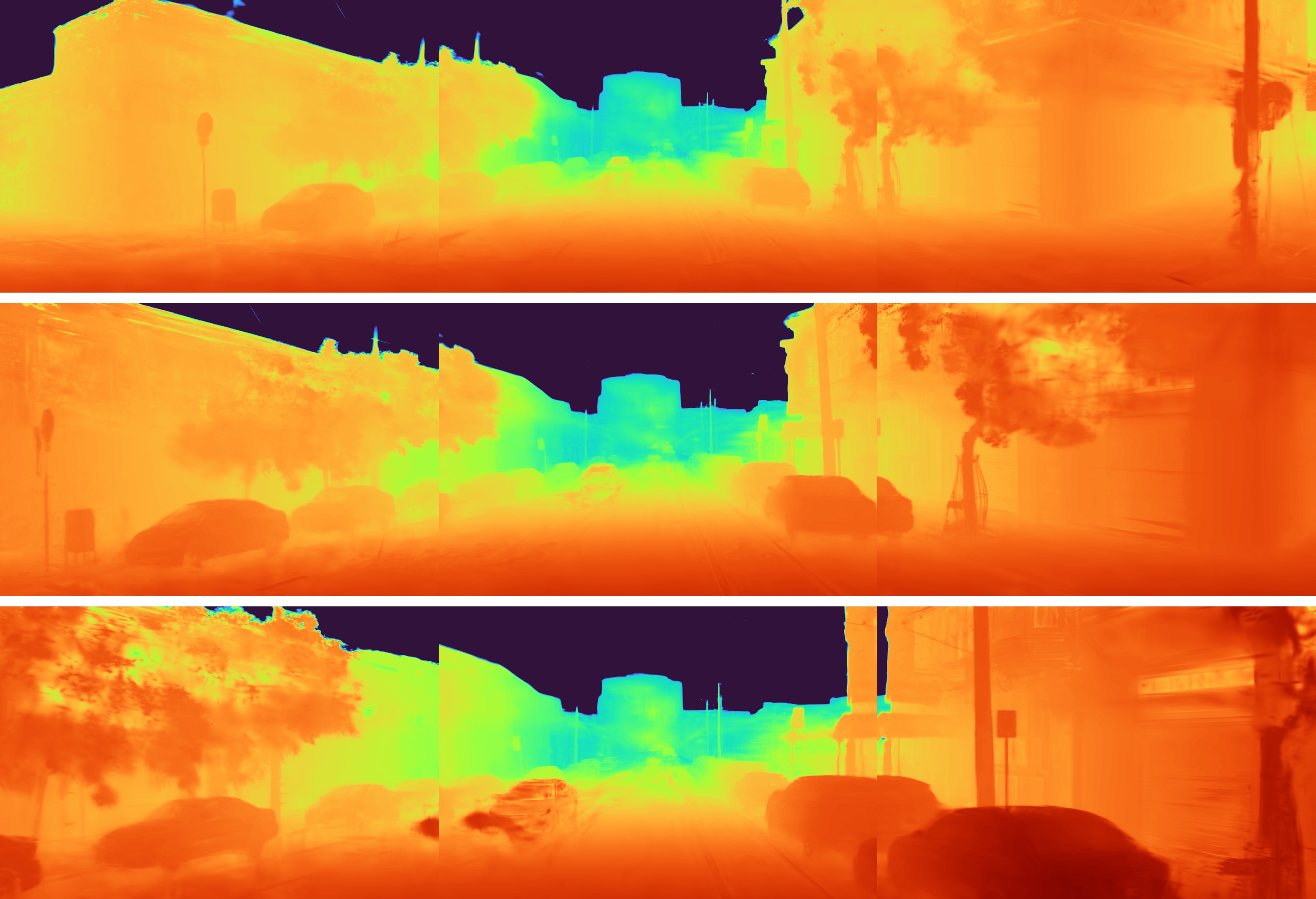}}
 \\
    \end{tabular}
    \caption{ {\color{cyan}\textbf{Visualization of urban scene synthesis.  
    } \textbf{Left}: rendered RGB images. \textbf{Right}: rendered depth maps. Best viewed with zoom-in.}}
\label{fig:urban}
\end{figure} 
\subsubsection{Experimental setup} 
We evaluate the Waymo-NOTR Dynamic-32 split curated by~\cite{yang2023emernerf} from Waymo Open Dataset~\cite{sun2020scalability}, with 32 sequences of calibrated images captured by five pinhole cameras and LiDAR point clouds. These driving sequences contain rich challenging dynamic objects. 
We divide each of them into clips of 50 consecutive frames for training and use images at resolution $640 \times 960$ from three frontal cameras in both training and evaluation. For reconstruction, we train and test on all frames of each clip. For novel view synthesis, we take out one frame every ten frames for testing and train on the remaining frames. The warm-up steps for all experiments is set to 3,000 unless otherwise specified.

\subsubsection{Results}
Table~\ref{tab:waymonotr} shows that our 4DGS (see Section~\ref{4dgsinthewild}) can achieve comparable and even better performance against those complex pipelines specifically designed for urban scenes.
Meanwhile, our fully explicit and unified representation supports highly efficient rendering, achieving superior efficiency over all competitors except 3DGS.
As shown in Figure~\ref{fig:urban}, 4DGS works well under diverse lighting and weather conditions. It faithfully reconstructs high-frequency texture details and correctly models the geometry for both dynamic and static regions.

\subsubsection{Ablations} 

\begin{table}[t]
\caption{{\color{cyan}
\textbf{Ablation studies on the dynamic urban scene.} All metrics are evaluated on the first clip of \texttt{seg104444}.
}}
\label{tab:ablation_urban}
\renewcommand\tabcolsep{3pt}
\renewcommand\arraystretch{1.2}
\scriptsize
\begin{tabular}{l|cc|cc}
\hline

\hline

\hline

\hline
\multirow{2}{*}{Settings} & \multicolumn{2}{c}{Recon.} \vline & \multicolumn{2}{c}{NVS} \\ 
 & PSNR$\uparrow$ & SSIM$\uparrow$ & PSNR$\uparrow$ & SSIM$\uparrow$ \\ 
\hline

\hline

4DGS & 37.05 & 0.964 & 29.45 & 0.880 \\
\hline
(a) +$\mathcal{L}_\mathrm{lidar}$ & 36.98 & 0.960 & 29.59 & 0.884 \\
(b) +$\mathcal{L}_\mathrm{sparse}$ & 36.17 & 0.961 & 30.22 & 0.898 \\
(c) +$\mathcal{L}_\mathrm{covt}$ & 36.08 & 0.956 & 30.49 & 0.904 \\
(d) +Rand. Perturb. & 35.60 & 0.953 & 30.59 & 0.907 \\
(e) +3D Warm-up & 36.08 & 0.956 & 31.46 & 0.915 \\
\rowcolor[gray]{.9}
(f) +$\mathcal{L}_\mathrm{rigid}$ (Full) & 36.94 & 0.959 & \bf 31.62 & \bf 0.916 \\

\hline

\hline
\end{tabular}
\end{table} 
We conduct ablation studies to investigate the impact of modifications for driving scenes. From Table~\ref{tab:ablation_urban}, we can draw these observations: (a) Introducing LiDAR supervision slightly increases performance for novel view synthesis, as relying solely on multi-view images to reconstruct the accurate geometry of textureless regions such as road surfaces is inherently difficult without additional prior constraints. (b) While strong capability of 4DGS,
it is challenging to constrain the behavior of Gaussians at a very novel time without static prior. $\mathcal{L}_{sparse}$ effectively addresses this by encouraging sparse motion and significantly boosts the rendering quality in novel views. 
(c) Applying regularization to the covariance in time dimension can increase the lifetime of the 4D Gaussian, enabling it to better utilize multi-view supervision. 
(d) Temporal random perturbation further reduces undesired motions, effectively enhancing the performance for novel view synthesis. (e) Warm-up and (f) $\mathcal{L}_{rigid}$ both slightly improve the performance of novel view synthesis by introducing reasonable priors about motion.

\subsection{Experiments for 4D Gaussian segmentation}
\label{sec:exp_seg}

We evaluate 4DGS for 4D segmentation (see Section~\ref{sec:segment}) with focus on visualization.

\noindent \textbf{Implementation details}
We set feature dimensions $D=32$ and $k=16$ in k-nearest neighbors.
We train the features and scale mapping based on compact 4DGS (see Section~\ref{compact4dgs}), pre-trained on the {\it Sear Steak} sequence from the Plenoptic Video dataset. The training is conducted with 10,000 iterations at a batch size of 1. 
In inference, as \cite{cen2023saga} we cluster Gaussians in 4D space based on the learned features using HDBSCAN, achieving automatic scene decomposition. For 4D segments, we first extract the 2D feature from the rendered feature map at a user-specific 2D point and then filter the 4D Gaussians whose smoothed scale-gated feature similarity with the query feature is below a certain threshold (typically 0.8).
The filtered 4D Gaussians represent a certain component most similar to the 2D point and can be rendered from any views at any timestamp. 
For both scene decomposition and 4D segments, we adjust the input scale from 0 to 1 to achieve multi-granularity segmentation.

\noindent \textbf{Results}
Figure~\ref{fig:segment} shows the multi-granularity segmentation results. The learned features enable the 4D Gaussians to cluster automatically and adaptively as the scale varies. At scale 0, individual elements like the man's hair, clothing, and arms are distinctly segmented. As the scale increases to 1, the entire figure of the man is segmented as a whole. This multi-granularity segmentation capability also extends to 4D segments. For example, the Gaussians representing the man's whole body, head, or arm can be identified at different scales with minimal blurry floaters.
} 
\section{Conclusion}

We introduced 4D Gaussian Splatting (4DGS), a novel framework for dynamic 3D scene representation, formulated as a spatio-temporal 4D volume learning problem. By leveraging 4D Gaussian primitives with anisotropic ellipses and 4D spherical harmonics, 4DGS effectively captures complex scene dynamics and enables real-time, high-resolution, photorealistic novel view rendering.
{\color{cyan}
To improve efficiency, we proposed compact variants that reduce memory usage and prevent overfitting. Extensive experiments validate the visual quality and efficiency of 4DGS across tasks like novel view synthesis, 4D generation, and scene understanding, proving its versatility across real and synthetic scenarios. 
}

\section{Data Availability Statements}
The authors declare that all datasets supporting the results and analysis in this study are publicly accessible. 
Plenoptic Video dataset~\cite{li2021neural} is available at \url{https://github.com/facebookresearch/Neural_3D_Video/releases/tag/v1.0}. 
Techhnicolor dataset~\cite{sabater2017technicolor} is provided at \url{https://www.interdigital.com/data_sets/light-field-dataset}.
D-NeRF dataset~\cite{pumarola2021d} can be downloaded from \url{https://www.dropbox.com/scl/fi/cdcmkufncwcikk1dzbgb4/data.zip?rlkey=n5m21i84v2b2xk6h7qgiu8nkg&e=1&dl=0}.
Waymo Open Dataset~\cite{sun2020scalability} is released at \url{https://waymo.com/open/data/perception}.
Consistent4D dataset~\cite{jiang2023consistent4d} is available at \url{https://drive.google.com/file/d/1mJNhFKvzZ-8icAw6KC-W-sf7JmmmMUkx/view}.

\backmatter

\bibliography{sn-bibliography}

\end{document}